\documentclass[10pt, journal]{IEEEtran}
\usepackage{amsmath,amsfonts}
\usepackage{array}
\usepackage[caption=false,font=normalsize,labelfont=sf,textfont=sf]{subfig}
\usepackage{textcomp}
\usepackage{stfloats}
\usepackage{url}
\usepackage{verbatim}
\usepackage{graphicx}
\def\BibTeX{{\rm B\kern-.05em{\sc i\kern-.025em b}\kern-.08em
    T\kern-.1667em\lower.7ex\hbox{E}\kern-.125emX}}
\usepackage{balance}
\usepackage{amsmath,amssymb,amsfonts}
\usepackage{graphicx}
\usepackage[dvipsnames]{xcolor}
\usepackage{pifont}
\usepackage{textcomp}
\usepackage{algorithm}
\usepackage{algpseudocode}
\usepackage{soul}
\usepackage{graphbox} 

\definecolor{newblue}{rgb}{0.2, 0.4, 0.6}
\definecolor{newgreen}{rgb}{0.2, 0.6, 0.2}

\usepackage{fancyhdr} \fancypagestyle{plain}{
\fancyhf{}

\fancyfoot[R]{\scriptsize{\copyright 2023 IEEE. Personal use is permitted, but republication/redistribution requires IEEE permission. See http://www.ieee.org/publications/rights/index.html for more information.}}
\fancyhead[C]{\scriptsize{This article has been accepted for publication in IEEE Transactions on Medical Robotics and Bionics. This is the author's version which has not been fully edited and content may change prior to final publication. Citation information: DOI 10.1109/TMRB.2023.3320267}}
  }
\begin{document}
\title{Bimodal Camera Pose Prediction for Endoscopy}
\author{Anita Rau, Binod Bhattarai, Lourdes Agapito, and Danail Stoyanov
\thanks{This work was supported by the Wellcome/EPSRC Centre for Interventional and Surgical Sciences (WEISS) [203145Z/16/Z]; Engineering and Physical Sciences Research Council (EPSRC) [EP/P027938/1, EP/R004080/1, EP/P012841/1]; The Royal Academy of Engineering Chair in Emerging Technologies scheme; and the EndoMapper project by Horizon 2020 FET (GA 863146). For the purpose of open access, the author has applied a CC BY public copyright license to any author accepted manuscript version arising from this submission.}
\thanks{All authors are affiliated with University College London, London, UK (e-mail:\{a.rau.16, b.bhattarai, l.agapito, danail.stoyanov\}@ucl.ac.uk). A. R. is also affiliated with Stanford University, Stanford, USA, and B. B. with University of Aberdeen, Aberdeen, UK. The authors would like to thank Javier Morlana from University of Zaragoza for providing the COLMAP results for real colonoscopy sequences, and Sophia Bano from UCL and the anonymous reviewers and editors for their constructive comments.}}

\markboth{IEEE TRANSACTIONS ON MEDICAL ROBOTICS AND BIONICS}%
{Bimodal Camera Pose Prediction for Endoscopy}

\maketitle
\thispagestyle{plain}

\begin{abstract}
Deducing the 3D structure of endoscopic scenes from images is exceedingly challenging. In addition to deformation and view-dependent lighting, tubular structures like the colon present problems stemming from their self-occluding and repetitive anatomical structure. In this paper, we propose \textit{{SimCol}}, a synthetic dataset for camera pose estimation in colonoscopy, and a novel method that explicitly learns a bimodal distribution to predict the endoscope pose. Our dataset replicates real colonoscope motion and highlights the drawbacks of existing methods. We publish 18k RGB images from simulated colonoscopy with corresponding depth and camera poses and make our data generation environment in Unity publicly available. We evaluate different camera pose prediction methods and demonstrate that, when trained on our data, they generalize to real colonoscopy sequences, and our bimodal approach outperforms prior unimodal work. Our project and dataset can be found here: \url{http://www.github.com/anitarau/simcol}.
\end{abstract}


\section{Introduction}
Reliably reconstructing the colon during colonoscopy from raw endoscopic images could improve cancer screening quality and substantially impact patient outcomes. A 3D model can help determine colon walls that were not sufficiently screened for polyps, as unobserved surfaces would appear as holes in the reconstruction. The colonoscope could then be guided to previously unobserved surfaces by estimating the current camera pose relative to the location of the holes within the reconstruction. Such a thorough examination of the entire colon wall is essential to guarantee the detection and removal of all polyps, which can prevent colorectal cancer from developing and increase survival rates \cite{kaminski2010quality}. 
Without computer assistance, polyp detection is challenging, and some studies suggest missing rates of precancerous lesions of over 90\% \cite{rex2017polyp}. So, especially in regions that lack sufficiently trained colonoscopists, providing automatic mapping and image-based navigation could increase the availability and quality of cancer screening services. A 3D map could also indicate the size and invasion of polyps and improve diagnosis. But predicting 3D structures from images of endoscopic scenes is challenging, and there is a vast gap between the performance of existing methods in urban scenes and endoscopic scenes.

\begin{figure}
 	\includegraphics[width=\columnwidth]{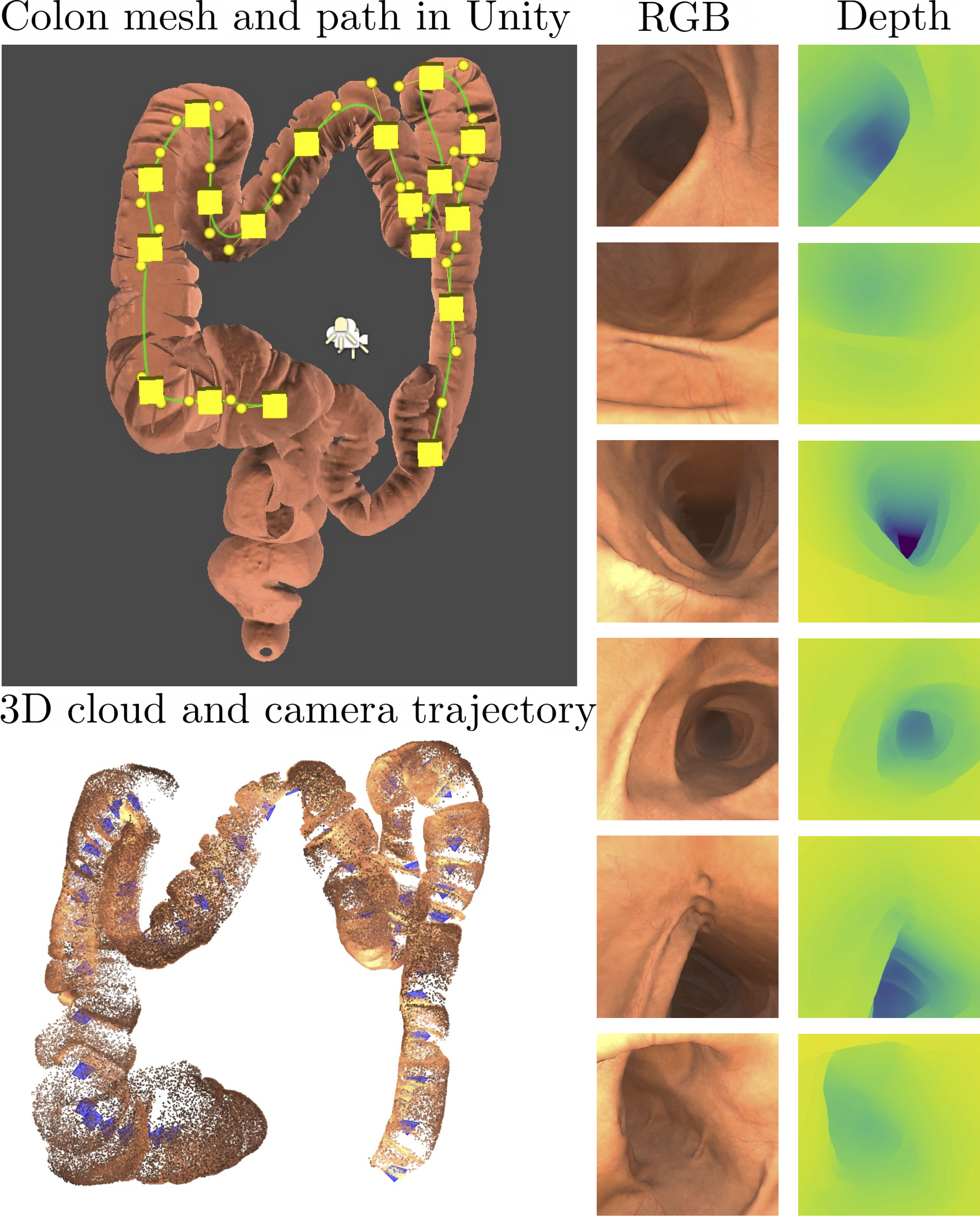}

	\caption{Our proposed dataset provides ground truth depth maps and camera poses. We show a virtual environment; an example virtual endoscope trajectory traveling through the reconstructed 3D cloud; and representative images with their depth maps next to them, where blue represents far surfaces.\label{fig:teaser_data}}
\end{figure}

Feature-based Structure-from-Motion (SfM) or Simultaneous Mapping And Localization (SLAM) are standard approaches for reconstructing outdoor or indoor scenes offline \cite{schoenberger2016sfm} or online \cite{mur2015orb}. Such methods detect features, establish correspondences, and estimate the relative camera motion and the 3D scene mutually and accurately. However, the application of feature-based systems to colonoscopic scenes suffers significant limitations that inhibit practical use in computer-assisted interventions (CAI). CAI systems have to address self-occlusion, camera-pose-dependent lighting and shadows, reflective surfaces, and a general lack of texture \cite{mahmoud2017slam}. Reconstructing the colon based on features thus remains (even in offline applications) an unsolved and challenging problem \cite{barbed2022superpoint}. 
\begin{table*}
\caption{An overview and comparison of colonoscopy or gastroscopy datasets}
{\def\arraystretch{1.4} 
\setlength{\tabcolsep}{3pt}
\begin{tabular}{|l|p{150pt}|c|c|c|c|c|c|c|c|c|}
\hline
{Dataset} & {Description} & {R/V/P} & {Public} & {Depths} & {Camera pose}& {Intrinsics} & {Tubular}  & \# {Frames} & \# {Trajectories} \\
\hline
\hline
Armin \textit{et al.} \cite{armin2017learning} & Simulated colonopscoy video &V&$\times$&?&\checkmark&?&\checkmark&30k&$>$15 \\\hline
Bae \textit{et al.}\cite{bae2020deep} &SfM labels from colonoscopy video&R&$\times$&$\times$&$\times$&?&\checkmark&$>$34k&51 \\\hline
Freedman \textit{et al.} \cite{freedman2020detecting} &Rendered synthetic dataset&V&$\times$&\checkmark&?&?&\checkmark&187k&?\\\hline
Fulton \textit{et al.}\cite{fulton2020comparing}&Colon phantom magnetic tracker dataset&P&\checkmark&$\times$&\checkmark&\checkmark&\checkmark&24k&7 \\\hline
Ma \textit{et al.} \cite{ma2019real}&COLMAP labels from colonoscopy video&R&$\times$&$\times$&$\times$&?&\checkmark&1.2m& 60\\\hline
Ozyoruk \textit{et al.} \cite{ozyoruk2021endoslam} &Ex-vivo porcine 3D scanned colon & R & some & \checkmark & \checkmark & \checkmark & $\times$ &  $>$8k & 18 \\\hline
Ozyoruk \textit{et al.} \cite{ozyoruk2021endoslam}& Simulated capsule endoscope in \textit{Unity}  & V & some& \checkmark & \checkmark & \checkmark & \checkmark &  22k & 1 \\\hline
Turan \textit{et al.}\cite{turan2018unsupervised} &Ex-vivo porcine stomach&R&$\times$&$\times$&\checkmark&\checkmark&$\times$&12k&$>$ 4 \\\hline
Widya \textit{et al.} \cite{widya2019whole} &Stomach with real meshed texture &R/V&$\times$&\checkmark&\checkmark&\checkmark&$\times$&?&7 \\\hline
Zhang \textit{et al.} \cite{zhang2020template} & Stereoscopic colonoscopy w/ features & V & \checkmark  & \checkmark & \checkmark & \checkmark & \checkmark & 12k & 16\\\hline
\textbf{Ours} & Simulated colonoscope in \textit{Unity} & V & \checkmark & \checkmark & \checkmark & \checkmark & \checkmark &  18k & 15 \\

\hline
\end{tabular}}\\ $ $ \\
R = Real, V = Virtually simulated, P = Physical phantom, ? = Could not be verified, $>$ could be partially verified but it appears there are more.

\label{tab:datacomparison}
\end{table*}
Learning methods do not necessarily rely on features and therefore have great potential to further 3D reconstruction during colonoscopy. Accurately predicted camera pose and scene depth can be an initial estimate within an SfM or SLAM pipeline, provide information about certainty, or, downstream, enable the direct regression of scene coordinates and camera locations in an end-to-end fashion \cite{tang2018ba, zhao2018learning}. Depth prediction during colonoscopy has already been studied and can be considered robust \cite{mahmood2018unsupervised, rau2019implicit, mathew2020augmenting}. However, predicting camera poses in this context remains challenging.

Working toward a learning-based system requires vast amounts of training data. However, labels for real colonoscopy are not readily available: depth and pose sensors cannot fit into a standard colonoscope. Therefore, existing methods have focused on surrogate data with ground truth (simulated or porcine) or self-supervised approaches. Unfortunately, existing simulated and porcine datasets are only partially publicly available. Further, ex-vivo porcine does not look similar to in-vivo human colonoscopy, and simulated datasets do not account for the authenticity of the simulated camera movement. We argue that for pose estimation, not only must the appearance of a synthetic dataset be realistic, but the simulated camera movements also must replicate the movements of a real colonoscope during colonoscopy.

One might argue that self-supervised methods are the remedy: They do not require labels at all as they instead mutually train a depth and a pose network \cite{ozyoruk2021endoslam} using warping losses. However, we show that such methods are \textit{not} guaranteed to converge to accurate depths and poses. When poses are bimodally distributed, we show that self-supervised methods can learn a distribution with one maximum, instead of two.

We, therefore, introduce a new, fully publicly available benchmark dataset that will enable the development and benchmarking of supervised camera pose prediction and 3D reconstruction methods during colonoscopy. The simulated camera poses replicate real camera movement during screening more closely than existing data and help networks generalize to real data. Additionally, we propose a novel supervised pose regression network that explicitly models a bimodal distribution through a combined classification-regression framework. The classification network classifies the camera movement into insertion or withdrawal, and the regression network predicts the difference from the respective class mean.

To work toward more accurate relative camera pose predictions during colonoscopy in this paper we:
\begin{itemize}
    \item create and publish an extensive synthetic dataset of over 18,000 images with rendered RGB images, depth  information, camera poses, camera intrinsics, and detailed documentation;
    \item explain why unsupervised methods are not readily applicable to infer relative camera poses accurately;
    \item propose a supervised bimodal approach that is robust towards a lack of features and more accurately predicts our dataset's bimodally distributed camera movement
\end{itemize}

\section{Related Work}

The availability of real colonoscopy training data with camera poses and scene depths is limited. Existing convolutional neural nets (CNNs) for camera pose estimation in colonoscopy thus focus mainly on synthetic data, labels from ex-vivo porcine, or unsupervised methods.

\noindent\textbf{Unsupervised depth and pose prediction:}
Self-supervised methods require neither ground truth nor pseudo labels. They depend instead on carefully designed loss functions.
 Turan \textit{et al.}, Ozyoruk \textit{et al.}, Yao \textit{et al.}, and Widya \textit{et al.} all employ warping errors to supervise a depth and pose prediction network in gastroscopy or colonoscopy \cite{turan2018unsupervised, ozyoruk2021endoslam, yao2021motion, widya2021learning}. Though an unsupervised approach enables training in an environment where it is near impossible to obtain good ground truth, we identify several shortcomings of existing approaches in this work. One drawback is that unsupervised methods can predict unimodal distributions even though the underlying data is bimodally distributed. We discuss this phenomenon in detail in Section \ref{sec:limitations}.
 
 \noindent\textbf{Integrated CNN \& SfM pipelines:}
Sub-tasks of the traditional SfM pipeline can be learned with CNNs; vice-versa, SfM outputs can drive CNNs. In both scenarios, the method heavily relies on SfM pseudo labels. However, standard SfM fails more often than not in colonoscopy, and methods that rely on their outputs can only learn to be as good as these pseudo labels.
 Ma \textit{et al.} \cite{ma2019real} incorporate the output of a depth network into their SfM pipeline and predict an initial pose using a recurrent neural net that is refined within a traditional SLAM pipeline. Their CNN is trained on sparse SfM-depth. Bae \textit{at al.} \cite{bae2020deep} propose a multi-view stereo algorithm for dense depth that is able to match patches between images but depends on an initial SfM camera pose and sparse depth estimation.
 
\noindent\textbf{Pose regression with CNNs:}
When simulated data or ex-vivo porcine data is available, networks can learn in a supervised manner. Armin \textit{et al.} \cite{armin2017learning} train a CNN on synthetic data to directly output the 6DoF vector describing the relative rotation and translation between two cameras during colonoscopy. The back-propagated error is a weighted sum of squared translation and rotation error. Turan \textit{et al.} \cite{turan2018deep} incorporate information from entire video sequences of the upper GI-tract using recurrent neural networks to enable time-consistent camera pose predictions. Their network directly regresses a 6DoF vector from two consecutive RGB images and their estimated corresponding depth maps. Zhang \textit{at al.} \cite{zhang2020template} predict camera pose by registering predicted disparity from synthetic stereo images to a pre-operative CT scan. However, the applicability to real colonoscopy is limited due to the lack of stereo images and CT scans during real procedures. While supervised approaches tend to be accurate, they require data that is not widely available.

\noindent \textbf{Datasets for camera pose estimation in colonoscopy:}
Different works have used a variety of datasets, but only a few published them. Table \ref{tab:datacomparison} enumerates the datasets referenced in relevant papers. To the best of our knowledge, there are only two fully available dataset for colonoscopy. One is based on a phantom \cite{fulton2020comparing} and investigates the effects of deformation. It was created using an electro-magnetic tracker attached to an endoscope that provides 6DoF poses. However, the structures are so repetitive that any forward movement could be interpreted as a backward movement looking at the previous fold rather than the next fold. The second fully available dataset is a synthetic dataset generated from a virtual colon and colonoscope \cite{zhang2020template}. It was designed for stereo matching and is characterized by pronounced synthetically generated features enabling the employment of feature matching methods. While the dataset is instrumental for stereo methods, stereo colonoscopy is not common and feature matching remains unreliable in real colonoscopy. Ozyoruk \textit{et al.} created a real and a simulated dataset and made some trajectories publicly available \cite{ozyoruk2021endoslam}. While it is the first of its kind for real data and is immensely helpful, its usefulness to the research community could be limited. Creating the real dataset required a porcine colon to be mounted to scaffolding. The mounting prevented the visualization of the typical shape of the colon and the camera's pointing towards the lumen. As a result, the published trajectories consist mostly of colon wall images. A virtual capsule endoscope generated the simulated dataset in Unity while a user "steered" the camera through a colon mesh. While this may replicate a colonoscopy with a magnetic pill-shaped capsule, the data does not reflect the movement of a standard colonoscope. Only 0.004\% of consecutive frame pairs in the published trajectory have a non-zero rotation \cite{EndoData}. A fully documented monocular colonoscopy benchmark dataset based on the movement of a standard colonoscope with a clear distinction between training and test data is still missing.
\section{Methods}
Our motivation is based on a prominent prior work on self-supervised depth and pose prediction \cite{ozyoruk2021endoslam}. We develop and make available a novel dataset and describe the data generation process and properties of our data. Then we propose a novel, bimodal approach to camera pose estimation that is better suited to learning bimodally distributed data. Our model, the full dataset, the documentation, and the data generation environment will be publicly available.

\subsection{A study on the limitations of self-supervised depth and pose prediction in colonoscopy.}

\label{sec:limitations}
\begin{figure}
	\centering
  \includegraphics[width=0.9\linewidth]{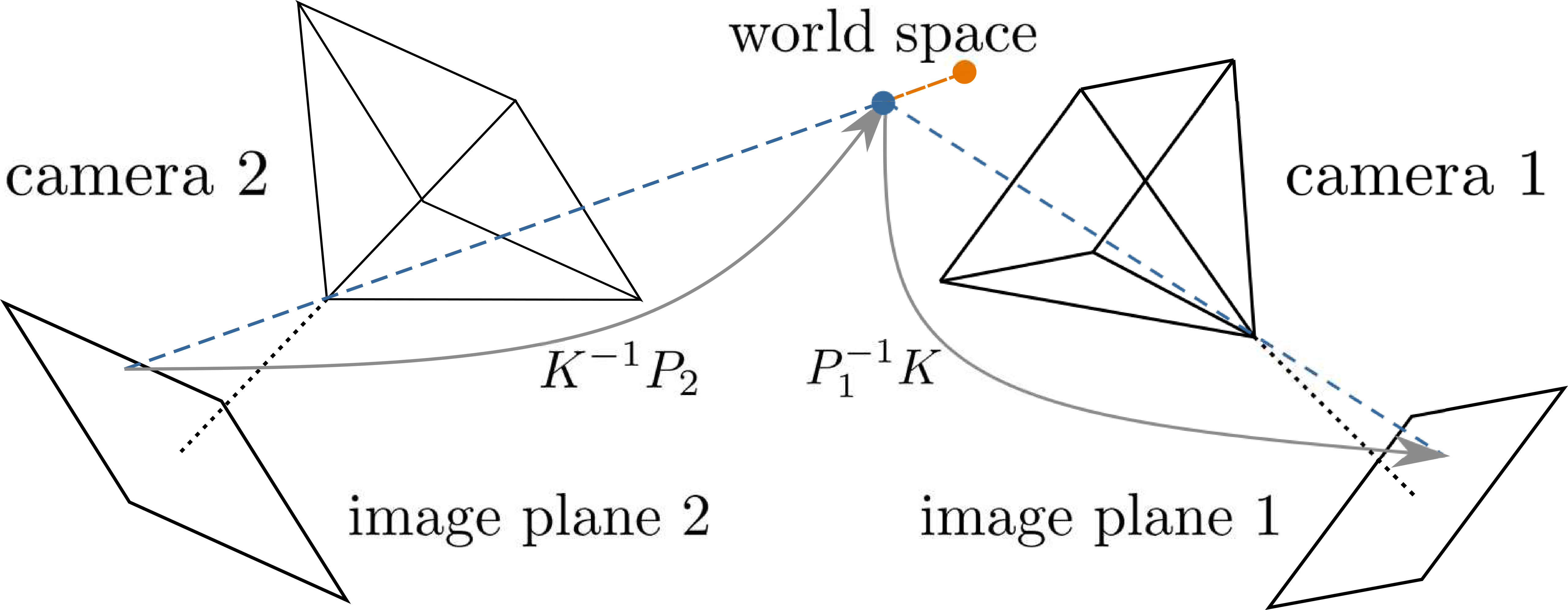}
 	\caption{Illustration of image warping. The orange point is occluded in image 2 and would not appear in warped image 1.} \label{fig:cameras}
\end{figure}

Ideally, a network can learn reliable depth and pose estimates from image pairs. Depth and pose are inherently related through the geometry of the scene. 
Self-supervised approaches are based on the idea that predicted depth and relative pose between two cameras could be used to warp an image to a nearby frame. A loss function then compares the warped image to the target image. We define the pose of a camera in a reference world space at time $\tau$ as a $4\times 4$ matrix
\begin{align*}
    \textbf{P}_{\tau} = \begin{bmatrix}\textbf{R}_{\tau} & \textbf{t}_{\tau} \\
    \textbf{0} & 1
\end{bmatrix},
\end{align*}
where $\textbf{R}$ is a $3\times 3$ rotation matrix and $\textbf{t}$ is a $3\times 1$ translation vector. We can project a pixel into the respective camera frame using the camera intrinsics. 

As illustrated in Figure \ref{fig:cameras}, we can \textit{warp} pixels between images.
We project a pixel to its camera space using the intrinsic matrix $\textbf{K}$. The camera's pose in the world space can then project a point $\textbf{h} = [x,y,z,1]$ in homogeneous coordinates into world space. The inverse of a second pose lets us project the point into the new camera space from where $\textbf{K}$ maps it to the new image plane. We summarize:
\begin{align}
    \textbf{h}' = \textbf{P}_1^{-1}\textbf{P}_2\textbf{h} =: \mathbf{\Omega} \textbf{h},
\end{align}
where $\textbf{h}'$ is the image of a point $\textbf{h}$ in a new camera frame and $\mathbf{\Omega}$ is a $4\times 4$ projection matrix.
It is important to note that world points occluded by structures in one image can not be retrieved through warping. For a thorough discussion of projective geometry, we refer to Hartley and Zisserman \cite{hartley2003multiple}. 

Similarly to projecting points between camera spaces, we can map a camera pose to a new pose in the world space. In other words, we rotate and translate a camera. To map camera 1 to the pose of camera 2, we write 
\begin{align}
    \textbf{P}_2 = (\textbf{P}_1\textbf{P}_1^{-1})\textbf{P}_2=\textbf{P}_1(\textbf{P}_1^{-1}\textbf{P}_2) = \textbf{P}_1\mathbf{\Omega}.
\end{align}
In the context of mapping camera poses we refer to $\mathbf{\Omega}$ as the \textit{relative camera pose}. Self-supervised methods technically learn the warping parameter $\mathbf{\Omega}$, but it can be interpreted as the relative camera pose.

To understand how self-supervised methods learn $\mathbf{\Omega}$,
we define the relevant losses that compare a target image to the predicted warped frame. Let Images $I_{\tau}$, $I_{\tau+k}$ be two nearby RGB images that are $k$ frames apart. For synthetic data, let $D_{\tau}$, $D_{\tau+k}$ be the corresponding ground truth depth maps and let $\mathbf{\Omega}_{\tau \rightarrow \tau + k}$ be the projection matrix that projects points from camera frame $\tau$ to camera frame $\tau + k$. Further, let hats denote the predictions of a network: $\hat{D}_{\tau}$ and $\hat{\mathbf{\Omega}}_{\tau \rightarrow \tau + k}$. Note that networks can parameterize camera translations and rotations differently, but other representations can be mapped to $4 \times 4$ projection matrices \cite{blanco2010tutorial}.

Then image ${I}_{\tau}$ can be warped to look like ${I}_{\tau+k}$ and the warped image is denoted through a tilde as $\tilde{{I}}_{\tau + k}(\mathbf{\Omega}_{\tau + k \rightarrow \tau}, D_{\tau+k}, I_{\tau})$, based on the inverse projection, target depth, and reference image used. For dense warping, we use \textit{inverse warping}; that is, we warp the target image back to the reference image rather than warping the reference to the target.
The reprojection loss for an image pair $(I_{\tau}, I_{\tau+k})$ is defined as
\begin{align}
    L_R = \sum_{i=-k,k}\sum_{\tau}||I_{\tau+i} - \tilde{I}_{\tau+i}(\hat{\mathbf{\Omega}}_{\tau+i\rightarrow\tau},\hat{D}_{\tau+i}, I_{\tau})||_1.
\end{align}
Other relevant losses are the geometric consistency loss $L_G$ \cite{ozyoruk2021endoslam} that compares the warped depth $\tilde{D}_{\tau+k}$ to the projected depth $D'_{\tau+k}$,
and the structural similarity index measure (SSIM), here referred to as $L_S$, as defined in \cite{wang2004image}. 

\begin{figure*}
	\centering
 \begin{minipage}{\columnwidth}
 \centering
     \includegraphics[width=\linewidth]{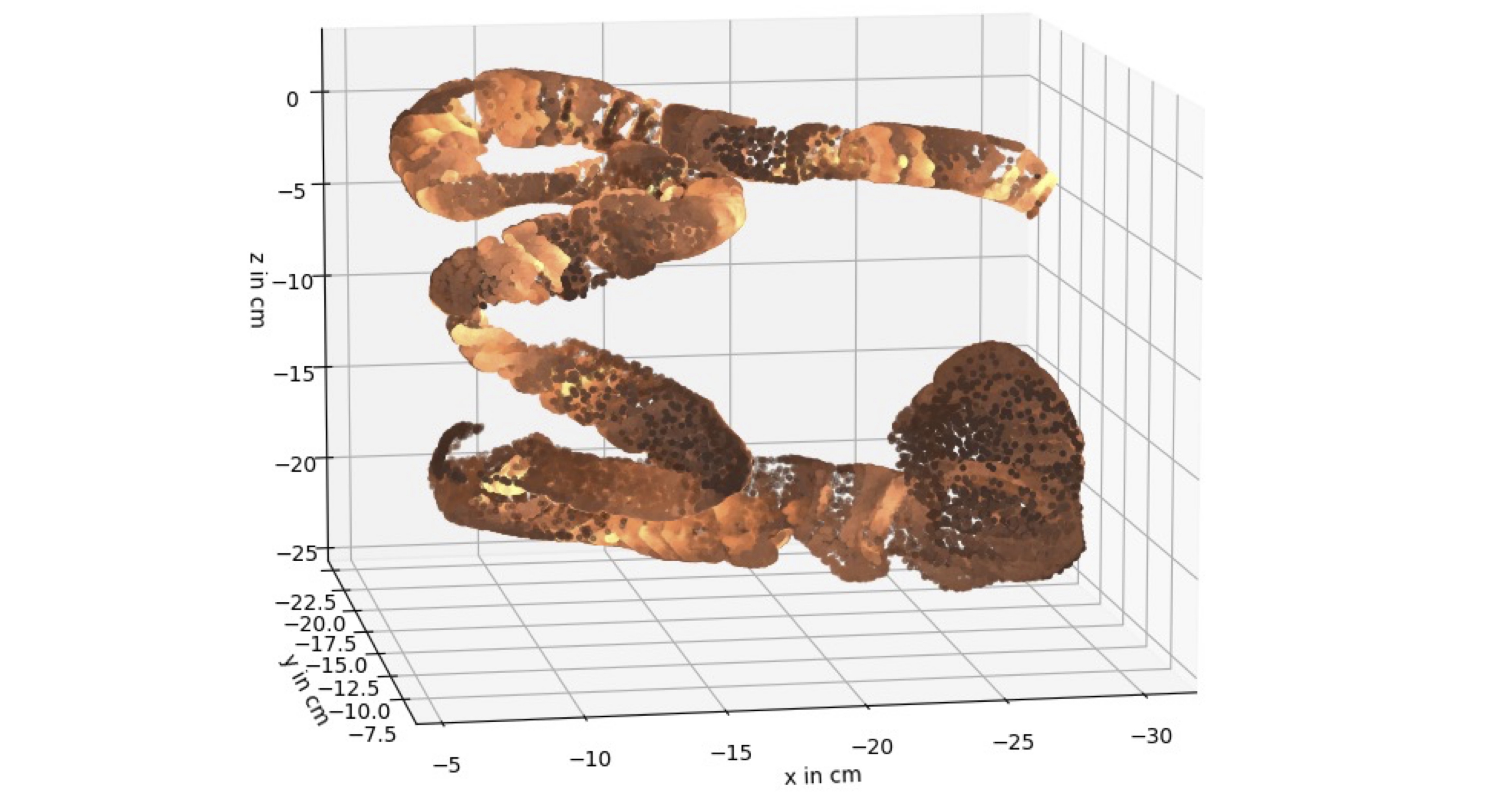}
     \footnotesize{(a)}
 \end{minipage}
  \begin{minipage}{\columnwidth}
  \centering
     \includegraphics[width=\linewidth]{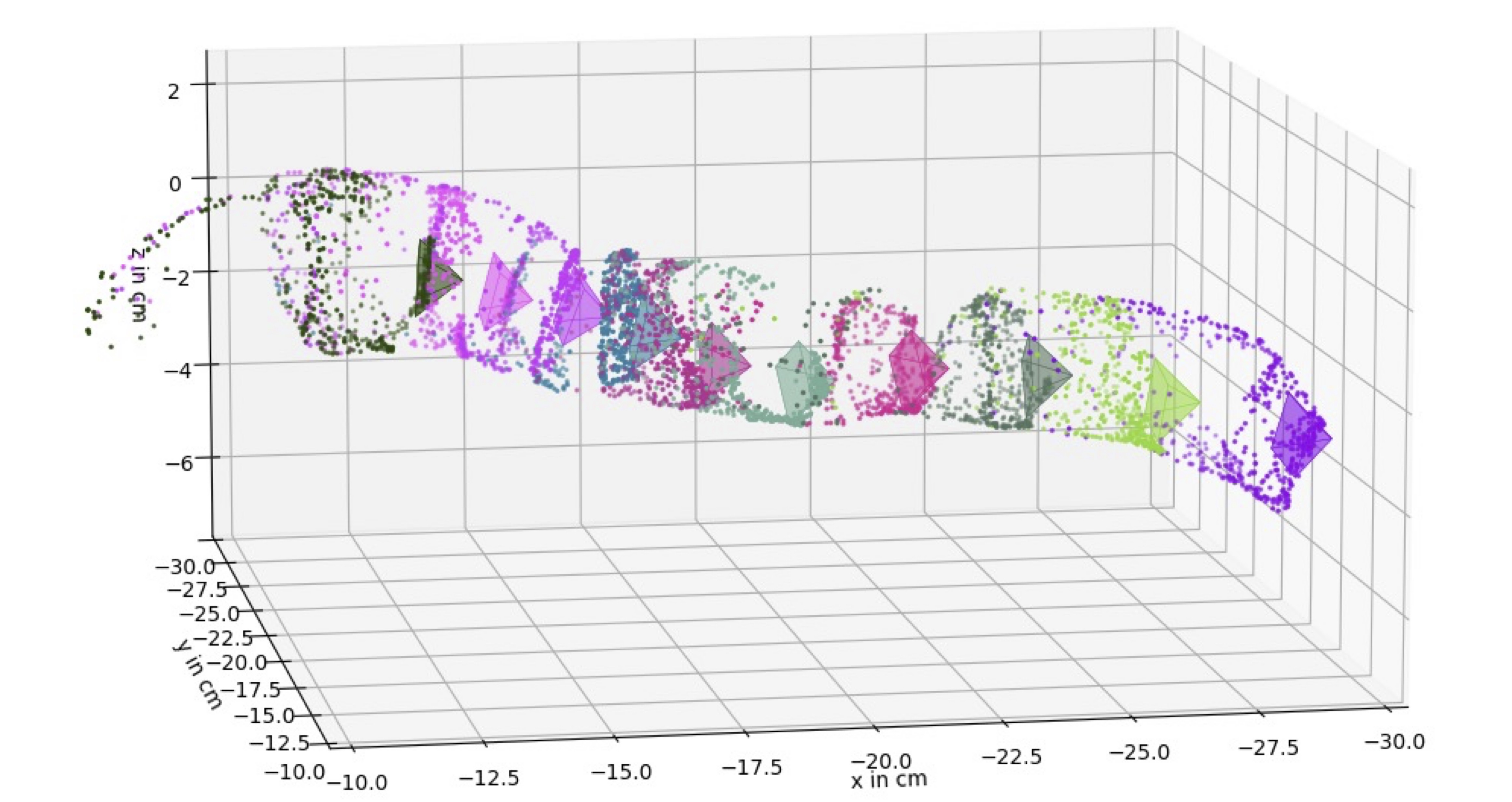}
     \footnotesize{(b)}
 \end{minipage}
	\caption{Projecting ground truth depth and camera poses into world space replicates the colon mesh. (a): Every tenth frame of one test trajectory is projected into the same world space. (b): Every twentieth frame of a smaller section in the same test sequence is projected into world space together with the respective camera poses. Each camera and its point cloud have the same random color, and clouds are subsampled for better visualization. \label{fig:3d_data}}
\end{figure*}
But due to the geometry of the colon, and the related characteristic camera movement, mapping between image plane and 3D world is ambiguous (and certainly not bijective):

\noindent\textbf{Illumination inconsistency:} Camera-pose-dependent lighting interferes with warping losses: The same surface can look very different from slightly different positions because the light source is attached to the endoscope. Adjusting for illumination inconsistency by scaling the intensity values of an image pair to have the same mean \cite{ozyoruk2021endoslam} only linearly approximates the illumination differences. It neglects, for instance, that colon folds throw shadows and block light from reaching farther parts of the lumen. Illumination is an important cue for understanding geometry in the mostly textureless colon. Its inconsistency challenges the applicability of warping-based methods in the colon.

\noindent\textbf{Smoothness inertia:} Networks tend to predict smooth depth maps when trained with naive warping loss, and this is often even reinforced by a smoothness loss \cite{zhou2017unsupervised}. While this might be a reasonable approach for smooth surfaces like bronchi, the geometry of the colon is characterized through folds and thus sharp steps in depth.

\noindent\textbf{Back-projection ambiguity: } Due to self-occlusion warping errors and geometric inconsistency often are not minimized by the ground truth depth and pose---the network converges to a local minimum.

\noindent\textbf{Wide field of view:} The wide frustum of colonoscopes emphasizes nearby structures over-proportionally in the  image.

We show experimentally how these drawbacks can lead to a sub-optimal accuracy of the predicted camera pose. 

\subsection{\textit{SimCol}: A new dataset}

We propose a virtual dataset for pose and depth prediction in colonoscopy. Following the pipeline introduced in \cite{rau2019implicit}, we use a \textit{Unity} simulation environment to generate our data; however, we additionally generate ground truth camera poses. The colon mesh used for simulation is based on a computer tomography scan of a real human colon \cite{ozyoruk2021endoslam}.

We define a path in the center of the rigid colon mesh defined through 18 \texttt{WayPoints} which the virtual colonoscope follows, simulating a traversal through a colon.  
Our dataset aims to reflect typical camera movements during colonoscopy. Fully inserting the colonoscope into the colon is challenging and can at times require two operators, the application of hand pressure on the abdomen, and repeated retraction and reinsertion of the colonoscope \cite{williams2009insertion}. Therefore, screening for polyps is performed during withdrawal, when the scope is slowly pulled backwards along the colon's centerline. When folds present themselves, the tip is flexed and rotated to observe the entire mucosa. Reinsertion and re-withdrawal also often occur \cite{rex2003missed}. 

Similarly, the virtual camera follows the centerline of the colon while rotating around it. The camera outputs images with size 475x475 and replicates the intrinsics of a real colonoscope. Two virtual light sources are attached to the left and the right of the camera. Each time a user renders images along the trajectory, the  \texttt{WayPoints} are independently offset by a random translation of up to 2mm, and a rotation of up to 20 degrees, which we chose experimentally. 

When generating new data, a user can change the rotation, position, randomization, and number of \texttt{WayPoints} arbitrarily. The user can also adjust the interpolation parameters between \texttt{WayPoints}. Adjusting a \texttt{WayPoint}  will affect all frames' camera poses between the previous and the following \texttt{WayPoint}. The camera intrinsics, image resolution, and light sources can also be adjusted in our Unity project. 
Given all poses as $4\times 4$  projection matrices describing the camera's pose in the world reference frame, we assume the first camera ($\tau$) of each pair to be at the origin. Thus we learn the projection $\mathbf{\Omega}_\tau = {\textbf{P}}_{\tau}^{-1} {\textbf{P}}_{\tau+k}$, that describes the position of camera 2's ($\tau+k$) origin as seen from camera 1. Accordingly, a positive z-coordinate of $\mathbf{\Omega}$ indicates that camera 2 is in front of camera 1. As \textit{Unity} uses a left-handed system, while the \textit{Python} package \textit{Scipy} uses a right-handed system, we transform the \textit{Unity} camera poses during training and testing using the projection ${\textbf{P}}^{right} = {\textbf{M}}{\textbf{P}}^{left}{\textbf{M}}$, where $\textbf{M}$ is a $4\times 4$ identity matrix with $-1$ as the second diagonal element. After transformation, the z-axis points forward from the optical center, while the y-axis points upward and the x-axis to the right.
\begin{figure}

\centering
	\begin{minipage}{0.45\columnwidth}
		\centering
  \includegraphics[width=\linewidth]{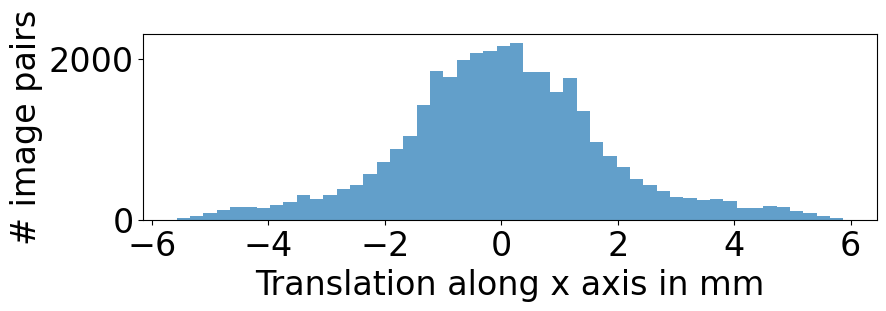}
	\end{minipage}
	\begin{minipage}{0.45\columnwidth}
	\centering
 \includegraphics[width=\linewidth]{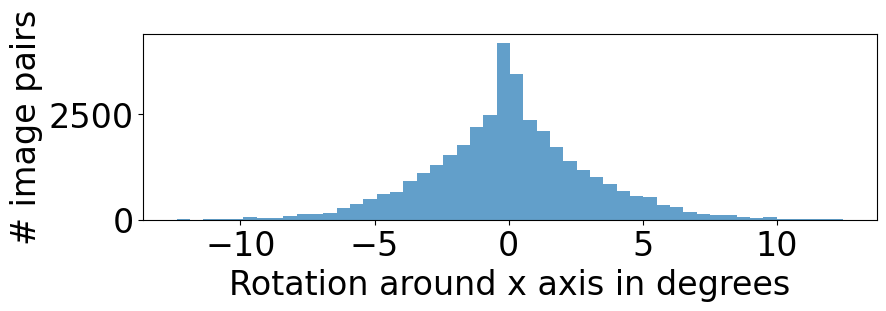}
	\end{minipage} \vspace{1mm} \\
	\begin{minipage}{0.45\columnwidth}
	\centering
 \includegraphics[width=\linewidth]{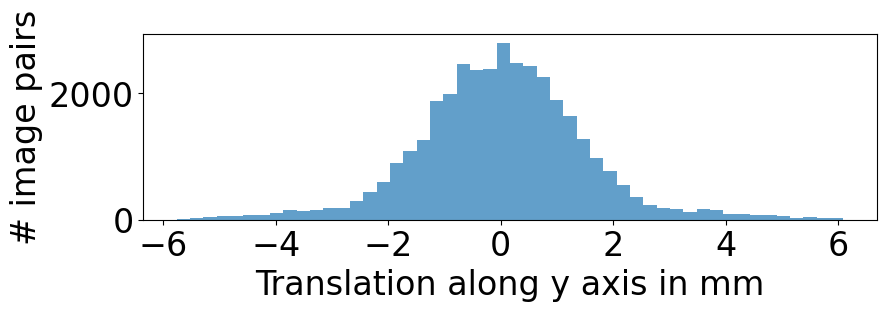}
	\end{minipage}
	\begin{minipage}{0.45\columnwidth}
	\centering
 \includegraphics[width=\linewidth]{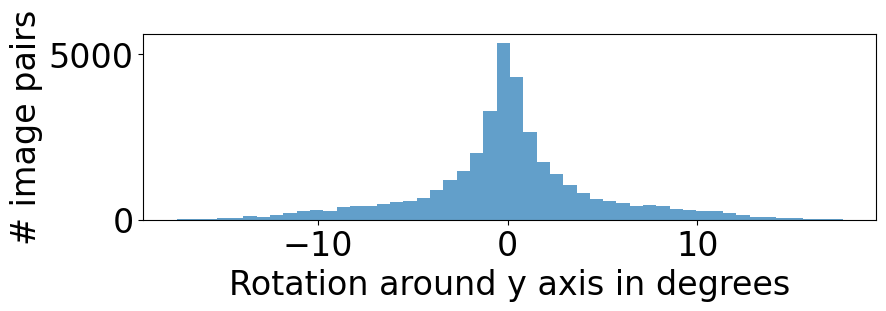}
	\end{minipage} \vspace{1mm} \\
	\begin{minipage}{0.45\columnwidth}
		\centering
  \includegraphics[width=\linewidth]{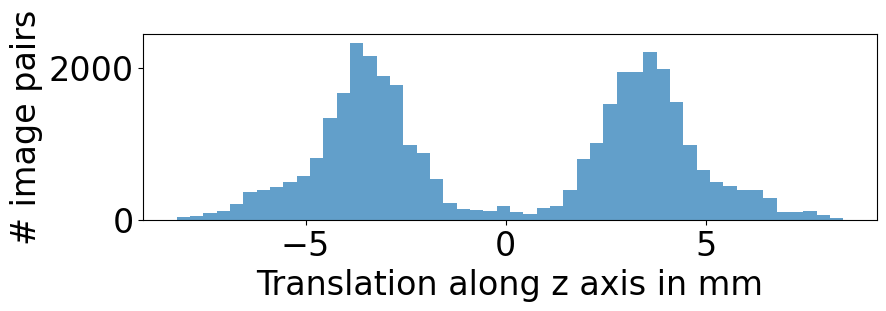}
	\end{minipage}
	\begin{minipage}{0.45\columnwidth}
		\centering
  \includegraphics[width=\linewidth]{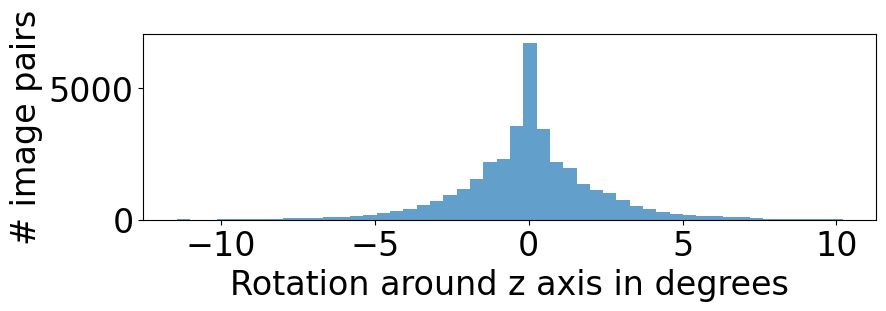}
	\end{minipage}

	\caption{Histograms of relative $x$-, $y$-, and $z$- translations and rotations between all image pairs in all 15 sequences of our dataset that are $k=5$ frames apart. The translation along the z-axis is clearly bimodal. This observation corresponds to the endoscope's predominant insertion and withdrawal movement within the colon.}
	\label{fig:histo}
\end{figure}
\begin{algorithm}
\caption{Data generation in Unity}\label{alg:cap}

\begin{algorithmic}

\For{{wayPoint} \textbf{in} \texttt{WayPoints}}
    \State randomize {wayPoint} pose
    \State time = 0
\While{time $<$ timePerWayPoint}
    \State $\text{cam.position} \gets GetPosition(wayPoint, time)$
  \State $\text{cam.rotation} \gets GetRotation({wayPoint}, {time})$
  \State render RGB image
  \State render depth map
\State update time
\EndWhile
\EndFor

\end{algorithmic}
\end{algorithm}
Figure \ref{fig:teaser_data} shows example images, depths, and trajectories next to the colon mesh in its \textit{Unity} environment. The Figure shows the camera path in green and the \texttt{WayPoints} in yellow. Algorithm \ref{alg:cap} gives an overview of the data generation process. Furthermore, we show how the depth maps and poses can be used to project RGB images into 3D space in Figure \ref{fig:3d_data}(a), and we visualize the ground truth camera poses of a smaller section and their respective point clouds in (b). Lastly, we plot histograms of the x-, y-, and z-rotations and translations for consecutive frame pairs in Figure \ref{fig:histo}. It is important to note that the translation along the z-axis is clearly bimodal, as we would expect in a real scenario. Further, the rotations range in an interval of about [-2,2] degrees. 

\subsection{A new approach to predict bimodal camera pose}
\begin{figure}
\includegraphics[width=\columnwidth]{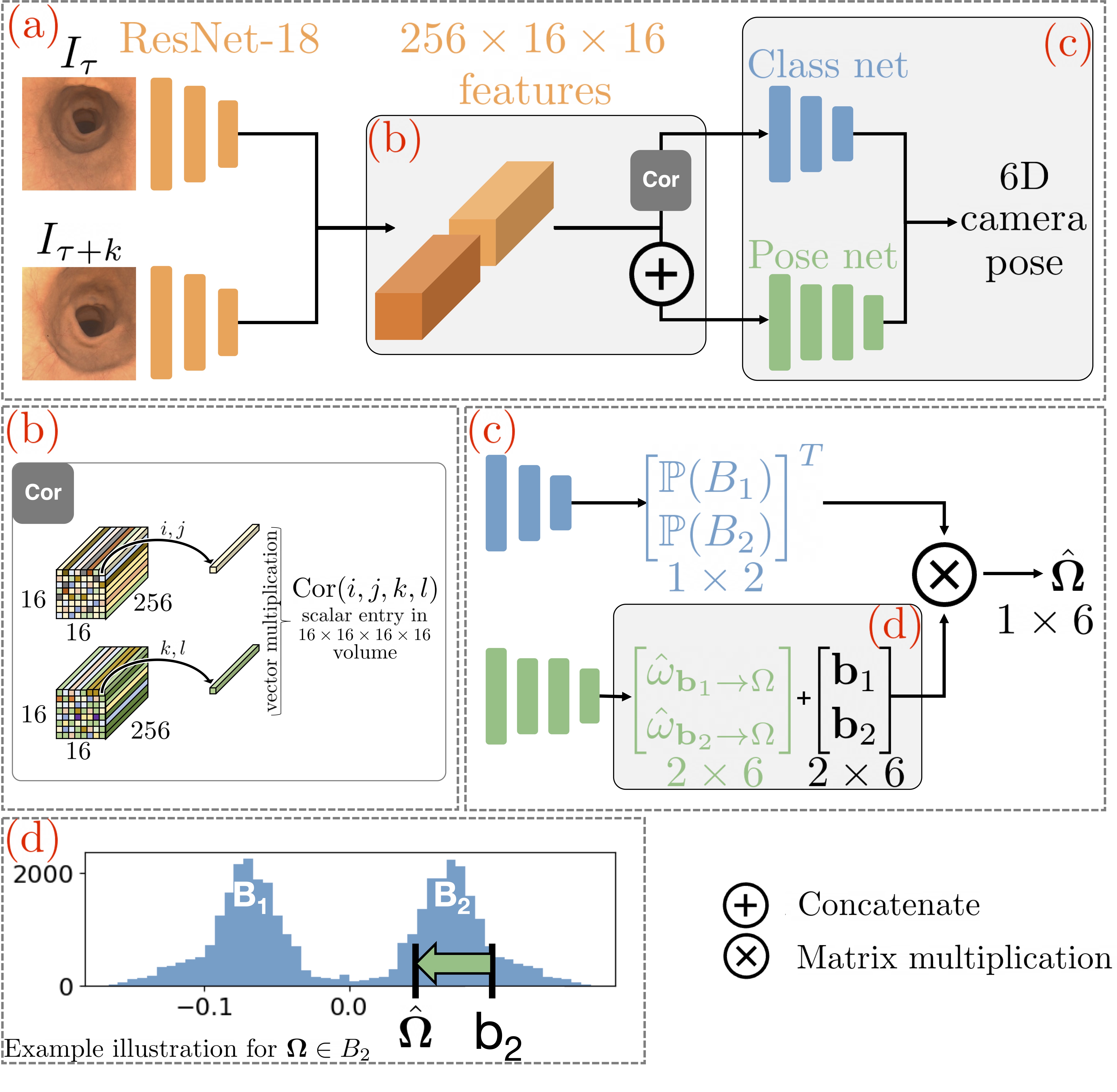}
\caption{Architecture of our bimodal model. A) Overview of the proposed approach. B) Illustration of the Correlation layer. C) Combining the outputs of Class Net and Pose Net. D) The output $\hat{\omega}_{\textbf{b}\rightarrow\mathbf{\Omega}}$ of Pose Net sets off the approximate bin centers. Illustration in 1D.}
	\label{fig:archi}
\end{figure}
Inspired by \cite{mahendran2018mixed}, we propose an explicitly bimodal model to learn bimodally distributed camera pose. Employing a supervised model classifying poses into insertion or withdrawal first, and offsetting downstream pose predictions accordingly, the model is forced to learn two modes instead of a single mode as observed in previous work. The required data for supervision was introduced in the previous section. This section describes our model. An overview of our approach can be found in Figure \ref{fig:archi}. We pass two images through a pre-trained ResNet-18 feature extractor and then through two branches: a correlation layer followed by a class net and a pose net.

\begin{figure}
    \centering
	\begin{minipage}{\linewidth}
		\centering
		\includegraphics[width=0.49\columnwidth]{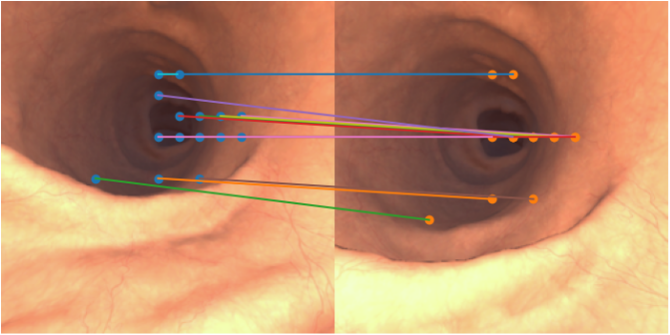}
        \includegraphics[width=0.49\columnwidth]{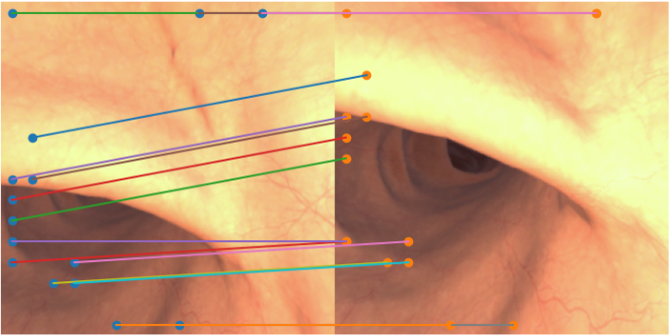}\\
        \small{Correlation layer (Cor)}
	\end{minipage}
	\begin{minipage}{\linewidth}
		\vspace{2mm}
		\centering
        \includegraphics[width=0.49\columnwidth]{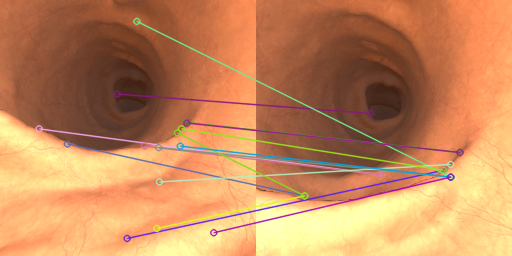}
        \includegraphics[width=0.49\columnwidth]{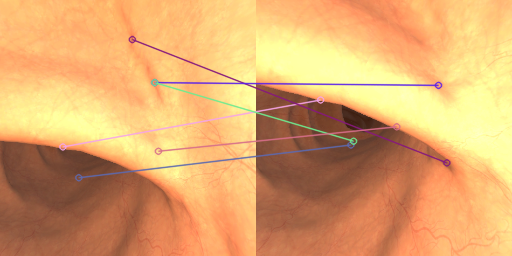} \\
        \small{SIFT features}
	\end{minipage}
	\begin{minipage}{\linewidth}
		\vspace{2mm}
		\centering
        \includegraphics[width=0.49\columnwidth]{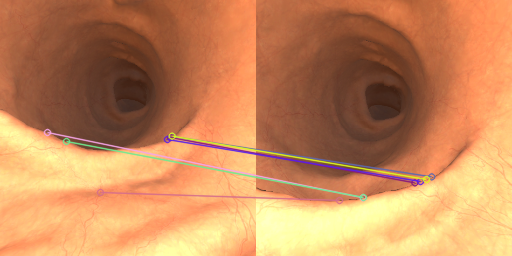}
        \includegraphics[width=0.49\columnwidth]{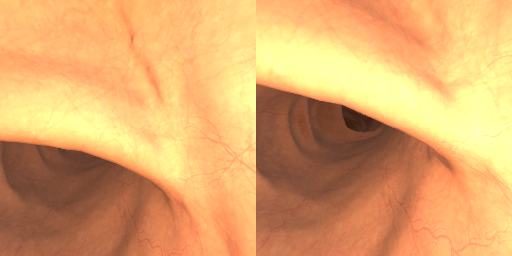} \\
        \small{ORB features}
	\end{minipage}

    \caption{Top: Visualization of the correlation layer used in this work. The argmax lets us illustrate the pixels with the highest correlation in the other image. Note that our network does not use explicit matches, but correlation volumes instead. Center: Features matches using SIFT features \cite{lowe2004distinctive} according to the standard OpenCV pipeline \cite{opencv}. Bottom: Features matches using ORB features \cite{rublee2011orb} according to the standard OpenCV pipeline \cite{opencv}. The correlation layer is more robust towards a lack of features.}
    \label{fig:corr}
\end{figure}

\noindent \textbf{Correlation layer:} 
As introduced in \cite{rocco2018neighbourhood} and used for pose regression in \cite{ding2020improving}, the down-sampled feature embeddings are passed through a layer in which the correlation between each feature pair is exhaustively computed. The output is a $16 \times 16 \times 16 \times 16$ correlation volume. The volume is flattened and normalized along the first two dimensions to indicate matches of $I_{\tau+k}$ in $I_{\tau}$, and along the last two dimensions for the opposite case. Unlike \cite{rocco2018neighbourhood,ding2020improving} our correlation layer has no learnable weights. Yet, it is crucial for the convergence of the classification net as it decouples the features for pose regression from the class net: The class net does not directly depend on the learned features but on their similarity across images. The correlation layer matches features between an image pair, and although they are too noisy to be useful for direct pose estimation, their signal helps distinguish between forward and backward movements easily. Employing the correlation layer, the classification accuracy of our model has reached 99\% on the validation set after only one epoch. For visualization only, we return the argmax along the feature dimension and receive a 16x16 grid with the location of the most similar feature in the other image in each pixel. We upsample the grid matches to the centers of their respective patches in the $475\times475$ image and show the 15 matches with the highest correlation in Figure \ref{fig:corr}. Experimenting with the argmax as input to the class net, we found the volume more useful. For comparison, Figure \ref{fig:corr} also shows two representative examples of hand-crafted features producing wrong, few, or no matches at all.

\noindent \textbf{Class net:} 
The output of the correlation layer is passed on to the class net that classifies relative poses into insertion $t_{z} > 0$ or withdrawal $t_{z} < 0$ movements, where $t_{z}$ denotes the translation along the $z$ axis based on the projection matrix  $\mathbf{\Omega}_\tau = {\textbf{P}}_{\tau}^{-1} {\textbf{P}}_{\tau+k}$. In other words, the translation along the $z$-axis is classified into two \textit{bins}. Our network first embeds each image of a pair $(I_{\tau},{I}_{\tau+k})$ into a $256\times16\times16$ feature vector using a ResNet-18 architecture referred to as $f$. The embedded images are then concatenated along the feature dimension and passed into both the classification net $\mathcal{C}$ and a pose regression net $\mathcal{R}$. The class net is a simple 3-layer fully connected network with dropout layers with high dropout probability (0.5) as proposed in \cite{mahendran2018mixed}. It outputs a 2D vector that is passed through a softmax layer and can be interpreted as the probability $\mathbb{P}$ that the input image is from bin ${B_1}$ or bin ${B_2}$, respectively:
\begin{align}
    \mathcal{C}(f(I_{\tau}), f(I_{\tau+k})) &= [\mathbb{P}(t_{{z}} < 0) \quad \mathbb{P}(t_{z} > 0)] \nonumber \\&= [\mathbb{P}(\mathbf{\Omega}_\tau \in B_1) \quad \mathbb{P}(\mathbf{\Omega}_\tau\in B_2)], \nonumber
\end{align}
where ${B_1}$ and ${B_2}$ represent two sets of relative poses ${P}$, such that ${P} \in B_1 \iff {P}^{-1} \in B_2$.
As we assume that each class is normally distributed we train the classification network with cross entropy loss
\begin{equation}
    L_{class} = - \sum_i \mathbb{P}(B_i) \log \hat{\mathbb{P}}(B_i).
\end{equation}

\noindent \textbf{Representation of relative pose:}
There are different ways to represent camera poses. As a rotation matrix has many almost zero entries, it is not a good representation for a network to learn. Therefore, we use the logarithms of unit quaternions in this work. They are represented as a 3D vector that can be mapped to the 4D unit quaternion as proposed in \cite{brahmbhatt2018geometry}. The pose error 
\begin{equation}
    L_{pose} = |\hat{{\textbf{t}}}-{\textbf{t}}|\exp^{-\beta} + \beta + |\log \hat{{\textbf{q}}} - \log {\textbf{q}}|\exp^{-\gamma} + \gamma
\end{equation}
is a weighted sum of rotation error and translation error, where the weights $\beta$ and $\gamma$ are learned as proposed in \cite{kendall2017geometric} and adapted to log quaternions in \cite{turkoglu2021visual} with initial values $0$ and $-3$, respectively. The final pose output of our network is a 6D vector representing the 3D translation and the 3D log quaternion. For convenience, we continue to refer to the relative pose in the new representation as $\mathbf{\Omega}$.

\noindent\textbf{Pose net: }
The pose net $\mathcal{R}$ is a simple 4-layer fully convolutional regression net with Relu activations as used in \cite{ozyoruk2021endoslam}. $\mathcal{R}$ outputs a 6D pose estimate for each of the two classes relative to each of the two bins as proposed in \cite{mahendran2018mixed}.
 Let \textit{bin1}, \textit{bin2} be the approximate centers of $B_1$ and $B_2$ describing the translation along the z-axis and $\textbf{b}_1 = [0, 0, bin1, 0, 0, 0]^T$. Then $\mathcal{R}$ predicts the difference between the ground truth pose in 6D representation and both $\textbf{b}_1$ and $\textbf{b}_2$. We denote the differences as $\hat{\mathbf{\omega}}_{\textbf{b}_1\rightarrow \mathbf{\Omega}}$ and $\hat{\mathbf{\omega}}_{\textbf{b}_2\rightarrow \mathbf{\Omega}}$. 
The pose net $\mathcal{R}$ outputs
\begin{equation}
\mathcal{R}(f(I_{\tau}), f(I_{\tau+k})) = [\hat{{\omega}}_{\textbf{b}_1 \rightarrow \mathbf{\Omega}_\tau} \quad \hat{\mathbf{\omega}}_{\textbf{b}_2 \rightarrow \mathbf{\Omega}_\tau}].
\end{equation}

\noindent \textbf{Final prediction:} 
The final pose prediction is a weighted sum of the probabilities from the class net and the regressed pose differences
\begin{align}
&\hat{\mathbf{\Omega}}_\tau= \begin{bmatrix} \mathbb{P}(\mathbf{\Omega}_\tau \in B_1) \\ \mathbb{P}(\mathbf{\Omega}_\tau \in B_2)\end{bmatrix}^T \cdot \begin{bmatrix} \textbf{b}_1 + \hat{{\omega}}_{\textbf{b}_1 \rightarrow \mathbf{\Omega}_\tau} \\  \textbf{b}_2 + \hat{{\omega}}_{\textbf{b}_2 \rightarrow \mathbf{\Omega}_\tau} \end{bmatrix}.
\end{align}

\noindent \textbf{Target function:}
Let $w_c$ be the weight for the class loss. Then our network is trained minimizing the target function 
\begin{equation}
    L = L_{pose} + w_c L_{class}.
\end{equation}

\section{Experiments}

In this section, we first observe the drawbacks of self-supervised pose networks. We then show that models trained on our data can better generalize to real colonoscopy sequences. Moreover, we show that supervised methods, too, can generalize to real data, making them suitable to be used for real procedures and lending themselves to domain adaptation methods. Lastly, we show that our novel bimodal method outperforms the unimodal baseline and is better suited to learning the camera movement during real colonoscopy. 

\subsection{Minima of self-supervised target functions are not guaranteed to minimize depth and pose errors}
\begin{figure}
	\begin{center}
  \includegraphics[width=0.04\columnwidth]{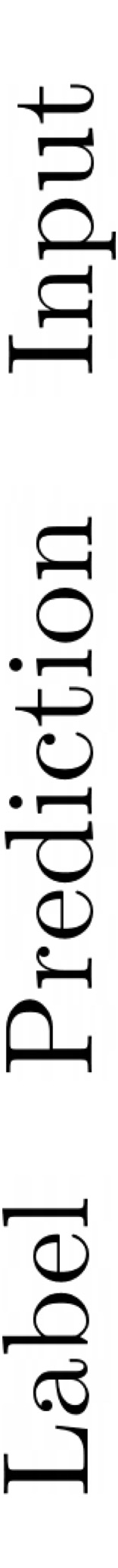}
		\includegraphics[width=0.174\columnwidth]{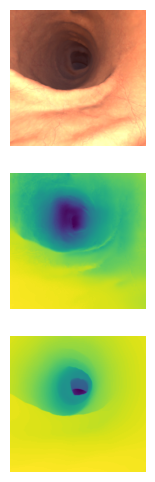}
		\includegraphics[width=0.174\columnwidth]{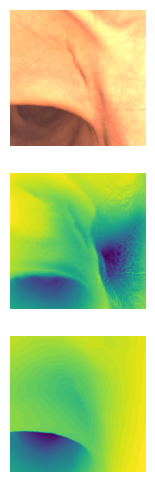}
		\includegraphics[width=0.174\columnwidth]{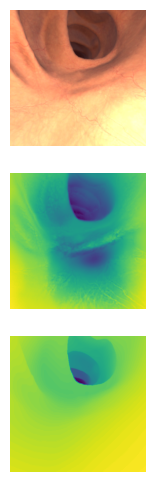}
		\includegraphics[width=0.174\columnwidth]{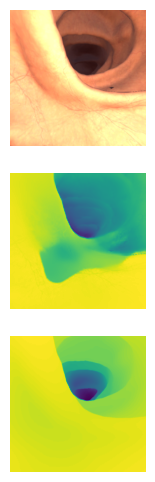}
		\includegraphics[width=0.174\columnwidth]{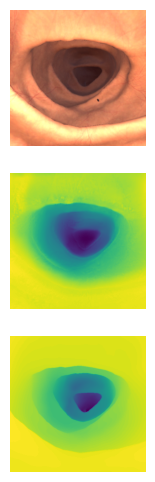}
	\end{center}
	\caption{Examples for predicted and ground truth (GT) depth maps using the self-supervised baseline. Scales are omitted as depths are predicted up to scale. Blue corresponds to a higher depth.}
	\label{fig:depths}
\end{figure}
\begin{figure}
	\begin{minipage}{0.24\textwidth}
		\centering
        \includegraphics[width=\columnwidth]{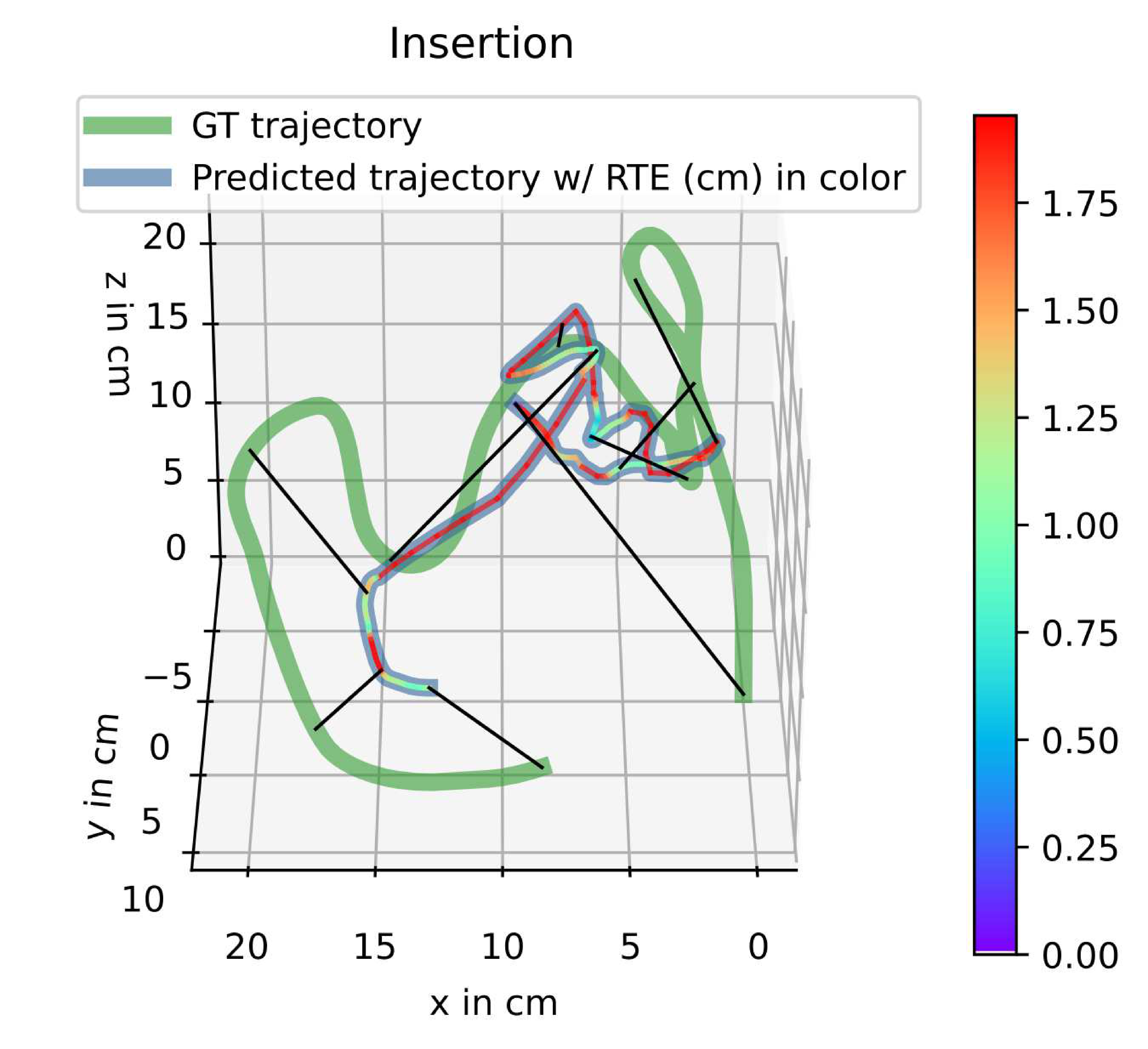}
	\end{minipage}
	\begin{minipage}{0.24\textwidth}
		\centering
        \includegraphics[width=\columnwidth]{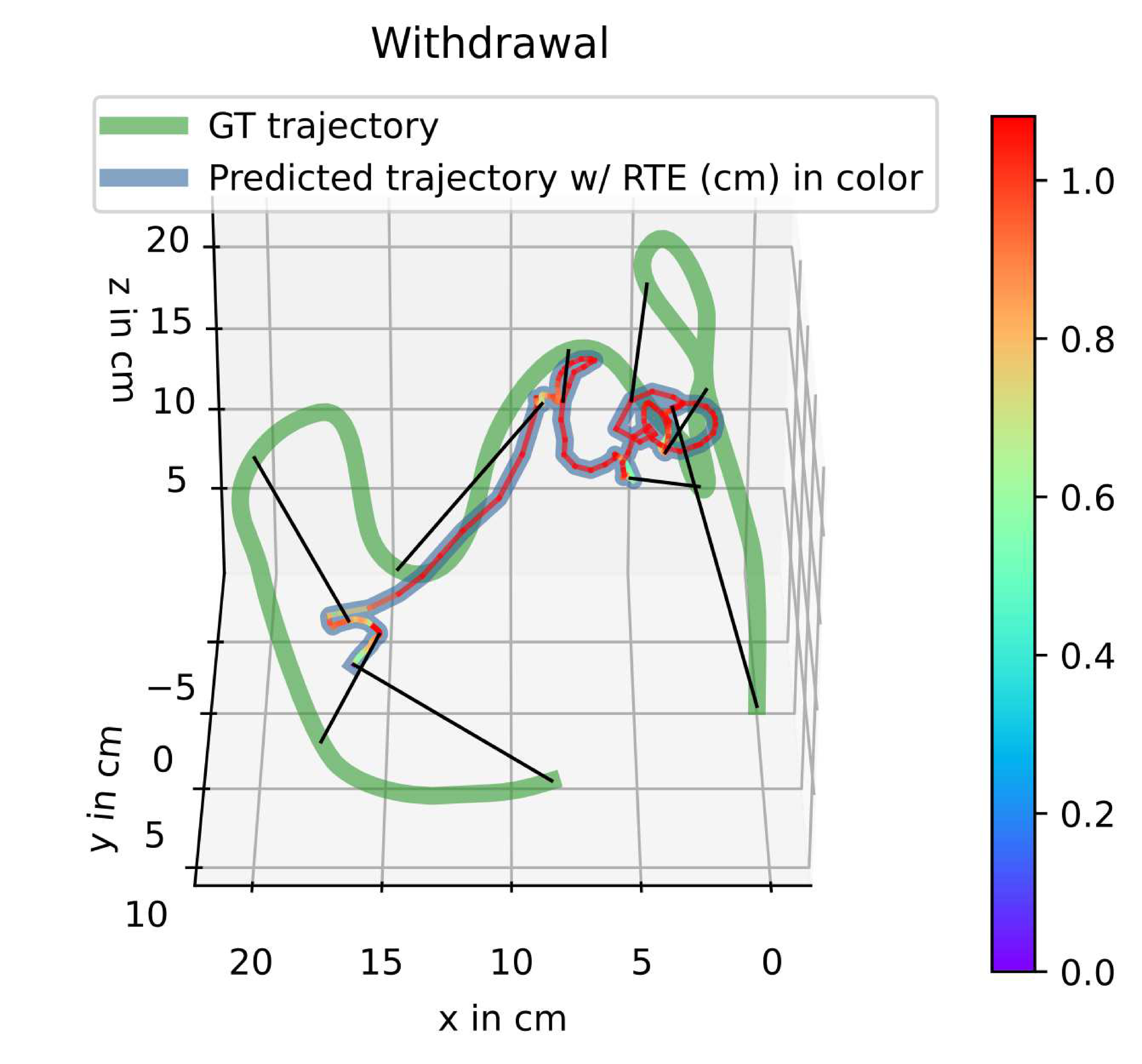}
	\end{minipage}
	\caption{Predicted camera trajectories using the self-supervised baseline are more accurate during insertion than during withdrawal. RTE is color coded in the predicted trajectory. Black links connect same time steps for a few examples. We aligned the predicted trajectories to the ground truth with an affine transformation for better visualization.}
	\label{fig:unsupertraj}
\end{figure}
We train the method proposed and published in \cite{ozyoruk2021endoslam} on our own data with all hyper-parameters set to their default values. To obtain the full trajectory of absolute poses, we multiply the relative poses. The pose of a camera $\tau$ in world space can be computed as {$\textbf{P}_\tau = \textbf{{P}}_1 \mathbf{\Omega}_{1} \cdot ... \cdot \mathbf{\Omega}_{\tau-1}$}, where each $\mathbf{\Omega}$ projects the initial pose $\textbf{P}_1$ sequentially to the subsequent camera pose. 
\begin{figure*}
\centering
	\includegraphics[width=0.32\textwidth]{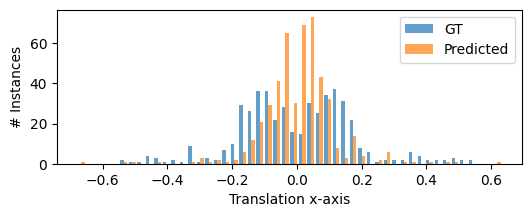}
	\includegraphics[width=0.32\textwidth]{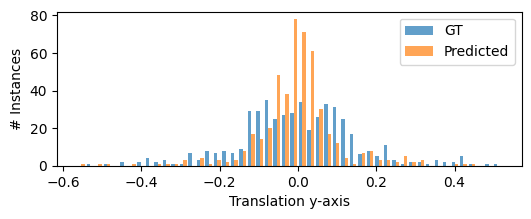}
	\includegraphics[width=0.32\textwidth]{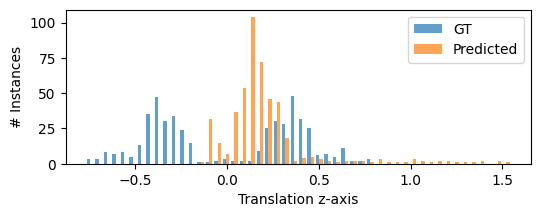}

	\caption{Minimizing standard self-supervised losses is not sufficient to learn a bimodal distribution. Histograms of predicted vs. ground truth (GT) poses on one sequence of our dataset (test set 1). Translations in cm. }
	\label{fig:comp_histos}
\end{figure*}
\begin{table*}
	\caption{Absolute translation errors,  Relative translation errors, and Rotation errors on our test trajectories}
	\setlength{\tabcolsep}{3pt}
	\begin{center}
	{\def\arraystretch{1.2}
	\begin{tabular}{|l|cccc|cccc|cccc|}
		\hline
		Test trajectory & \multicolumn{4}{c|}{1}&\multicolumn{4}{c|}{2}&\multicolumn{4}{c|}{3} \\
		Total trajectory length & \multicolumn{4}{c|}{105.1 cm}&\multicolumn{4}{c|}{103.6 cm}&\multicolumn{4}{c|}{102.2 cm} \\
		Mean step size & \multicolumn{4}{c|}{4.4 mm}&\multicolumn{4}{c|}{4.3 mm}&\multicolumn{4}{c|}{4.3 mm} \\
		Mean rotation per step & \multicolumn{4}{c|}{4.6°}&\multicolumn{4}{c|}{4.8°}&\multicolumn{4}{c|}{4.8°} \\
		\hline
		\hline
		&ATE(cm)&RTE(mm)&ROT(°)&Acc(\%)&ATE(cm)&RTE(mm)&ROT(°)&Acc(\%)&ATE(cm)&RTE(mm)&ROT(°)&Acc(\%) \\
		\hline
				COLMAP \cite{schoenberger2016sfm} &0.01 & 0.07 & 0.18& 8  &0.02 & 0.10 & 0.28 & 6 &0.37 & 0.68 & 1.42 & 7 \\
    \hline
		EndoSLAM \cite{ozyoruk2021endoslam}&14.9/17.8 & 3.10/4.26 & 1.3/5.7 & 100/41  &18.0/15.3 & 3.50/4.28 & 1.8/5.0 & 100/35 &14.9/15.7 & 3.24/4.16 & 1.6/5.5 & 100/37\\
EndoSLAM \cite{ozyoruk2021endoslam} bimodal w/$L_{c}$&12.4/19.1&2.44/4.23&1.5/5.8&100/41&18.8/11.6&2.80/5.76&1.9/5.8&100/37&19.0/14.4&2.94/4.90&1.7/5.1&100/42\\
		
		\hline
		
		Ours unimodal &7.08/12.5 & 0.75/0.79 &\textbf{1.3/1.3 }& 100/99 &\textbf{2.79/2.63}& 0.76/0.76 & \textbf{1.5/1.5 }&100/100 &7.56/10.0& 0.87/0.90 & 1.5/1.6 & 100/99 \\
		Ours bimodal &8.86/13.3 & 0.72/0.72&1.5/1.5 & 100/99 &2.83/6.03& 0.69/0.71 & 1.7/1.7&100/100 &6.17/9.62& 0.85/0.89 & 1.6/1.6 & 99/99 \\
		\textbf{Ours bimodal w/$L_{c}$} &\textbf{8.81/9.79} &\textbf{0.69/0.72}&1.5/1.5 & 100/99 &2.35/5.23& \textbf{0.67/0.70} & 1.6/1.6 &100/100 &\textbf{3.78/9.00}& \textbf{0.82/0.85} & 1.6/1.5 & 100/99 \\
				{Ours bimodal w/$L_{c}$ w/o Corr} &14.0/13.9 &{0.77/0.76}&1.7/1.6 & 99/99 &8.32/5.46&{0.73/0.71} & 2.1/2.0 &99/98 &{8.70/8.10}& {0.93/0.98} & 1.9/1.8 & 99/98 \\
		\hline
	\end{tabular}}\\ $ $ \\
	\end{center}
	ATE \& RTE, ROTation error, and accuracy for our test trajectories' forward/backward traversal. COLMAP performs global optimization and thus direction does not matter (only one value reported). The accuracy for regression methods (Ours and EndoSLAM) denotes the percentage of correctly predicted directions. The accuracy for COLMAP reports the percentage of frames from the trajectory that COLMAP was able to reconstruct. Note that although COLMAP yields the smallest errors, the method successfully reconstructs only a small fraction of each trajectory. Bold indicates the best result as the sum of forward and backward errors. W/o Corr refers to our model without the correlation layer in the classfication net.
	\label{tab:results}
\end{table*}
As the network predicts pose up to scale we measure the accuracy of the \textit{scaled} trajectory using the Absolute Translation Error (ATE), the Relative Translation Error (RTE), and the ROTation error (ROT). The losses are defined as
\begin{align}
    RTE &= \mu_{\tau}(||trans({{\mathbf{\Omega}}_\tau}^{-1}\hat{\mathbf{\Omega}}_\tau)||), \label{eq:rte}\\
    ATE &= \mu_{\tau}(||trans(\textbf{P}_{\tau}) - trans(\hat{\textbf{P}}_{\tau})||)\label{eq:ate}, \quad \text{and}\\
    ROT &= \mu_{\tau}(\frac{trace(Rot(\mathbf{\Omega}_\tau^{-1}\hat{\mathbf{\Omega}}_\tau))-1}{2} \cdot \frac{180}{\pi})\label{eq:rot}, 
\end{align}
where $\mu_{\tau}$ denotes the median over all steps $\tau$, $trans$ and $Rot$ denote the translation and rotation components of a projection matrix, and $||.||$ denotes the 2-norm. The ATE measures drift and the overall consistency of the predicted trajectory; however, it is prone to outliers. More robust than the ATE is the RTE. It measures the magnitude of the difference between the predicted and actual relative pose per step and reflects translation and rotation errors on a local level. As we do not apply global optimization or loop-closure, and all methods in this work predict each relative pose independently, the RTE has more significance for us. Lastly, the ROT measures only the magnitude of the rotation of the local errors.

The scaling factor is defined as 
\begin{align}
    s = \dfrac{\sum_{\tau} trans(\textbf{P}_{\tau})^T \cdot trans(\hat{\textbf{P}}_{\tau})}{\sum_{\tau}  trans(\hat{\textbf{P}}_{\tau})^T \cdot trans(\hat{\textbf{P}}_{\tau})  }, \label{eq:scale}
\end{align}
where $\textbf{P}_{\tau}$ denotes absolute camera poses. A different approach would be to compute an alignment projection for the prediction, as in \cite{sturm2012benchmark}. However, as we investigate the ability to predict correct directions, we only scale the predicted trajectory but refrain from translating or rotating it. If a network were to predict the exact opposite of the ground truth, a rotation of 180° could compensate for the error, which is not informative in our case.
We compute independent scales for the forward and backward trajectories.  

In Figure \ref{fig:depths}, we show that the network has learned to predict smooth yet sensible depth maps. Because the network is trained based on the self-supervision of warped RGB and depth images, one might assume that the network must have learned both depth and camera pose. Investigating the resulting trajectory as shown in Figure \ref{fig:unsupertraj}, we find that the network predicts the insertion movement considerably better than the opposite movement. 
The forward movement successfully predicts the three half-loops, though with drift. Predicting the motion in the opposite direction does not lead to a sensible trajectory. Further evidence that the performance of the algorithm varies depending on direction is presented in Figure \ref{fig:comp_histos}. The histogram of predicted z-values is not aligned with the ground truth values leading to a bias towards a positive translation along the z-axis. Finally, also Table \ref{tab:results} (EndoSLAM) underlines our findings, where we quantitatively evaluate the performance of the self-supervised approach on three different trajectories within our dataset. Most notably, the RTE and rotation errors are higher for the withdrawal than for the insertion for all trajectories. Even applying our bimodal architecture with class loss to the pose branch of EndoSLAM does not enable the self-supervised model to learn the bimodal distribution. We assumed a hypothetical setting in which class labels (insertion vs. withdrawal) are available during training, but depth and pose ground truths are not. As reported in Table \ref{tab:results} (EndoSLAM bimodal w/$L_c$), the bimodal architecture and the class loss cannot force the model to learn a multi-modal distribution. The pose network simply learns to compensate for the enforced bin offsets, such that output remains unimodaly distributed. 
\begin{figure*}
	\centering
 	\includegraphics[width=\textwidth]{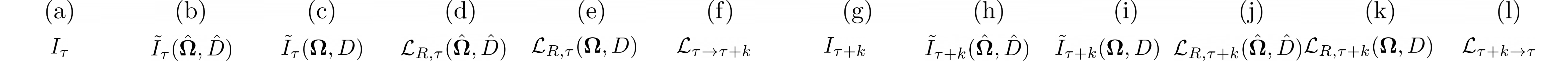}

	\includegraphics[width=\textwidth]{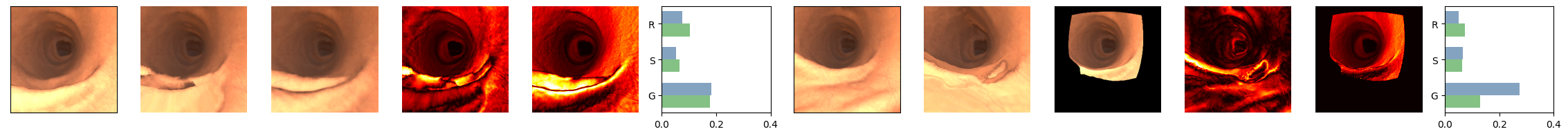}
	\includegraphics[width=\textwidth]{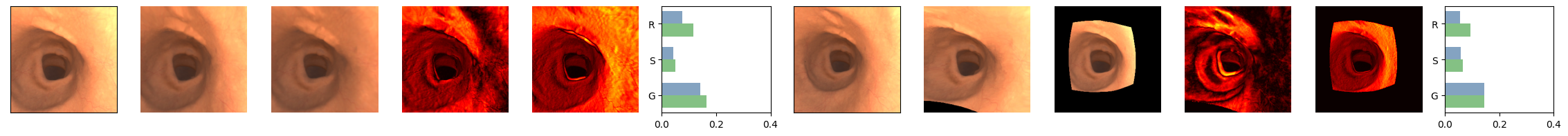}
	\includegraphics[width=\textwidth]{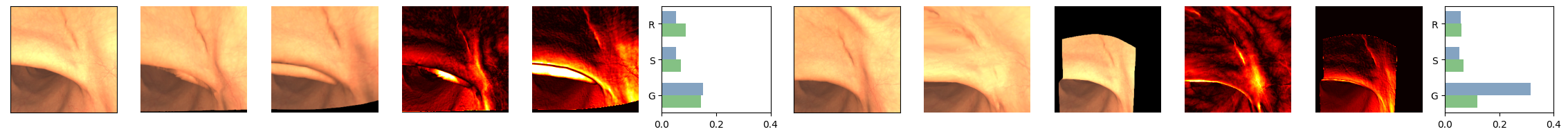}
	\includegraphics[width=\textwidth]{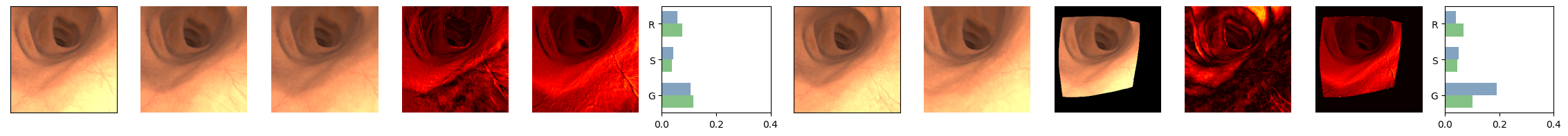}
	\includegraphics[width=\textwidth]{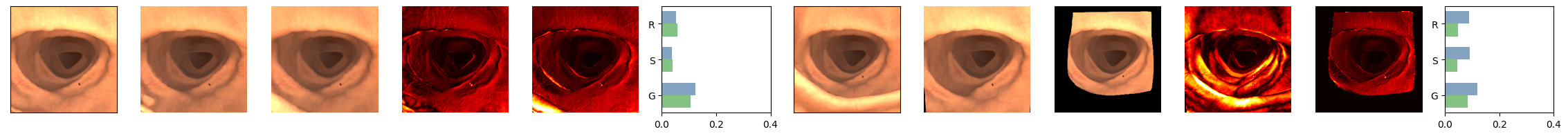}
	\footnotesize R=Reprojection loss, S=SSIM loss, G=geometric consistency loss.
	\caption{\color{black} Training losses of a self-supervised depth and pose network are not necessarily minimized at ground truth. Each row shows an image pair ($I_{\tau}$, $I_{\tau+k}$) and its induced training losses. The first six columns visualize the errors of warping $I_{\tau+k}$ to $I_{\tau}$, while the last six columns show the opposite case: (a) + (g) original images; (b), (c), (h), (i) warped images $\tilde{I}_{\tau+k\rightarrow \tau}$ and $\tilde{I}_{\tau\rightarrow \tau+k}$ using predicted depth and camera pose vs. using ground truth depth and pose; (d), (e), (j), (k) error maps for the Reprojection loss $L_R$, where dark areas represent low errors; and (f), (l) $L_R, L_S$ and $L_G$ for the \color{newblue}predicted \color{black}  and \color{newgreen}ground truth \color{black} inputs. If depth and pose were correctly learned we would expect (b) to equal (c), and (h) to equal (i). However, especially (h) fails to represent the correct camera movement.  Further, we would expect that using ground truth depth and camera pose leads to lower warping errors in (f) and (l) than using the predictions; yet, the errors induced by the ground truth are often higher (green bars larger than corresponding blue bars). In all examples the camera moves in negative z-direction between frames $\tau$ and $\tau+k$. \color{black}}
	\label{fig:lossesgtvspred}
\end{figure*}

\noindent\textbf{Dissecting insertion vs. withdrawal:}
We use the terms insert and withdraw when speaking of direction relative to the structure rather than forward/backward because the latter can be defined arbitrarily. So what is the difference between inserting and withdrawal movements? 
During training, the network predicts the forward and backward movement between an image pair $(I_{\tau}, I_{\tau+k})$. We show warped images in both directions based on ground truth depth and pose vs. warped images based on predicted (\^{}) depth and pose in Figure \ref{fig:lossesgtvspred}. We also show error maps and the resulting training losses.
Observe that column (h) is minimized during training to look like column (g). Moreover, column (h) should look like column (i) if the predicted depth and pose were correct. Note that even when using the ground truth values, the training losses are non-zero. Visually, there are significant differences between the ground truth warp and the predicted warp, especially along steps in the structure. Investigating the resulting errors in column (l) shows high errors for geometry consistency losses (D). 
 For the opposite movement, the images in columns (b) and (c) should look like column (a). But the ground truth warped images have many artifacts that originate in the self-occlusion of the scene and the sharp steps in depth. These artifacts lead to high warping errors for the ground truth, as shown in columns (k). The predicted warpings visually replicate nearby structures well, resulting in small reprojection errors, although wrong poses are predicted. Every example in column (f) has a higher reprojection error for the ground truth (green) than for the prediction (blue). The network has converged to a wrong optimum and predicts higher reprojection losses for the ground truth than for the predictions in 63\% of the test images.
 
We conclude that, due to the geometry of the colon, depth prediction and pose prediction do not necessarily improve or help each other. The different properties of the insertion and the withdrawal warping lead to different performances for the pose prediction. As a result, the unsupervised network learns a unimodal distribution of the z-translation shifted towards positive z-translations. The model in this case was not able to predict both forward and backward movements of the camera.
Because these are the main relative movements to be expected during colonoscopy, the bimodality needs to be accounted for. 

\subsection{Our data helps generalize to real data}
\begin{table*}
\caption{Robustness and generlizability to unseen real data}
\setlength{\tabcolsep}{3pt}
\begin{center}
{\def\arraystretch{1.2}
\begin{tabular}{|l|ccc|ccc|ccc|ccc|ccc|ccc|}
\hline

 & \multicolumn{3}{c|}{(a)} & \multicolumn{3}{c|}{(b)}& \multicolumn{3}{c|}{(c)}  & \multicolumn{3}{c|}{(d)} & \multicolumn{3}{c|}{(e)}& \multicolumn{3}{c|}{(f)} \\
\hline
\hline
Total trajectory length* &\multicolumn{3}{c|}{12.0}&\multicolumn{3}{c|}{14.6}&\multicolumn{3}{c|}{11.6}&\multicolumn{3}{c|}{11.9}&\multicolumn{3}{c|}{11.1}&\multicolumn{3}{c|}{15.0}\\
Mean step size*&\multicolumn{3}{c|}{0.7}&\multicolumn{3}{c|}{0.4}&\multicolumn{3}{c|}{0.9}&\multicolumn{3}{c|}{0.7}&\multicolumn{3}{c|}{0.6}&\multicolumn{3}{c|}{0.8}\\
Mean rotation per step in ° &\multicolumn{3}{c|}{2.1}&\multicolumn{3}{c|}{1.5}&\multicolumn{3}{c|}{2.1}&\multicolumn{3}{c|}{1.6}&\multicolumn{3}{c|}{2.3}&\multicolumn{3}{c|}{2.0}\\
\hline
\hline
ATE $\rightarrow$ &
\color{orange}\underline{1.2}& \color{red}{{3.5}}& \color{ForestGreen}7.9 &  \color{orange}\underline{1.4}& \color{red}{{6.2}}& \color{ForestGreen}8.4 & \color{orange}\underline{1.3}&\color{red}{{1.8}}&\color{ForestGreen}11. &
\color{orange}\underline{1.6}&\color{red}{\underline{1.6}}&\color{ForestGreen}5.6&
\color{orange}2.6& \color{red}{\underline{2.3}}&\color{ForestGreen}3.2&
\color{orange}\underline{1.5}&\color{red}{{6.5}}&\color{ForestGreen}7.5\\
ATE $\leftarrow$&
\color{orange}\underline{1.0}& \color{red}{3.2}&\color{ForestGreen} {2.3}&  \color{orange}\underline{1.2}&\color{red} {1.8}& \color{ForestGreen}{3.6} & \color{orange}\underline{0.7}&\color{red} {1.2}& \color{ForestGreen}{2.1} &
\color{orange}\underline{1.1}&\color{red} {6.6}& \color{ForestGreen}{6.5}& \color{orange}\underline{1.1}&\color{red} {11.}& \color{ForestGreen}{9.6} &
\color{orange}\underline{1.1}& \color{red}6.5&\color{ForestGreen} {{6.3}}\\
\hline
RTE $\rightarrow$&
\color{orange}\underline{0.3}& \color{red}{0.4}& \color{ForestGreen}{1.0}&
 \color{orange}\underline{0.2}&\color{red} {0.3}& \color{ForestGreen}{0.6 }& \color{orange}\underline{0.2}&\color{red} {0.3}& \color{ForestGreen}{1.7} &
 \color{orange}\underline{0.2}&\color{red} {0.3}& \color{ForestGreen}{0.5 }& \color{orange}\underline{0.3}&\color{red} {0.4}& \color{ForestGreen}{0.5 }&
\color{orange}\underline{0.2}& \color{red}{{0.5}}& \color{ForestGreen}0.7\\
RTE $\leftarrow$&
\color{orange}\underline{0.3}&\color{red} {{0.4}}& \color{ForestGreen}0.5&
 \color{orange}0.2&\color{red} \underline{0.1}& \color{ForestGreen}{0.2}& \color{orange}\underline{0.3}&\color{red} \underline{0.3}& \color{ForestGreen}{0.6}& 
 \color{orange}\underline{0.3}&\color{red} {0.5}&\color{ForestGreen} 1.1 & \color{orange}\underline{0.2}&\color{red} {1.0}&\color{ForestGreen}{1.0}&
 \color{orange}\underline{0.3}&\color{red} {0.4}& \color{ForestGreen}{0.5}\\
\hline
ROT $\rightarrow$&
\color{orange}2.0& \color{red}{\underline{1.7}}&\color{ForestGreen} 2.0&
\color{orange}1.2& \color{red}\underline{1.0}&\color{ForestGreen} {1.8} &
\color{orange}3.2& \color{red}\underline{1.9}&\color{ForestGreen} {2.4} &
\color{orange}\underline{1.0}& \color{red}{1.5}&\color{ForestGreen} {1.4}&
\color{orange}2.9& \color{red}{2.4}&\color{ForestGreen} \underline{2.0}&
\color{orange}\underline{1.1}& \color{red}{1.8}& \color{ForestGreen}{1.2}\\
ROT $\leftarrow$&
\color{orange}2.0& \color{red}{\underline{0.9}}&\color{ForestGreen} 2.7&
\color{orange}1.1& \color{red}\underline{1.0}&\color{ForestGreen} {1.4}&
\color{orange}3.1& \color{red}\underline{1.9}&\color{ForestGreen} {2.4}&
\color{orange}\underline{1.0}& \color{red}{2.8}& \color{ForestGreen}{1.9}&
\color{orange}3.2& \color{red}\underline{2.5}&\color{ForestGreen} {2.6}&
\color{orange}1.3& \color{red}{1.5}& \color{ForestGreen}\underline{1.1}\\
\hline

\end{tabular}}\\ $ $ \\
\footnotesize{Best method for each sequence and measure underlined.}
\end{center}
Comparison of three methods evaluated on six real sequence: (i) \color{orange} Our proposed method trained on our proposed dataset\color{black}, (ii) \color{red}EndoSLAM method trained on our dataset \textit{SimCol}\color{black}, and (iii) \color{ForestGreen} EndoSLAM method trained on EndoSLAM data \color{black}. ATE \& RTE in unknown scale and Rotation error in degrees for the forward ($\rightarrow$) and backward ($\leftarrow$) traversal. Errors relative to COLMAP labels.
* Absolute scale unknown.
\label{tab:real}
\end{table*}
\begin{figure*}
\centering
 \begin{minipage}{0.16\textwidth}
 \centering{\footnotesize{(a)}}\\
     \includegraphics[width=1.2\linewidth]{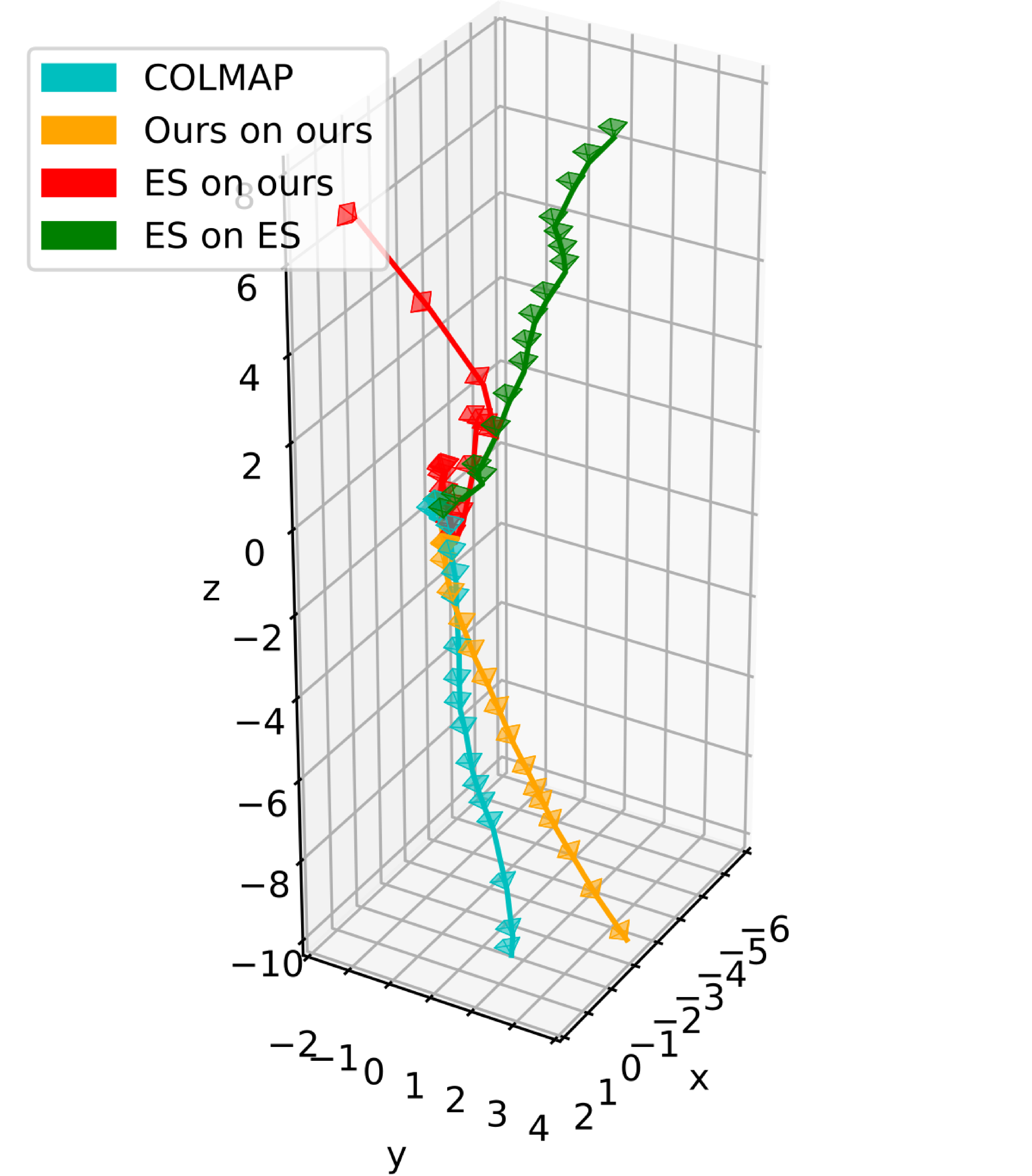} \end{minipage}
  \begin{minipage}{0.16\textwidth}
  \centering{\footnotesize{(b)}}\\
     \includegraphics[width=1.2\linewidth]{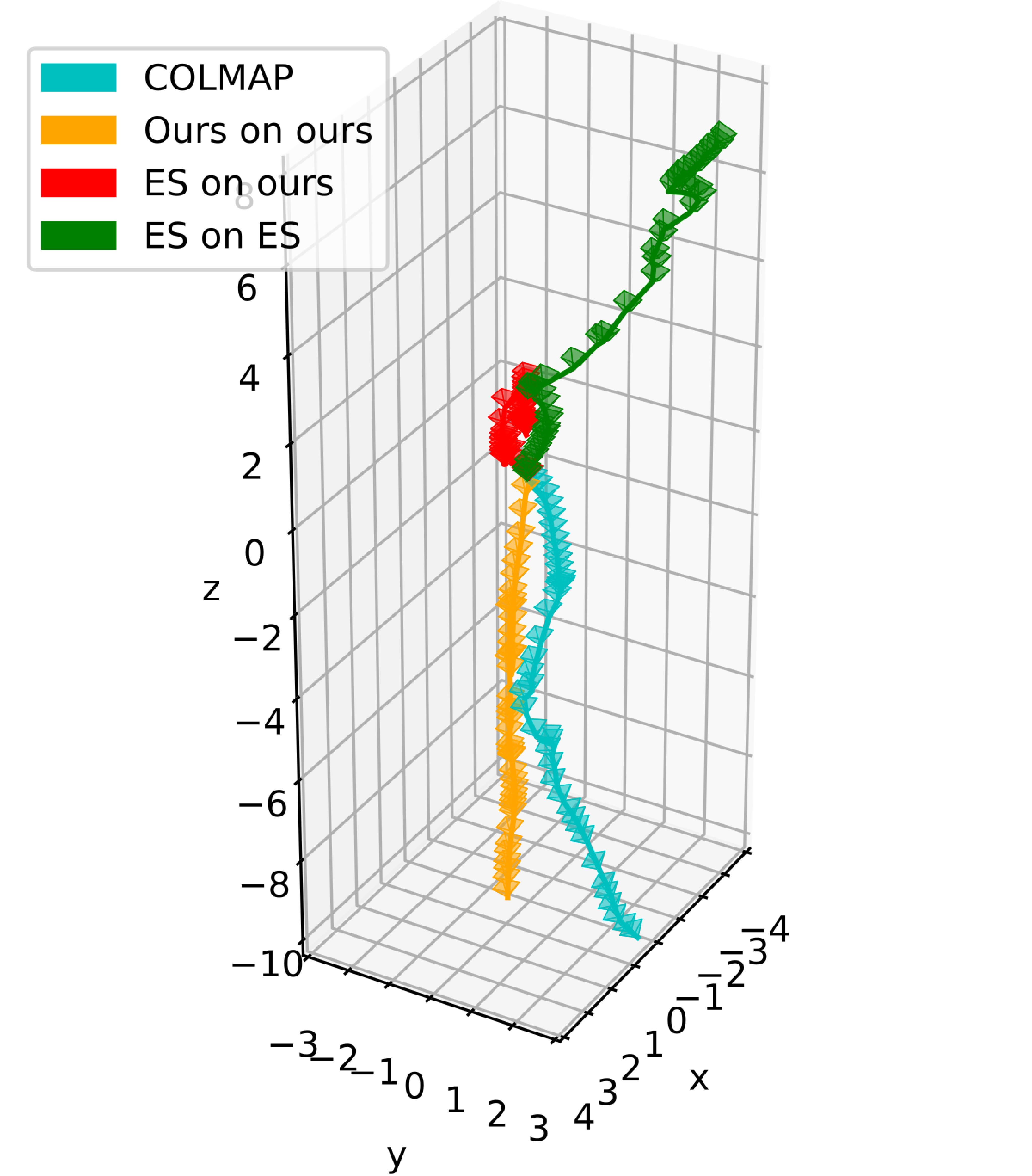}
 \end{minipage}
  \begin{minipage}{0.16\textwidth}
  \centering{\footnotesize{(c)}}\\
     \includegraphics[width=1.2\linewidth]{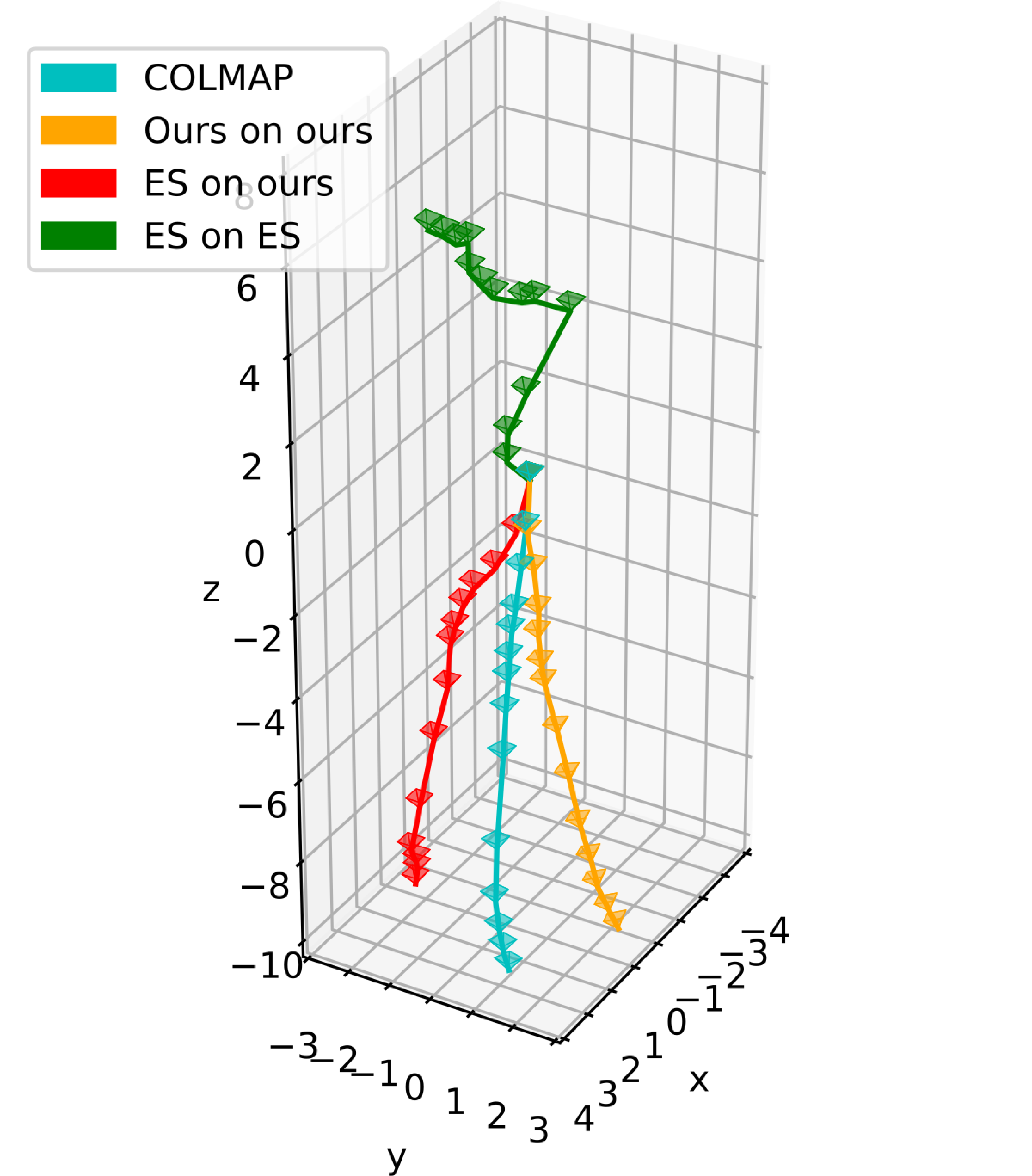}
 \end{minipage}
  \begin{minipage}{0.16\textwidth}
  \centering{\footnotesize{(d)}}\\
     \includegraphics[width=1.2\linewidth]{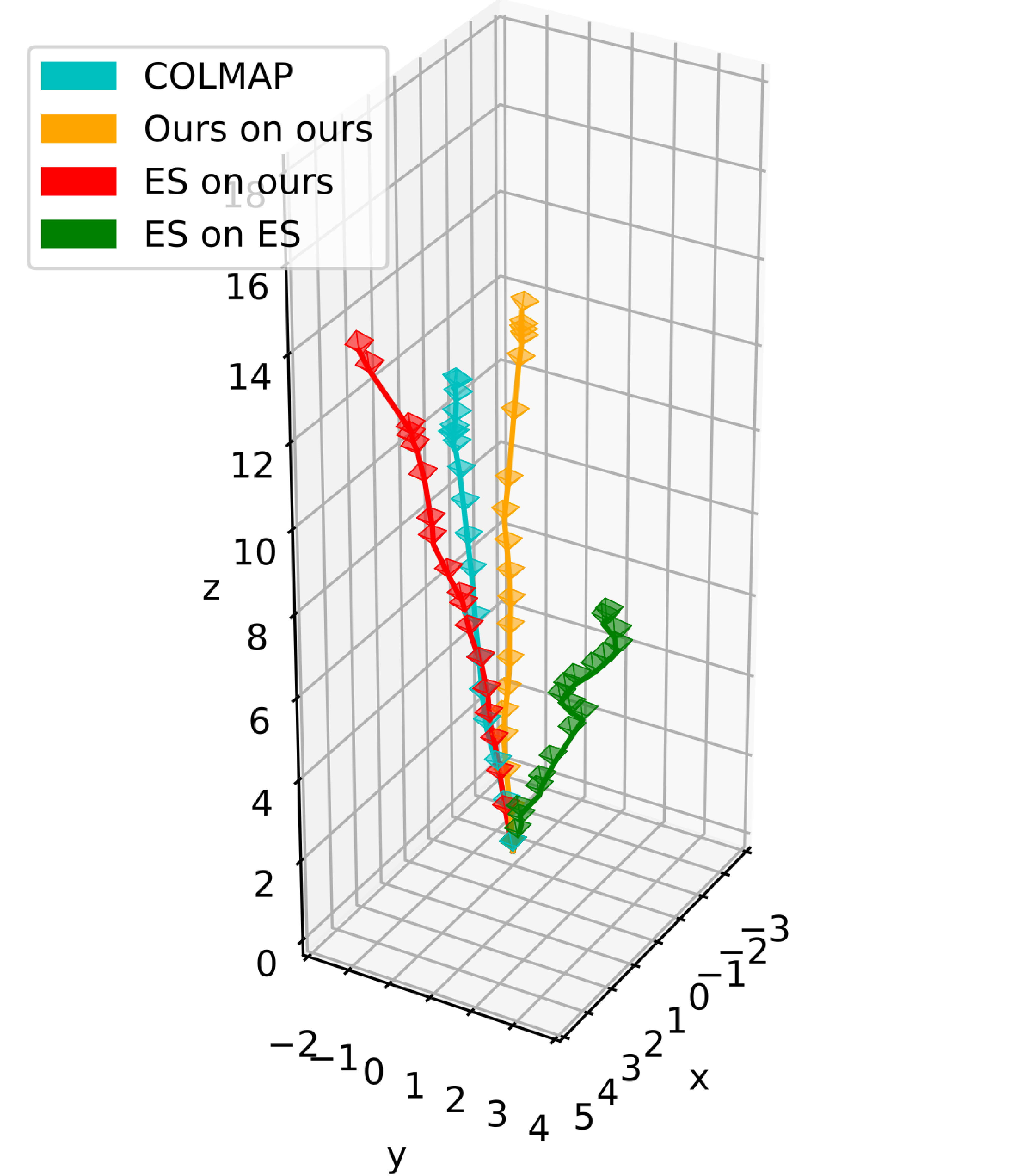}
 \end{minipage}
  \begin{minipage}{0.16\textwidth}
  \centering{\footnotesize{(e)}}\\
     \includegraphics[width=1.2\linewidth]{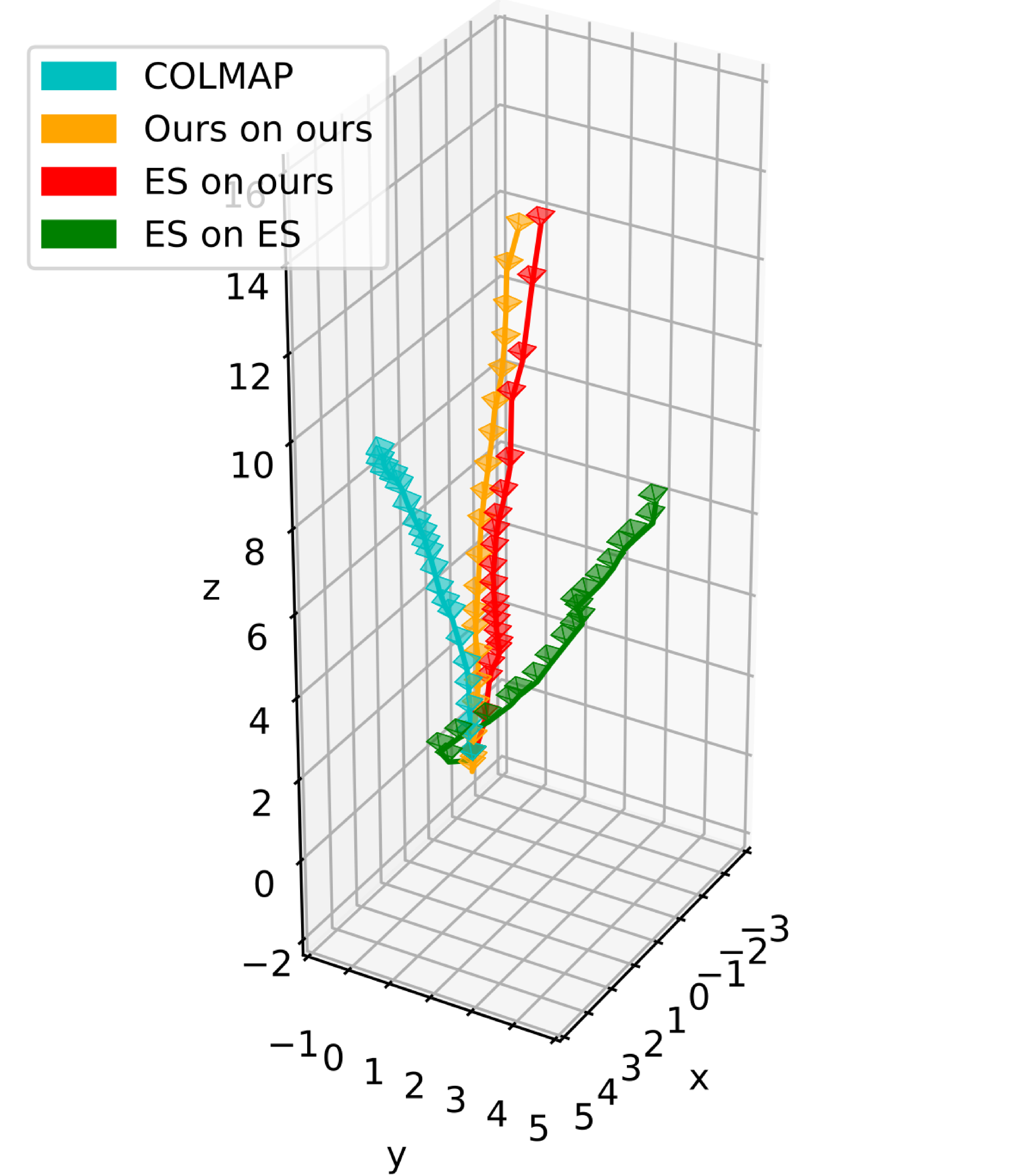}
 \end{minipage}
  \begin{minipage}{0.16\textwidth}
  \centering{\footnotesize{(f)}}\\
     \includegraphics[width=1.2\linewidth]{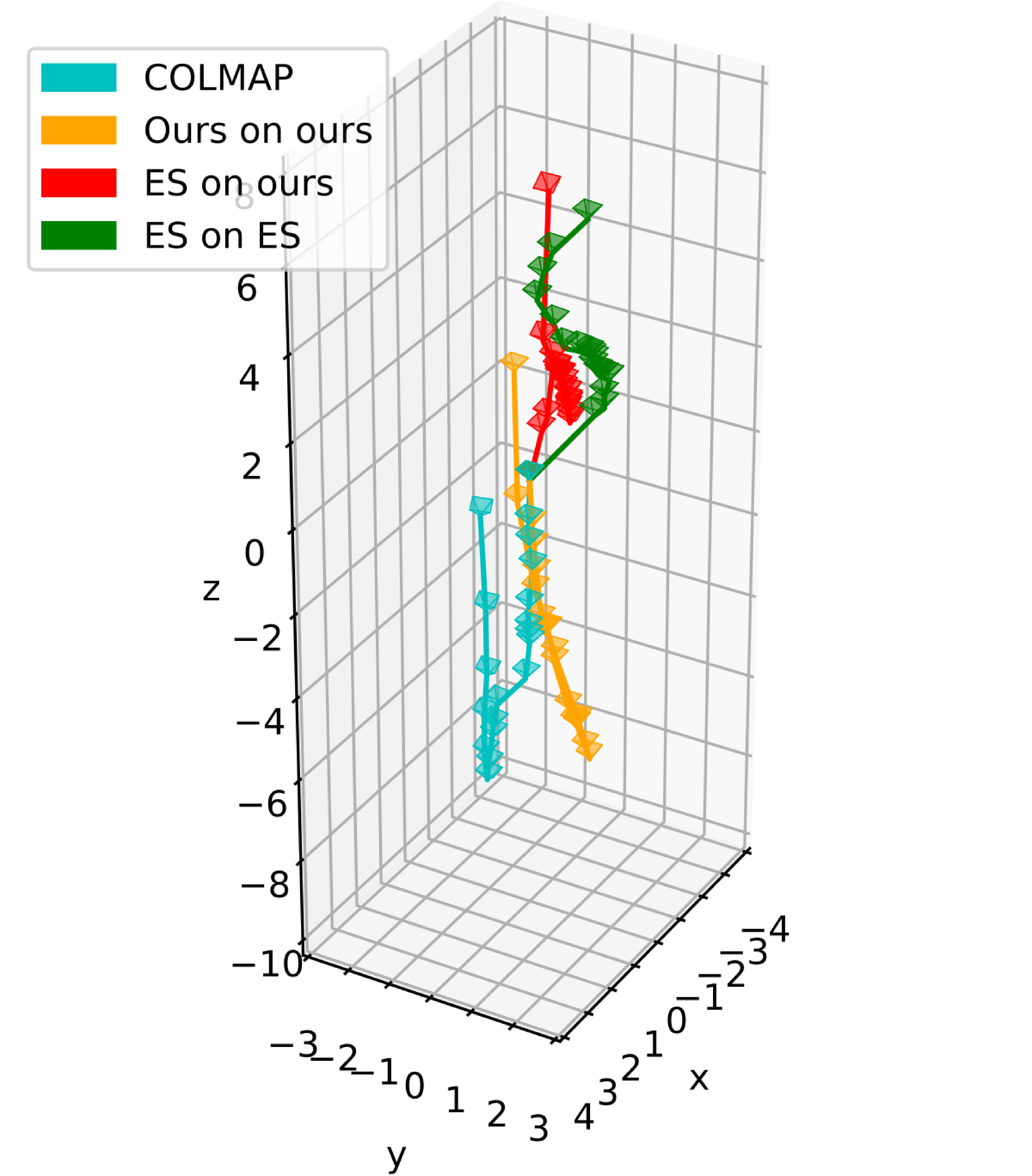}
 \end{minipage} 
 \\$ $ \\[0.1cm]
    \setlength{\tabcolsep}{0pt}
    {\renewcommand{\arraystretch}{1.2}
    \begin{tabular}{lcccccccccc}
    	\footnotesize{(a) \hspace{1pt}}
		&\includegraphics[align=c,width=0.09\linewidth]{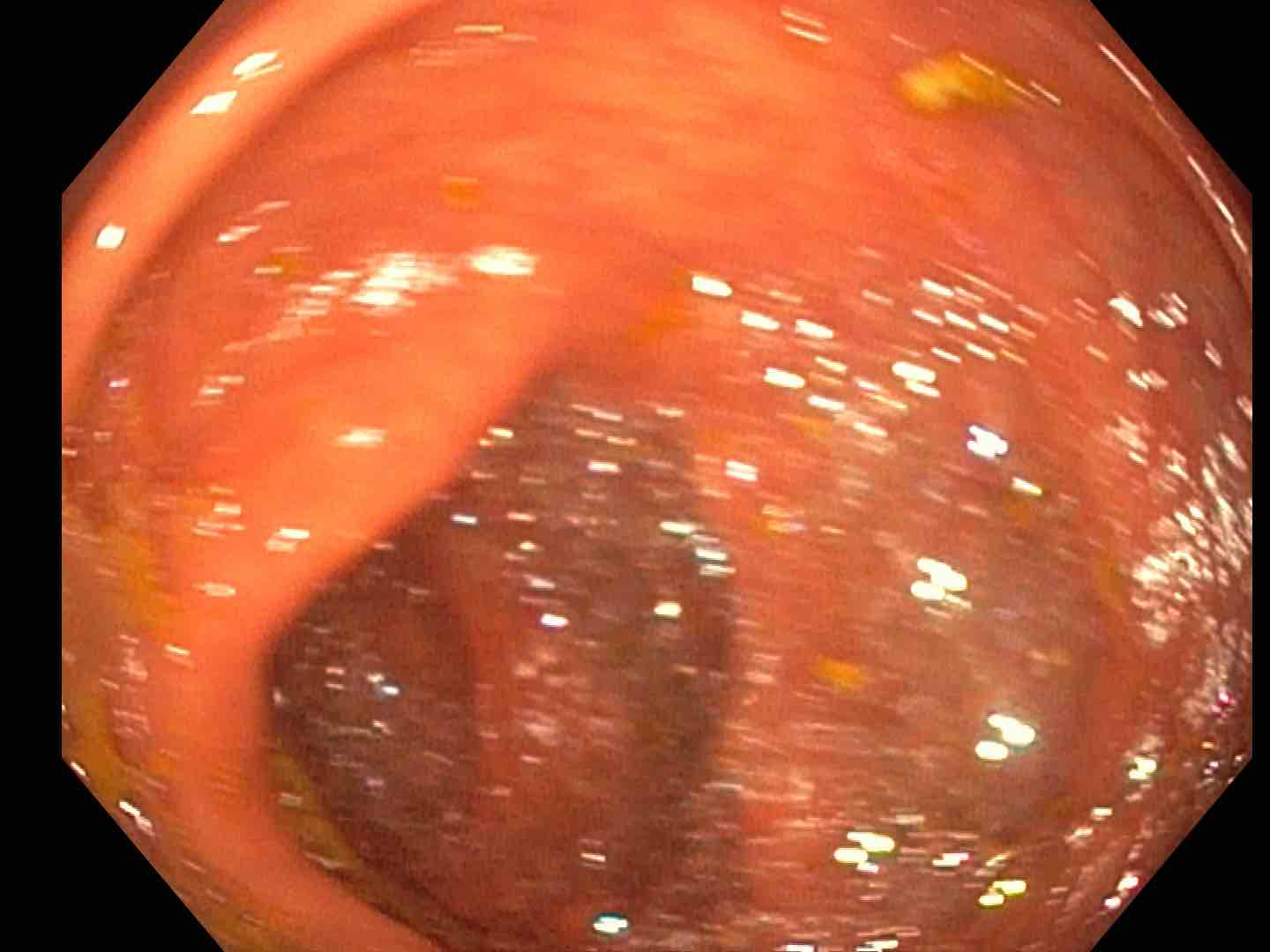}
		&\includegraphics[align=c,width=0.09\linewidth]{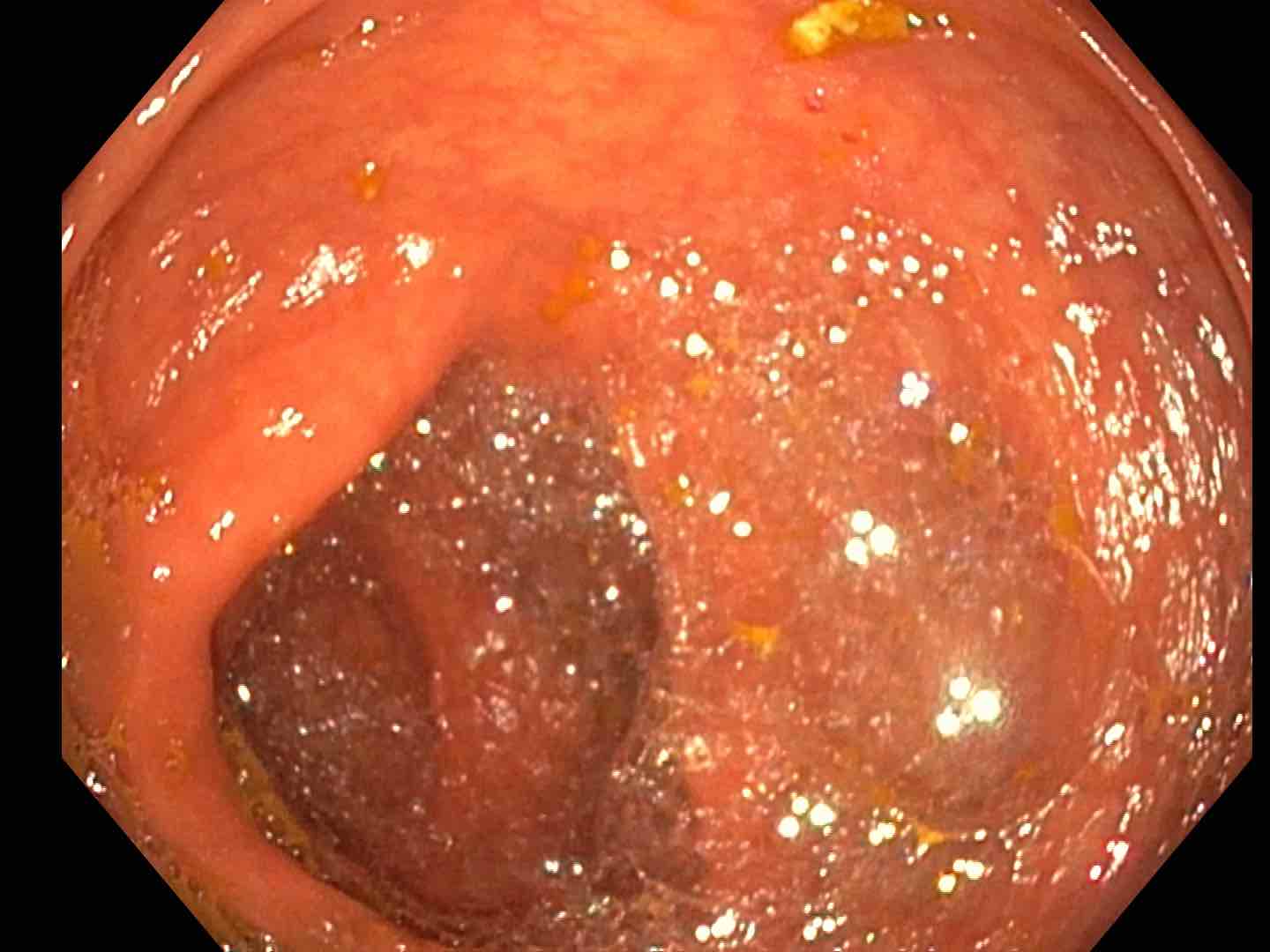}
		&\includegraphics[align=c,width=0.09\linewidth]{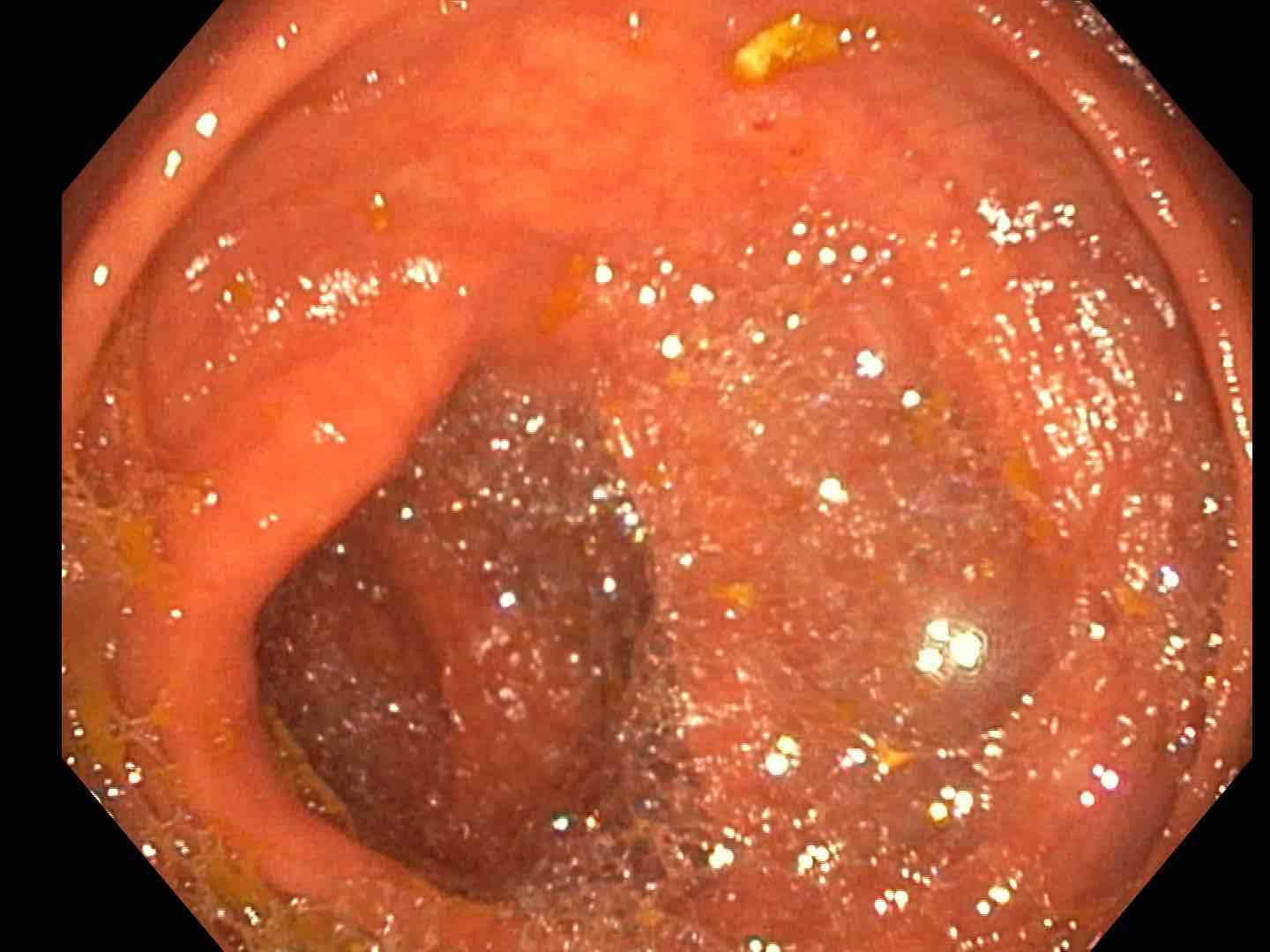}
		&\includegraphics[align=c,width=0.09\linewidth]{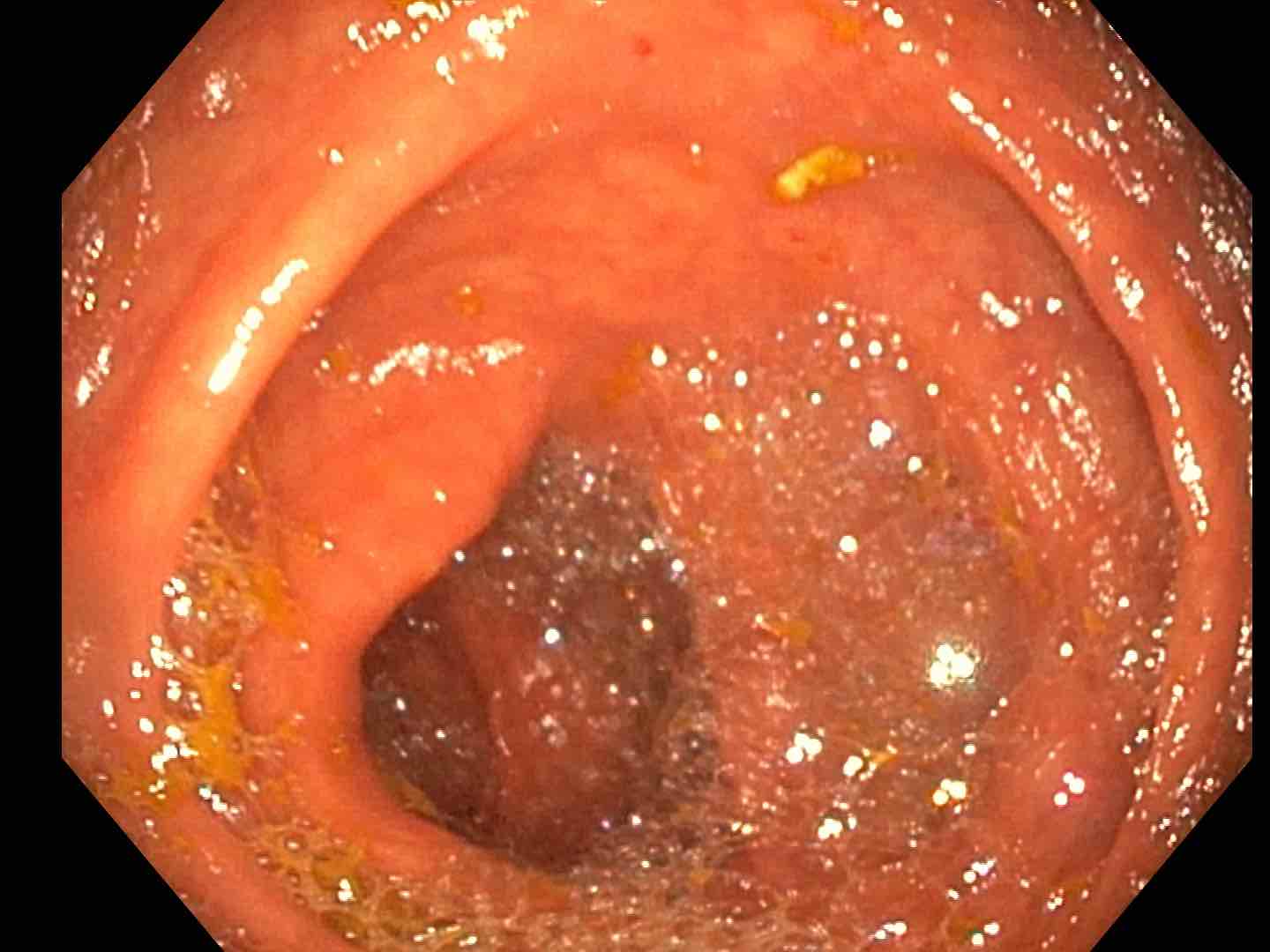}
		&\includegraphics[align=c,width=0.09\linewidth]{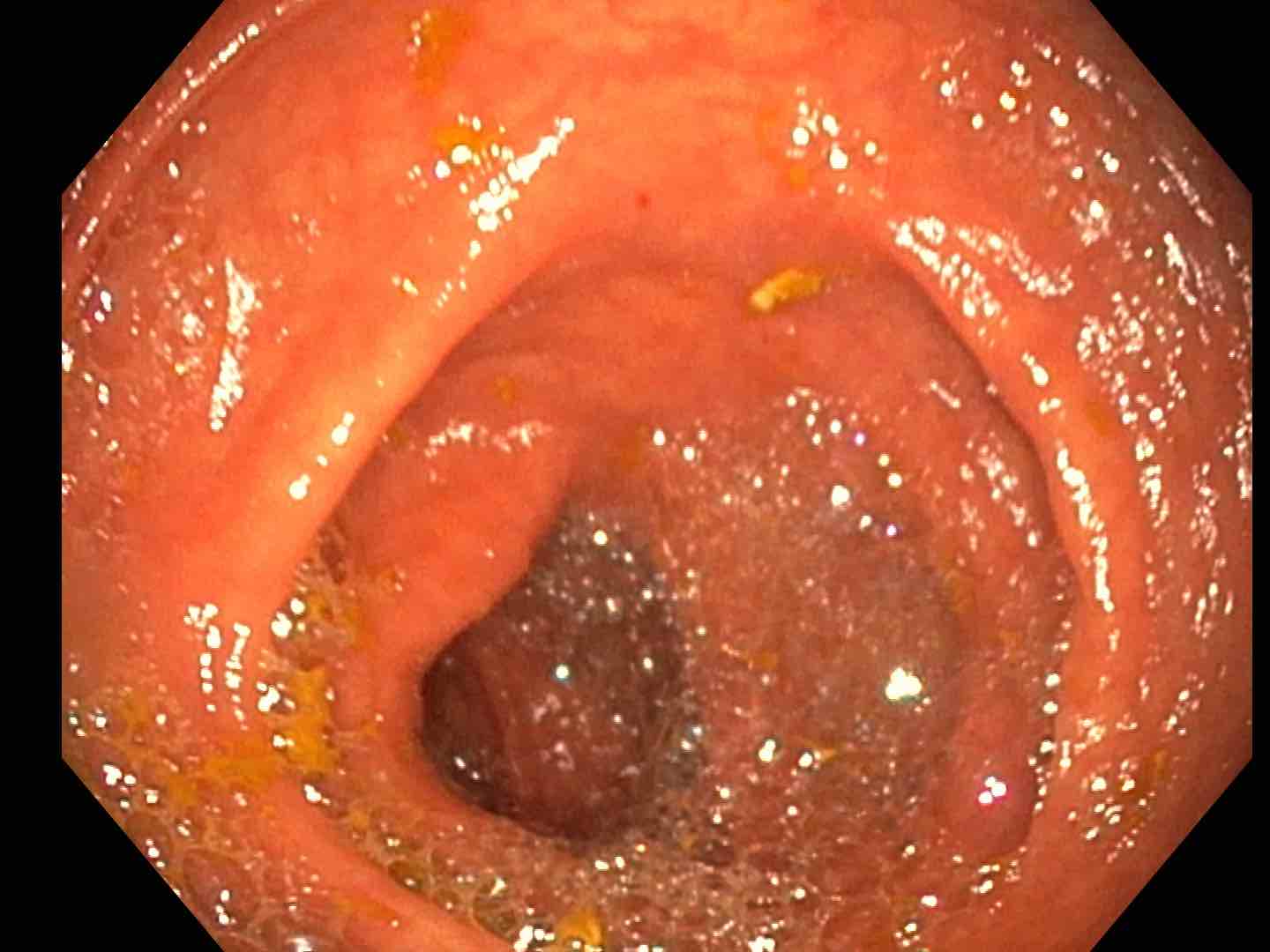}
		&\includegraphics[align=c,width=0.09\linewidth]{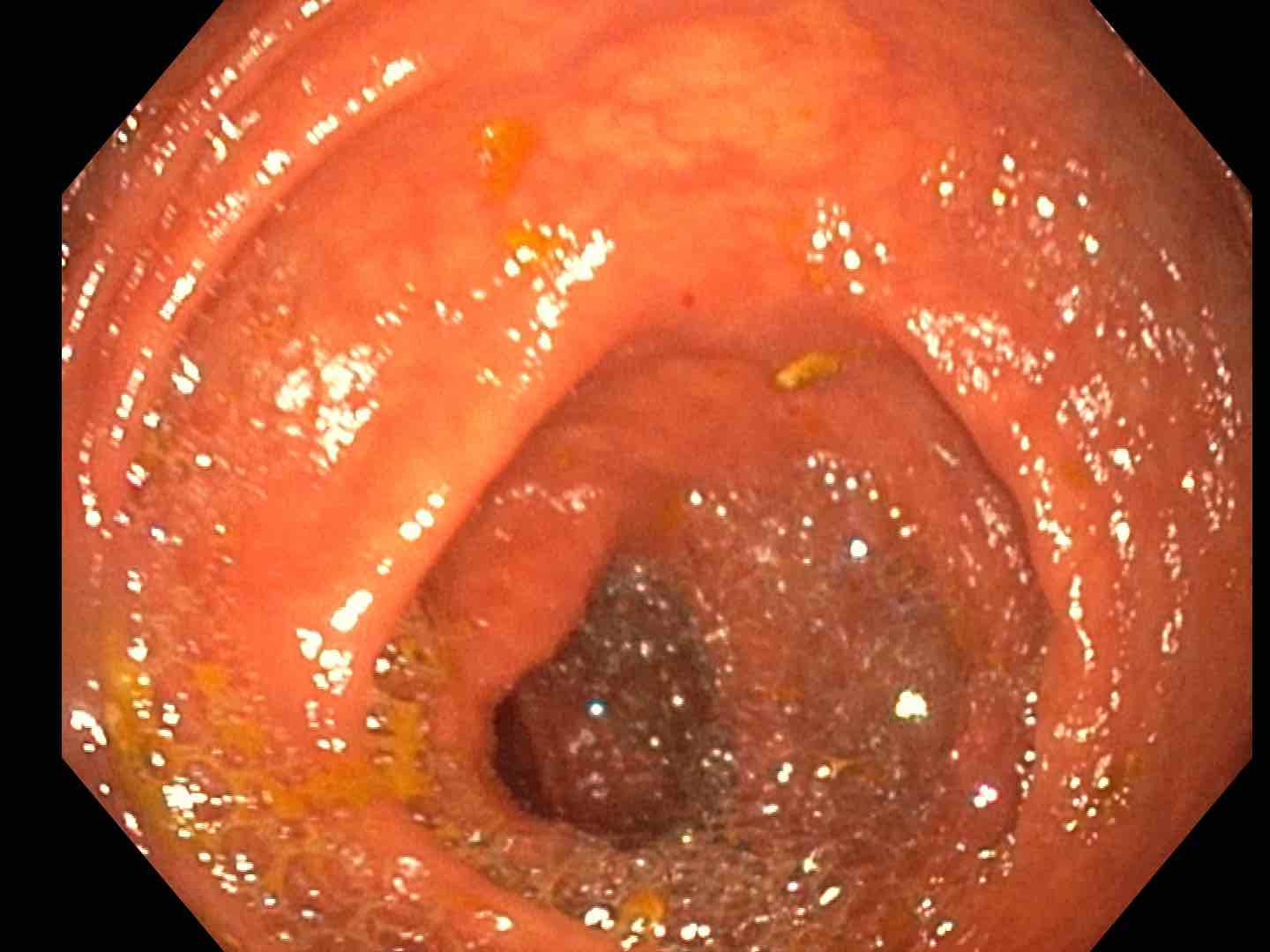}
		&\includegraphics[align=c,width=0.09\linewidth]{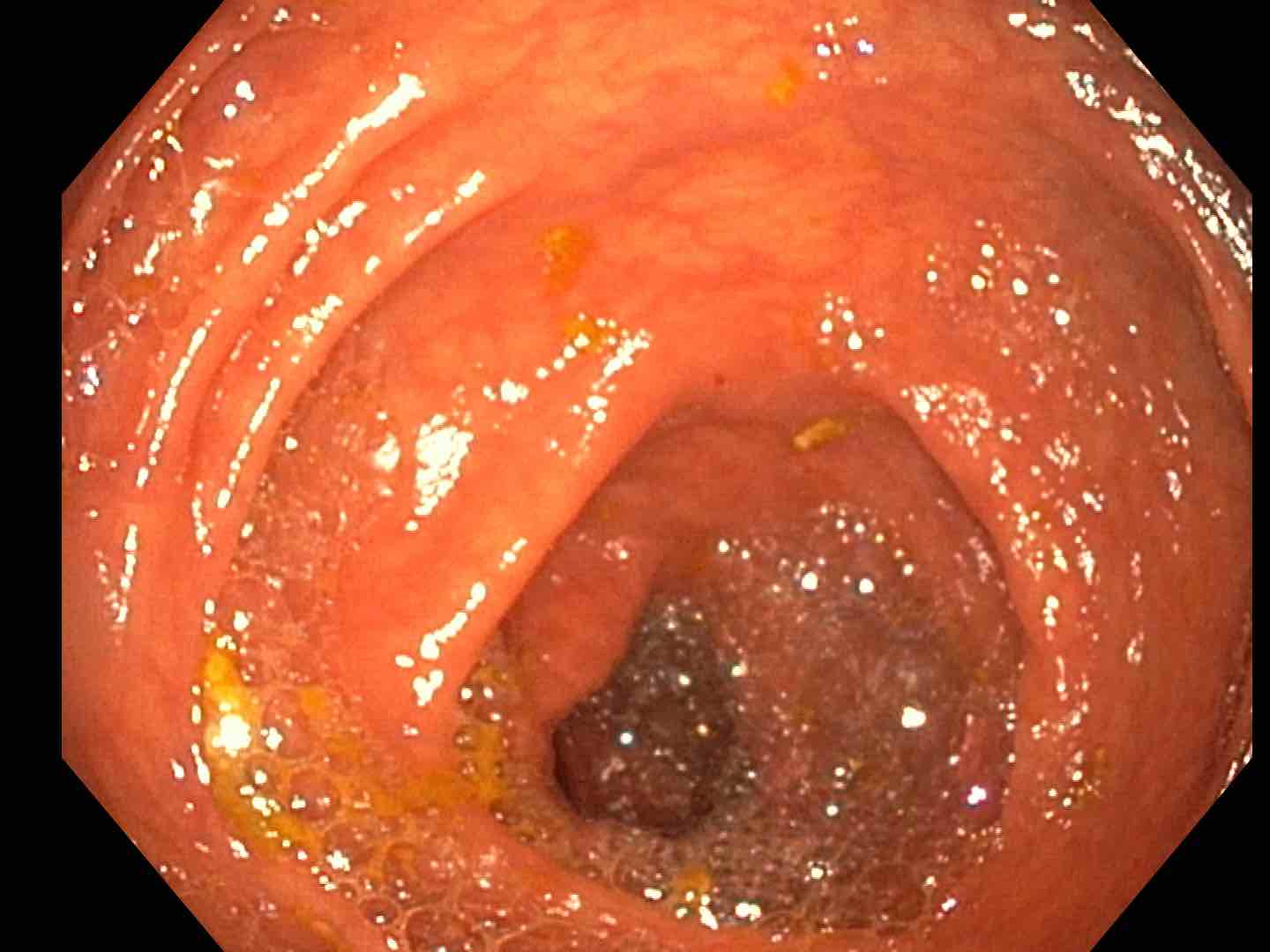}
		&\includegraphics[align=c,width=0.09\linewidth]{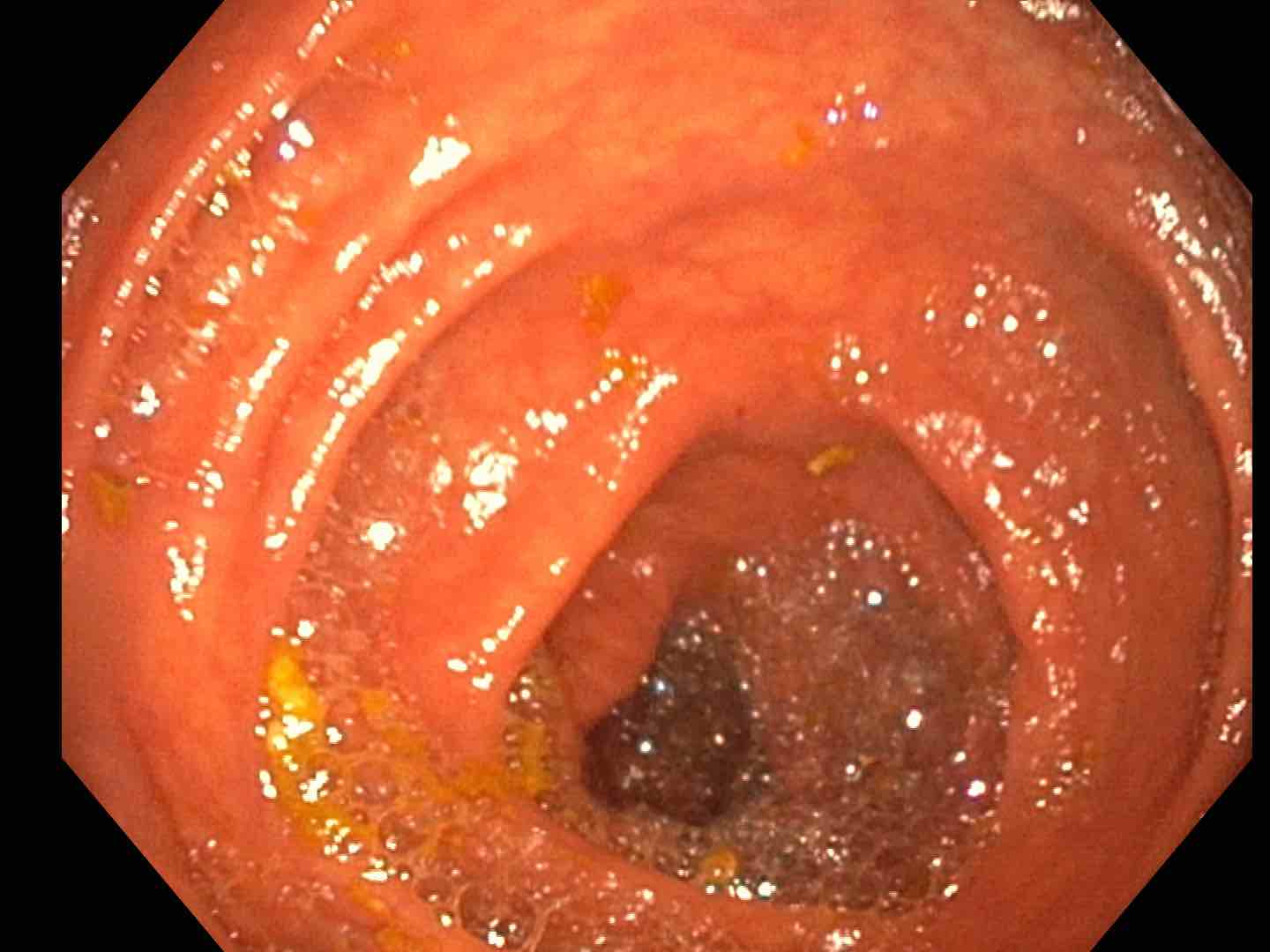}
		&\includegraphics[align=c,width=0.09\linewidth]{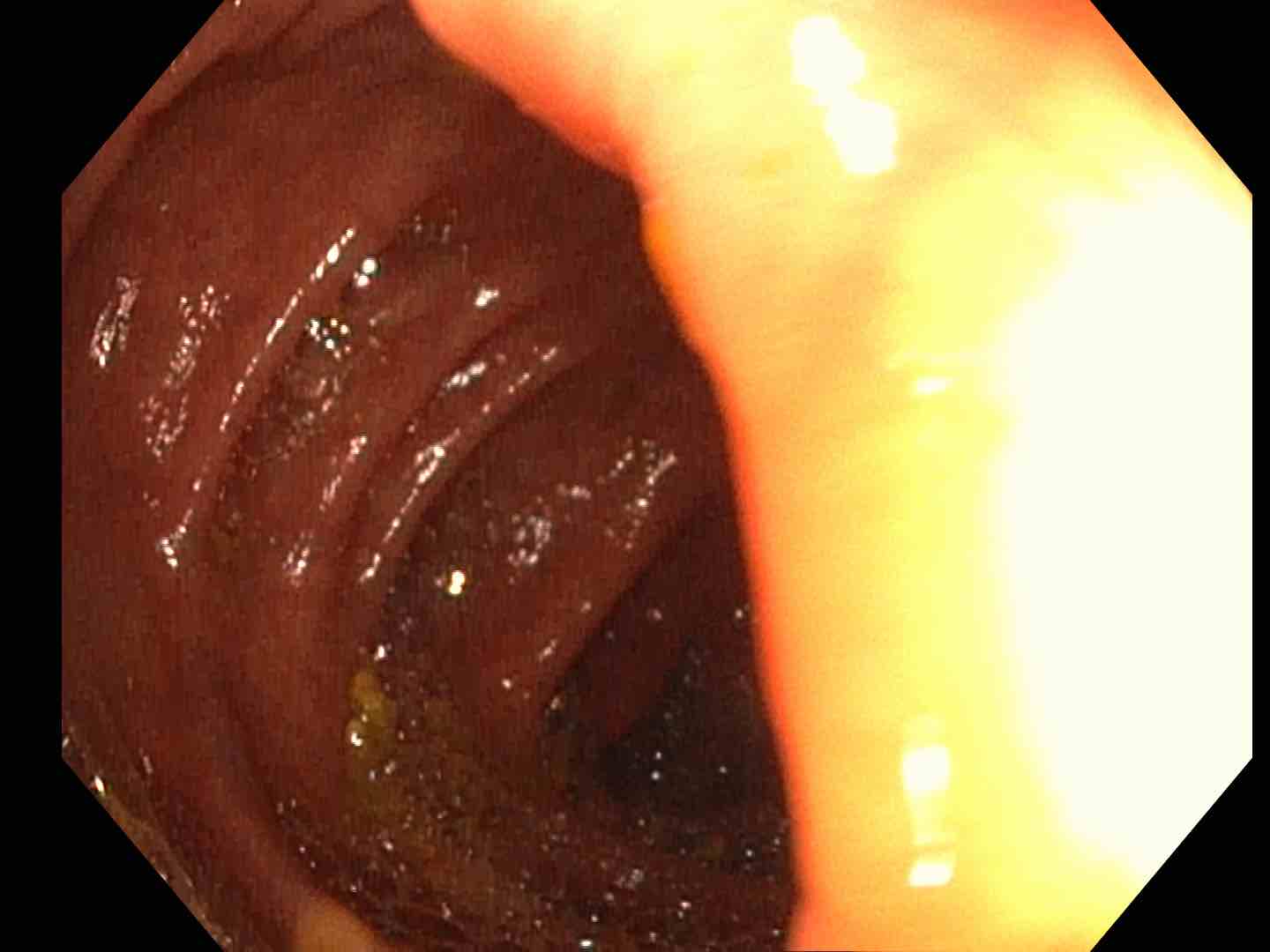}
		&\includegraphics[align=c,width=0.09\linewidth]{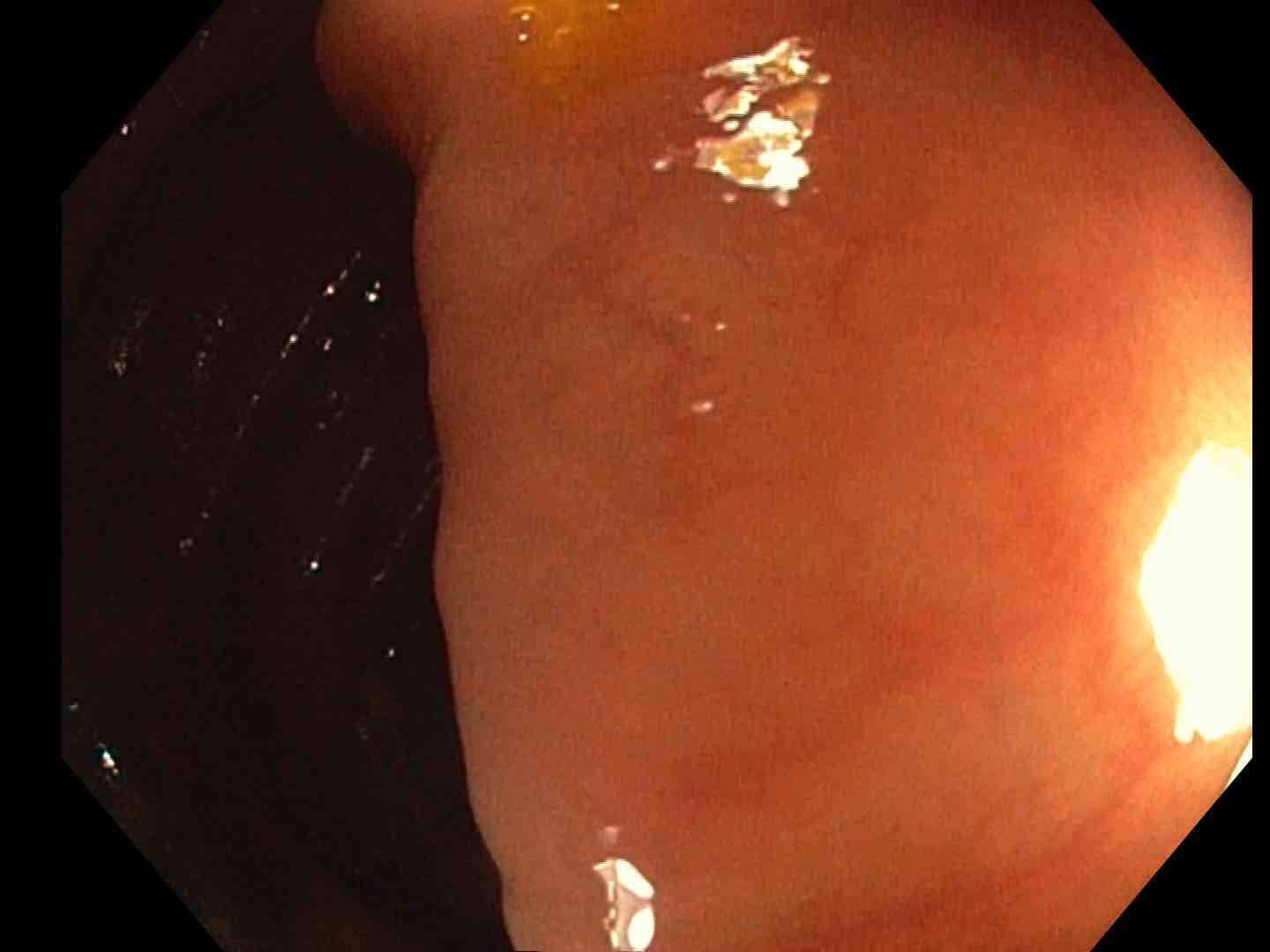}
	\\[0.43cm]
		\footnotesize{(b)}
		&\includegraphics[align=c,align=c,width=0.09\linewidth]{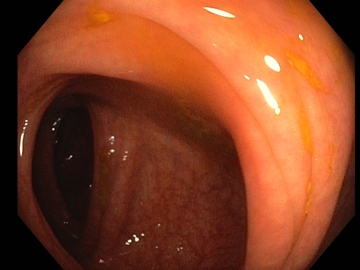}
		&\includegraphics[align=c,width=0.09\linewidth]{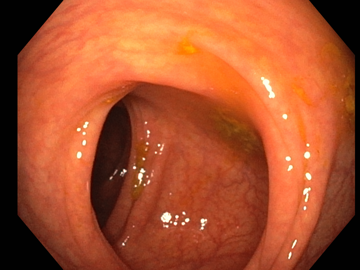}
		&\includegraphics[align=c,width=0.09\linewidth]{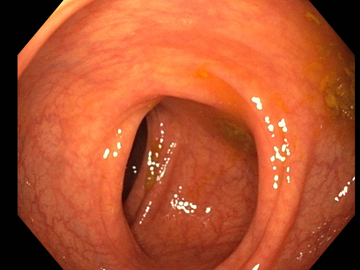}
		&\includegraphics[align=c,width=0.09\linewidth]{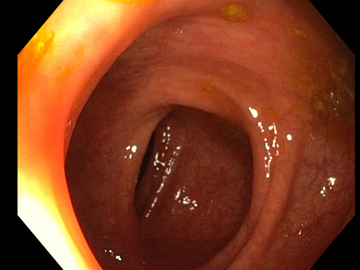}
		&\includegraphics[align=c,width=0.09\linewidth]{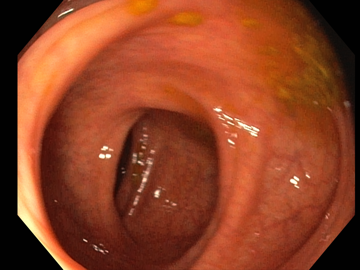}
		&\includegraphics[align=c,width=0.09\linewidth]{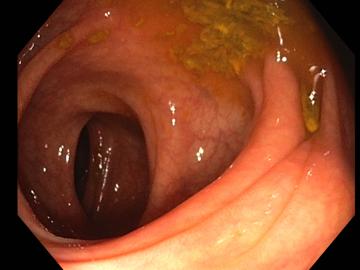}
		&\includegraphics[align=c,width=0.09\linewidth]{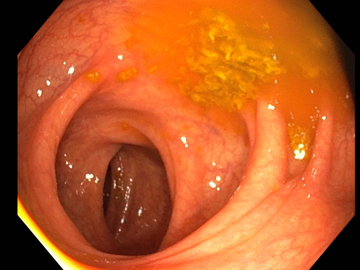}
		&\includegraphics[align=c,width=0.09\linewidth]{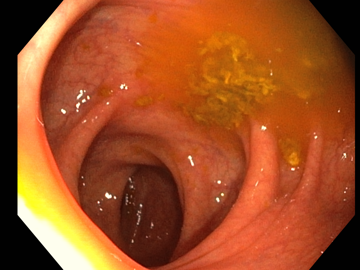}
		&\includegraphics[align=c,width=0.09\linewidth]{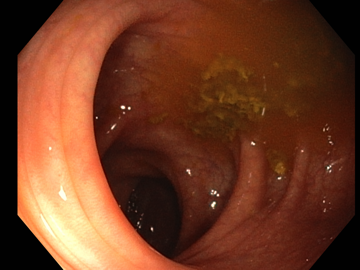}
		&\includegraphics[align=c,width=0.09\linewidth]{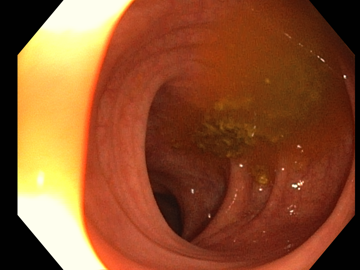}
	\\[0.43cm]
	\footnotesize{(c)}
		&\includegraphics[align=c,align=c,width=0.09\linewidth]{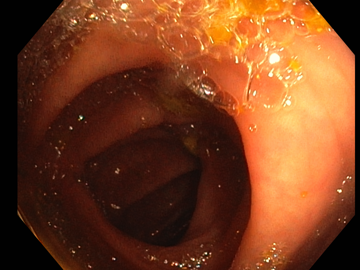}
		&\includegraphics[align=c,width=0.09\linewidth]{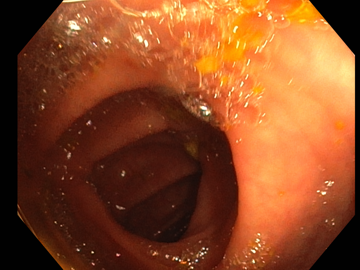}
		&\includegraphics[align=c,width=0.09\linewidth]{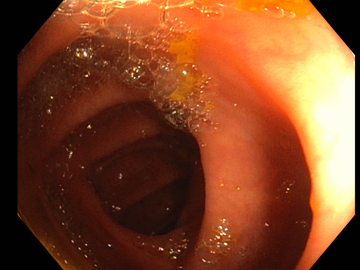}
		&\includegraphics[align=c,width=0.09\linewidth]{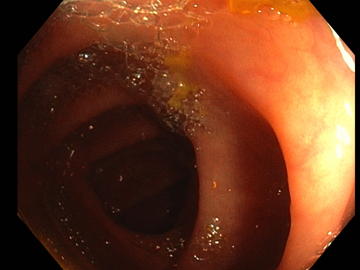}
		&\includegraphics[align=c,width=0.09\linewidth]{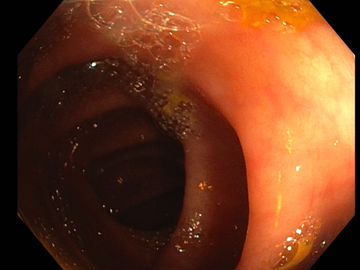}
		&\includegraphics[align=c,width=0.09\linewidth]{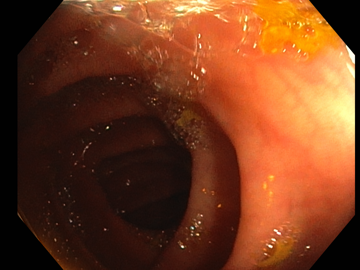}
		&\includegraphics[align=c,width=0.09\linewidth]{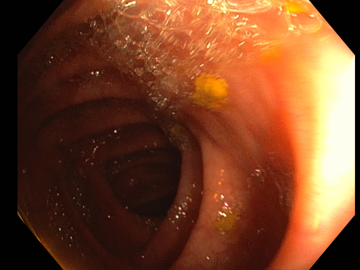}
		&\includegraphics[align=c,width=0.09\linewidth]{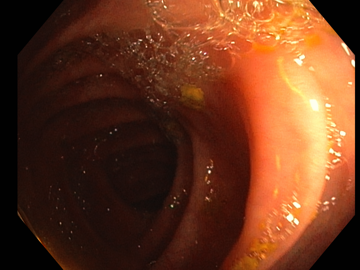}
		&\includegraphics[align=c,width=0.09\linewidth]{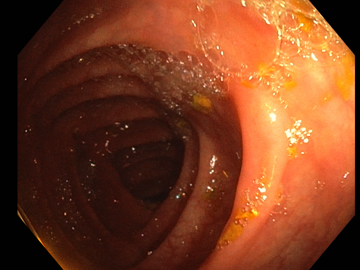}
		&\includegraphics[align=c,width=0.09\linewidth]{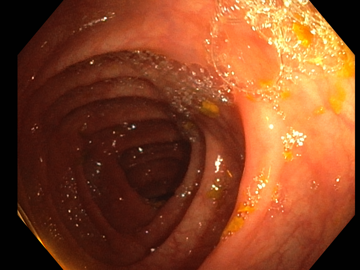}
	\\[0.43cm]
		\footnotesize{(d)}
		&\includegraphics[align=c,width=0.09\linewidth]{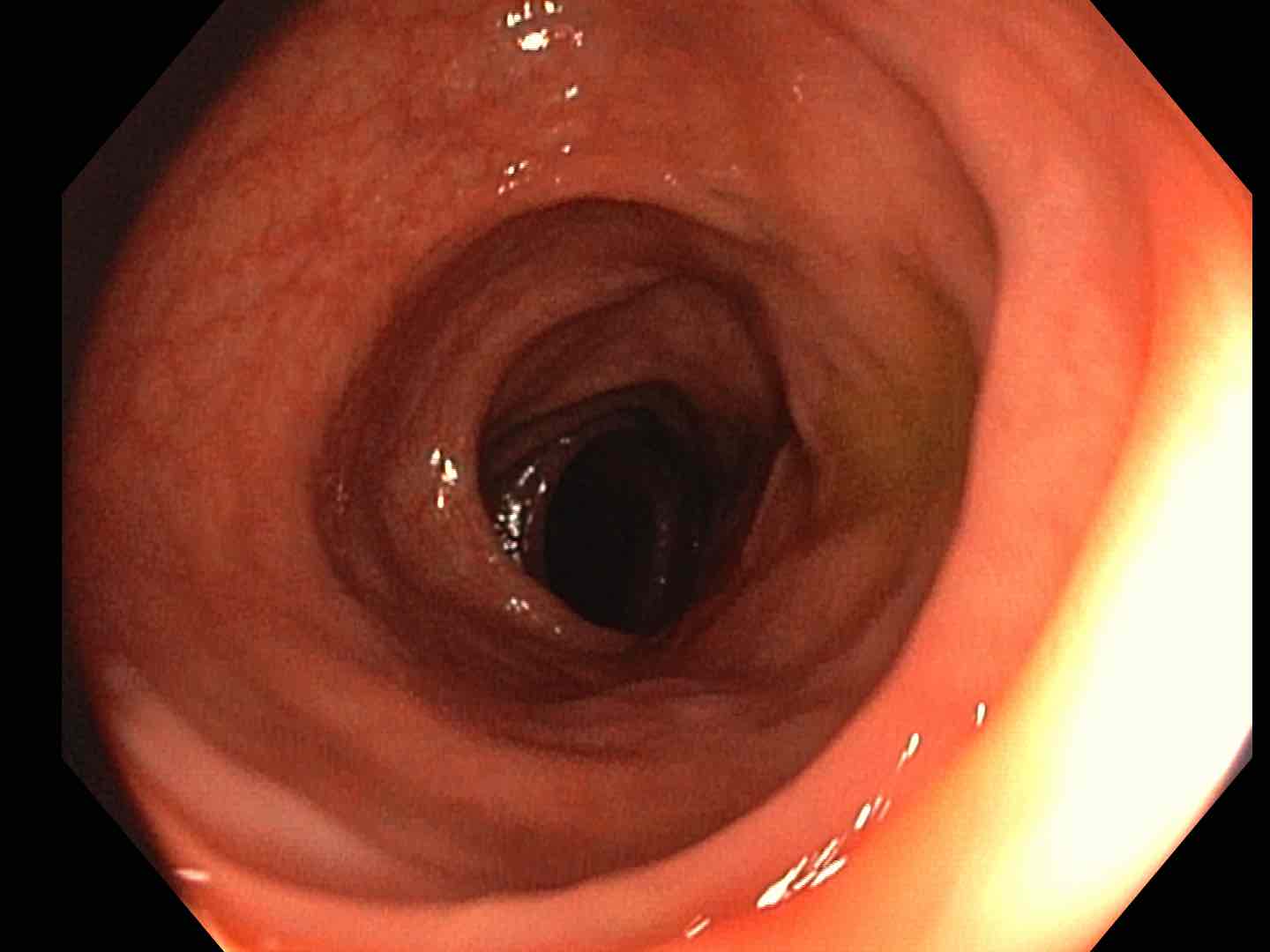}
		&\includegraphics[align=c,width=0.09\linewidth]{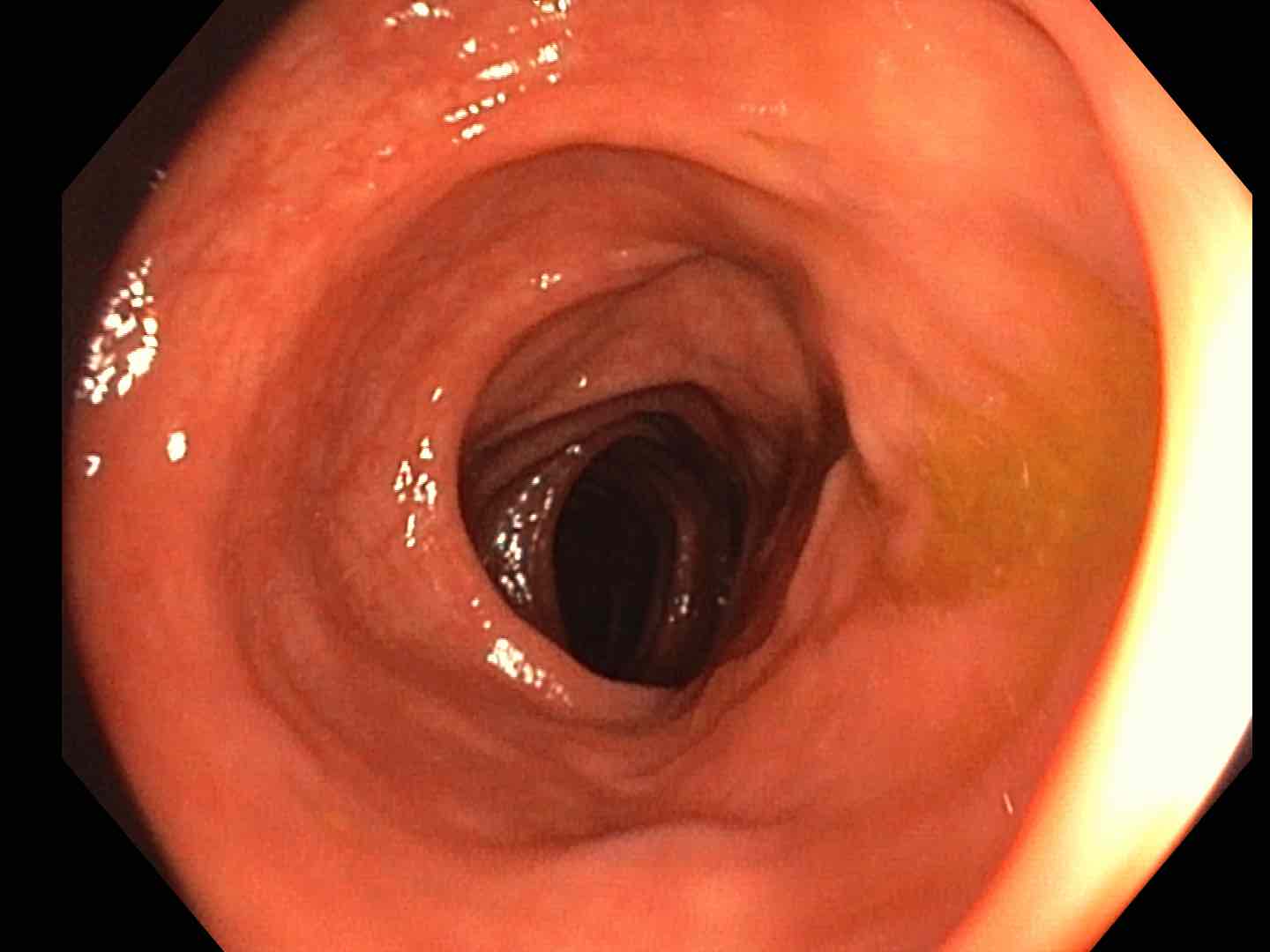}
		&\includegraphics[align=c,width=0.09\linewidth]{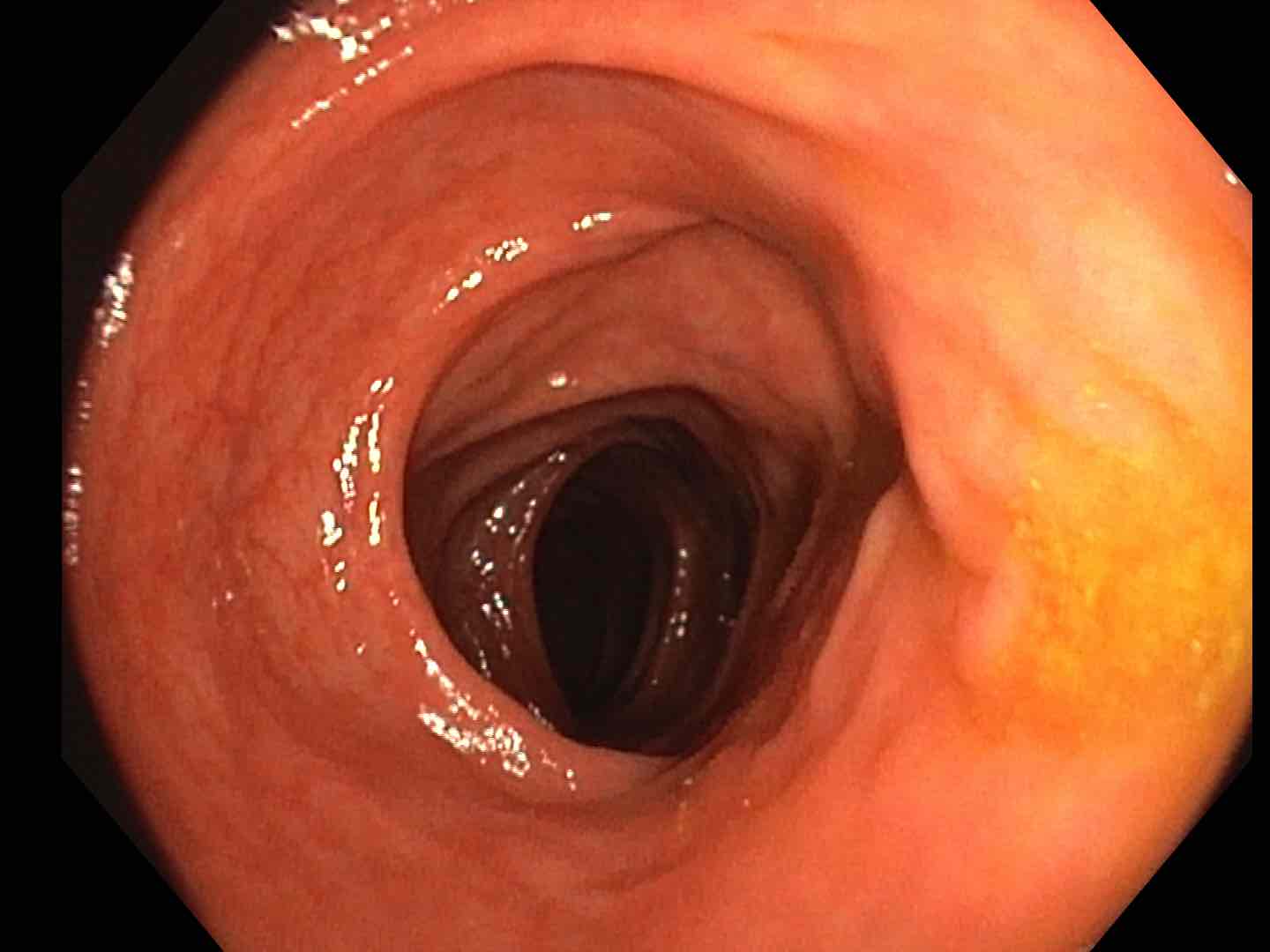}
		&\includegraphics[align=c,width=0.09\linewidth]{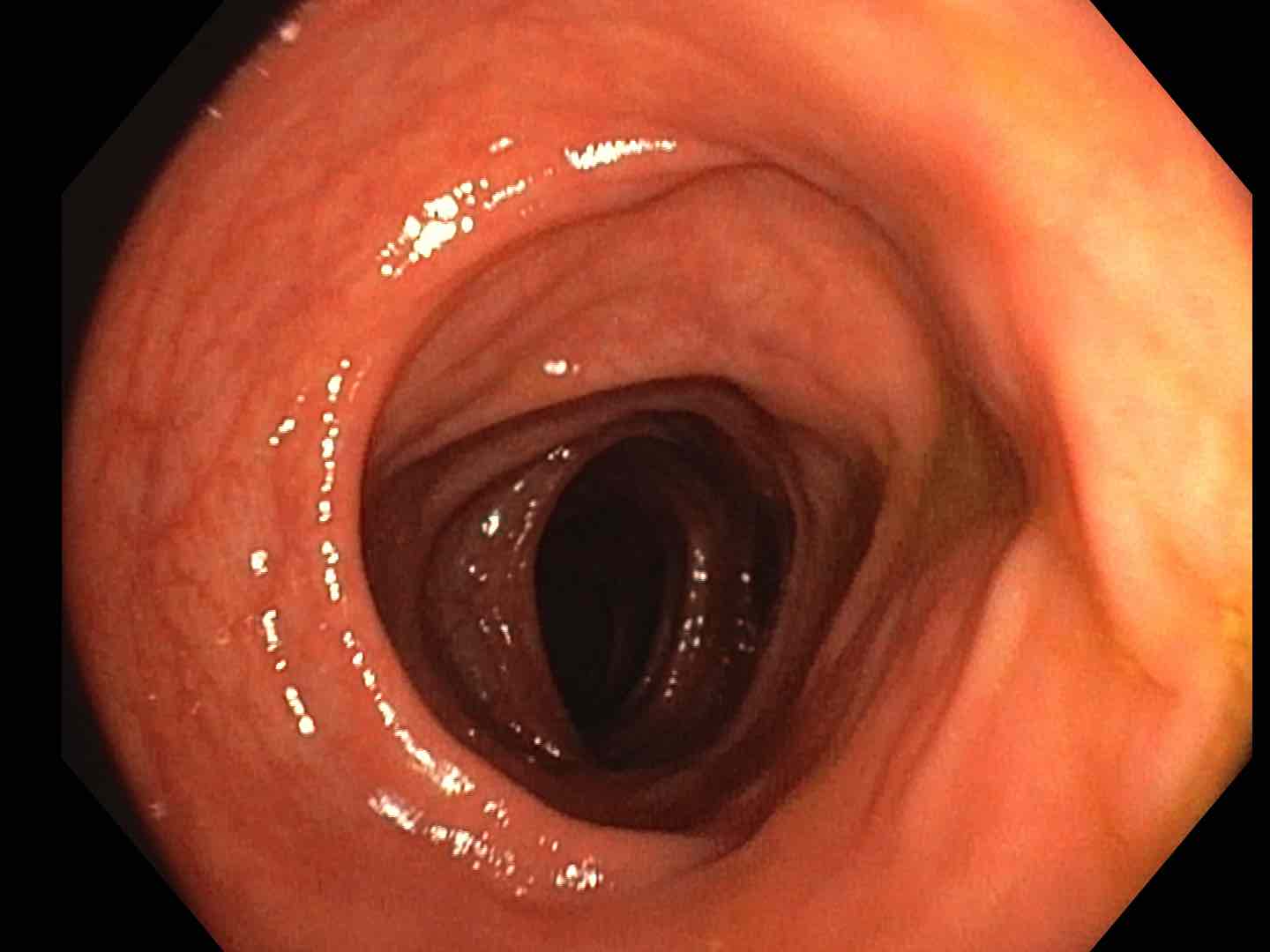}
		&\includegraphics[align=c,width=0.09\linewidth]{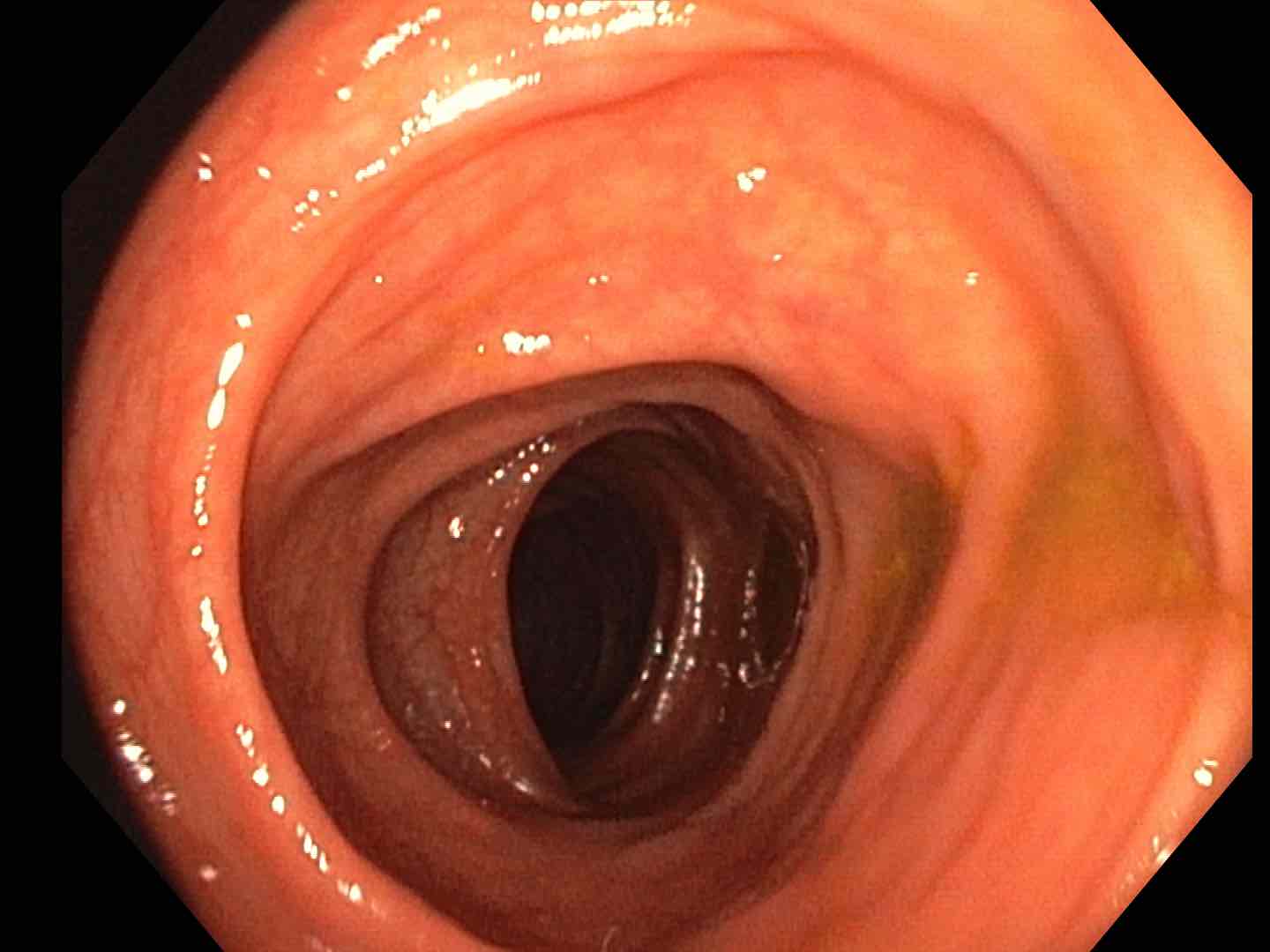}
		&\includegraphics[align=c,width=0.09\linewidth]{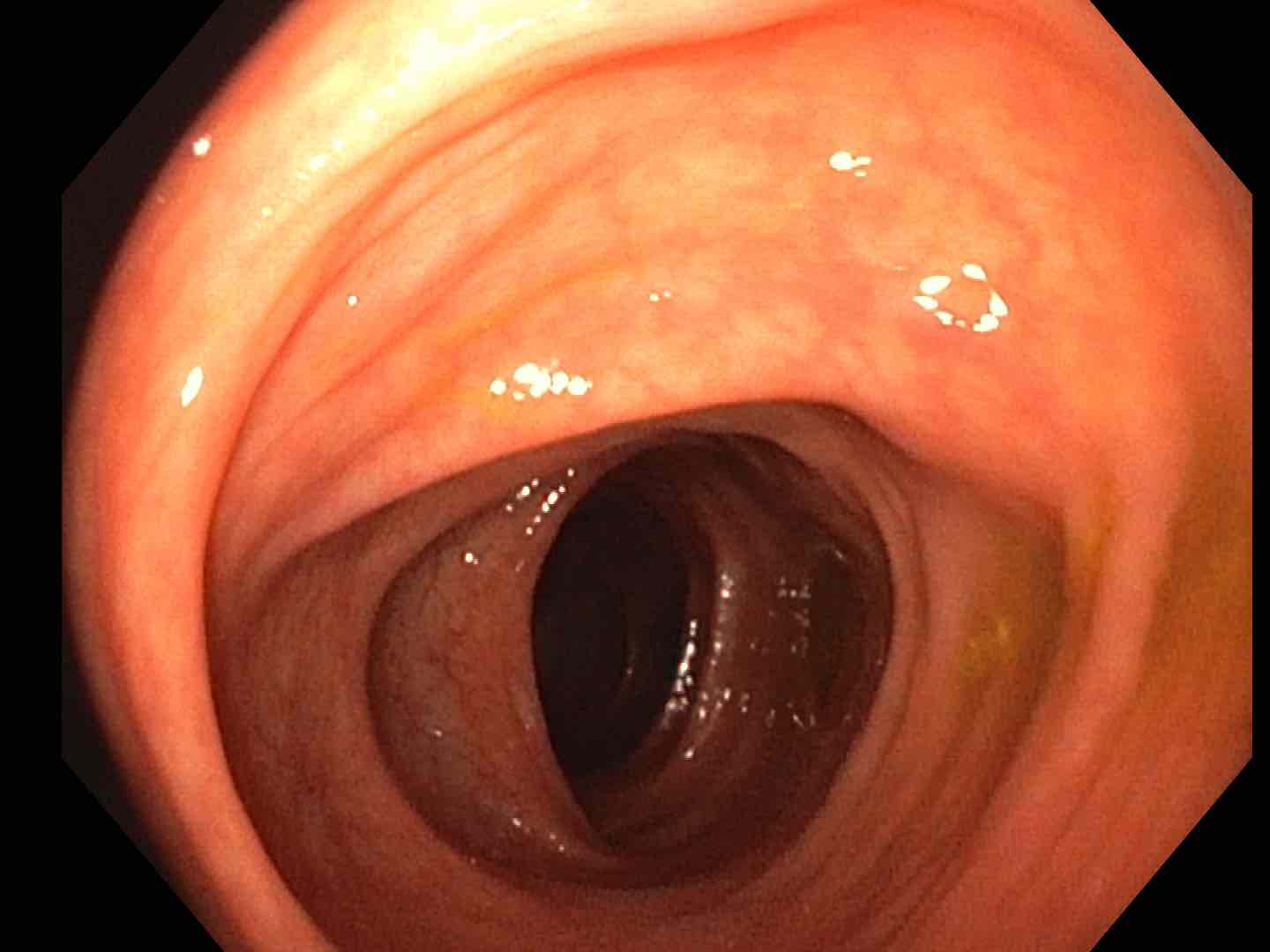}
		&\includegraphics[align=c,width=0.09\linewidth]{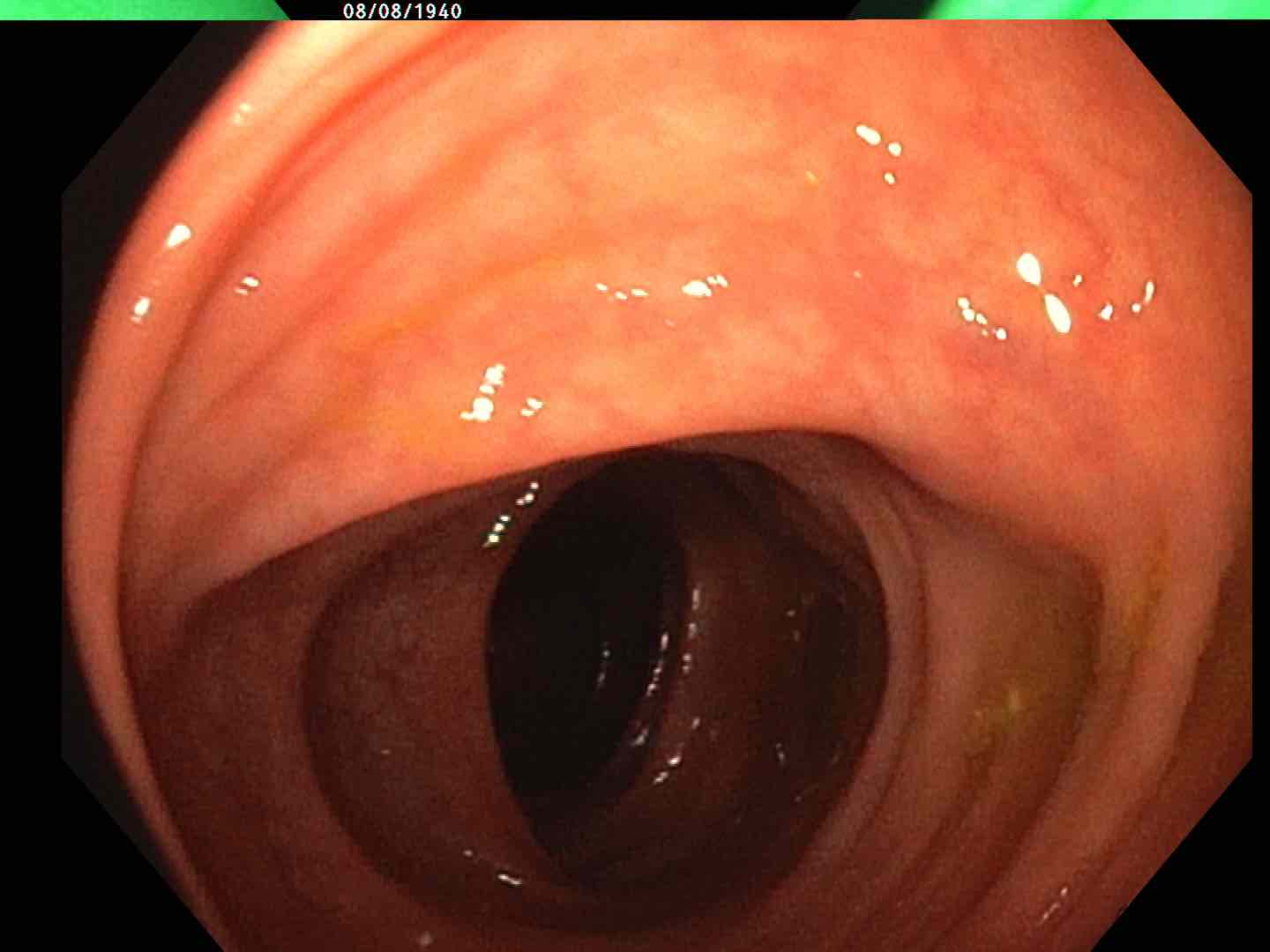}
		&\includegraphics[align=c,width=0.09\linewidth]{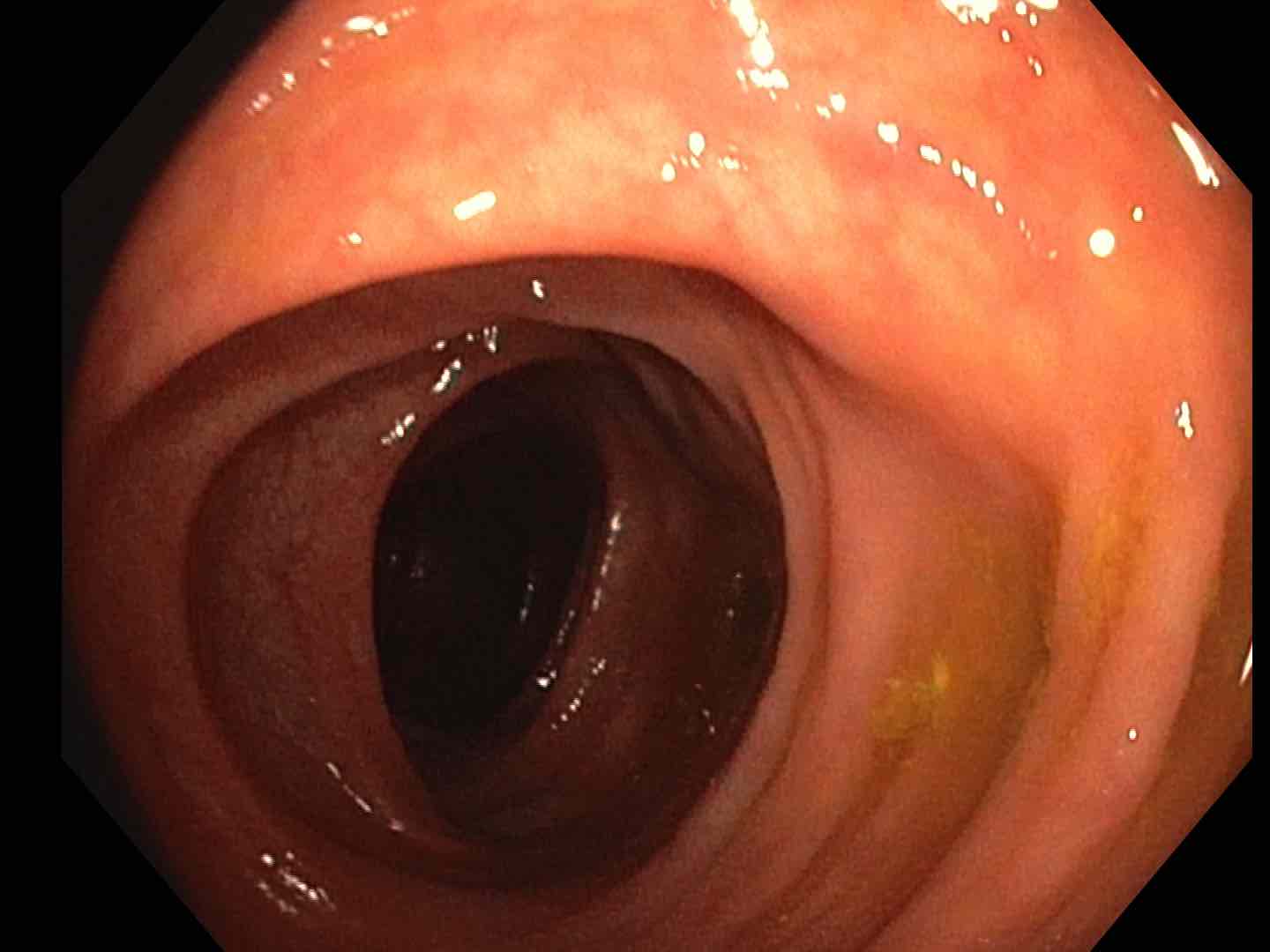}
		&\includegraphics[align=c,width=0.09\linewidth]{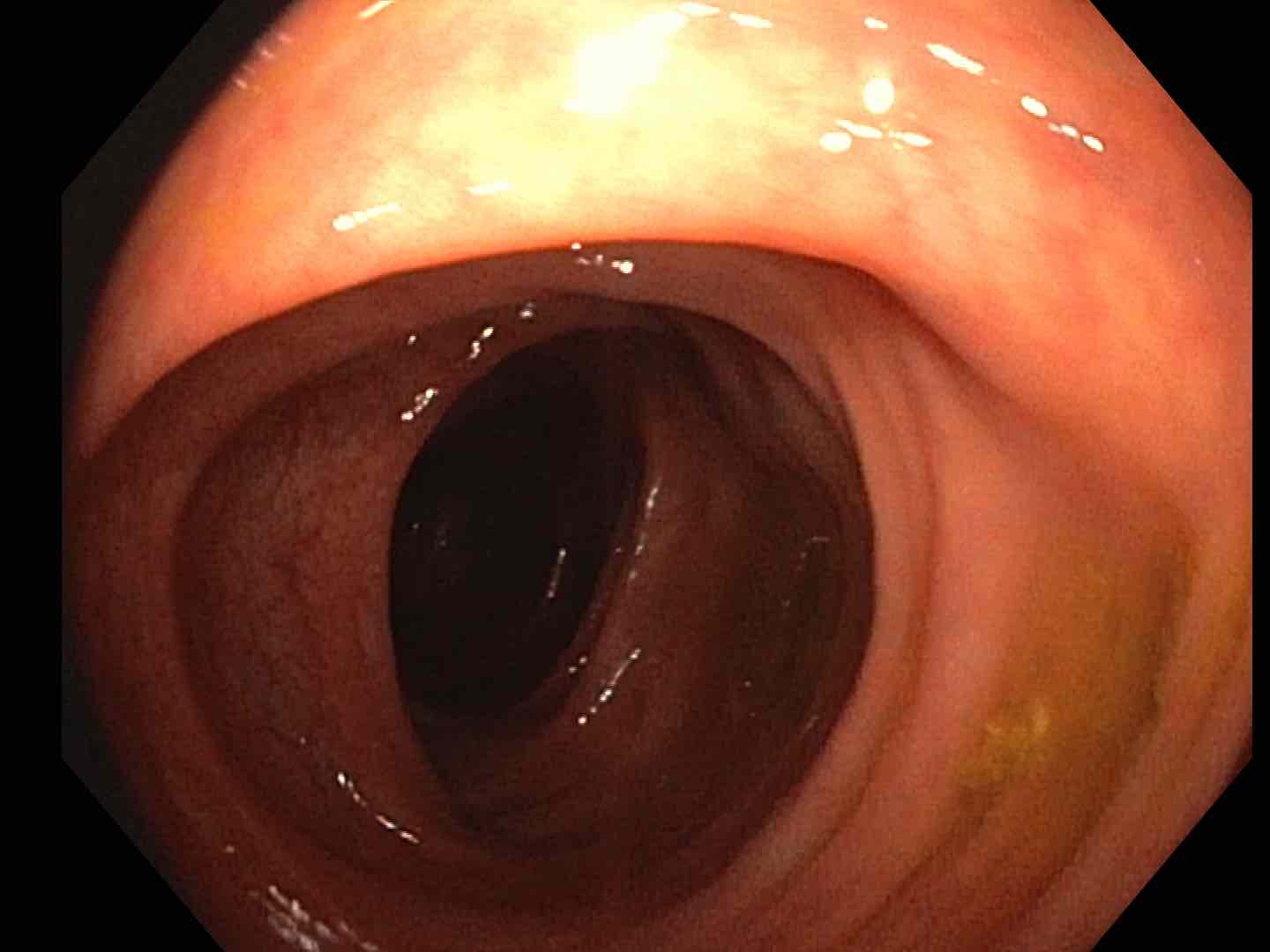}
		&\includegraphics[align=c,width=0.09\linewidth]{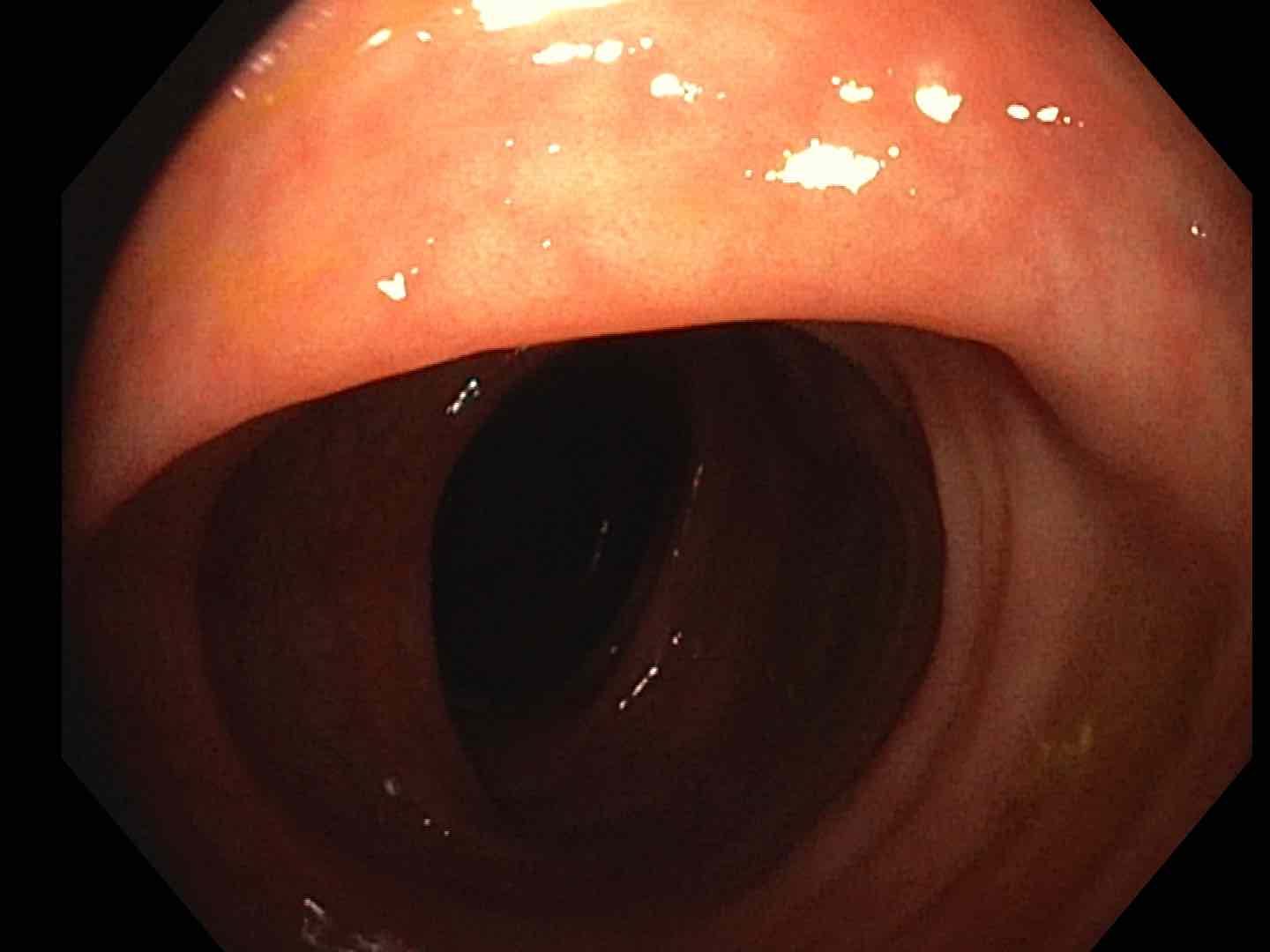}
	\\[0.43cm]
		\footnotesize{(e)}
		&\includegraphics[align=c,width=0.09\linewidth]{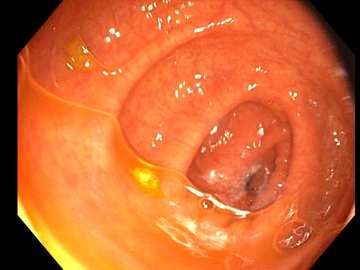}
		&\includegraphics[align=c,width=0.09\linewidth]{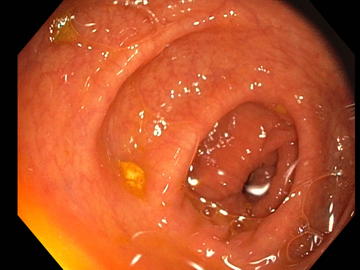}
		&\includegraphics[align=c,width=0.09\linewidth]{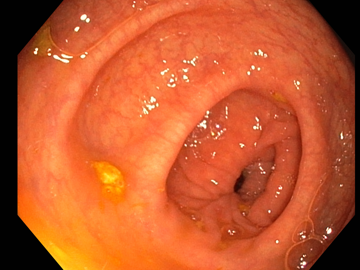}
		&\includegraphics[align=c,width=0.09\linewidth]{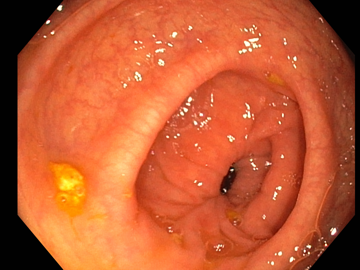}
		&\includegraphics[align=c,width=0.09\linewidth]{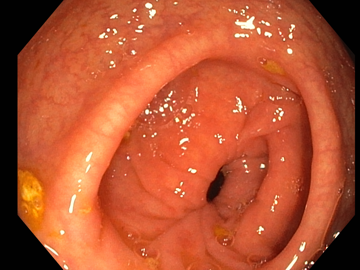}
		&\includegraphics[align=c,width=0.09\linewidth]{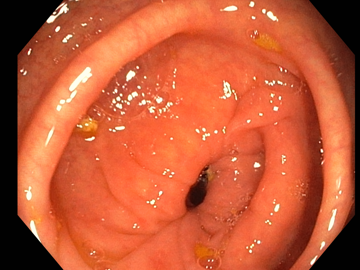}
		&\includegraphics[align=c,width=0.09\linewidth]{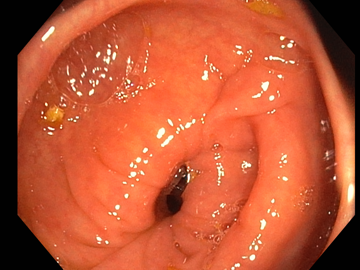}
		&\includegraphics[align=c,width=0.09\linewidth]{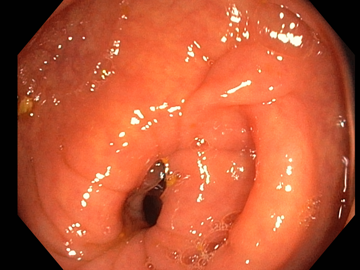}
		&\includegraphics[align=c,width=0.09\linewidth]{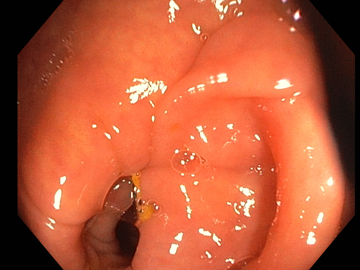}
		&\includegraphics[align=c,width=0.09\linewidth]{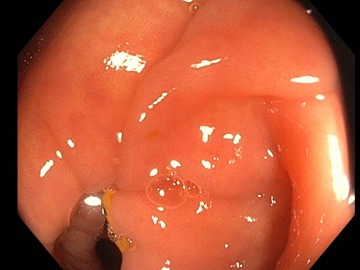}
	\\[0.43cm]
	    \footnotesize{(f)}
		&\includegraphics[align=c,width=0.09\textwidth]{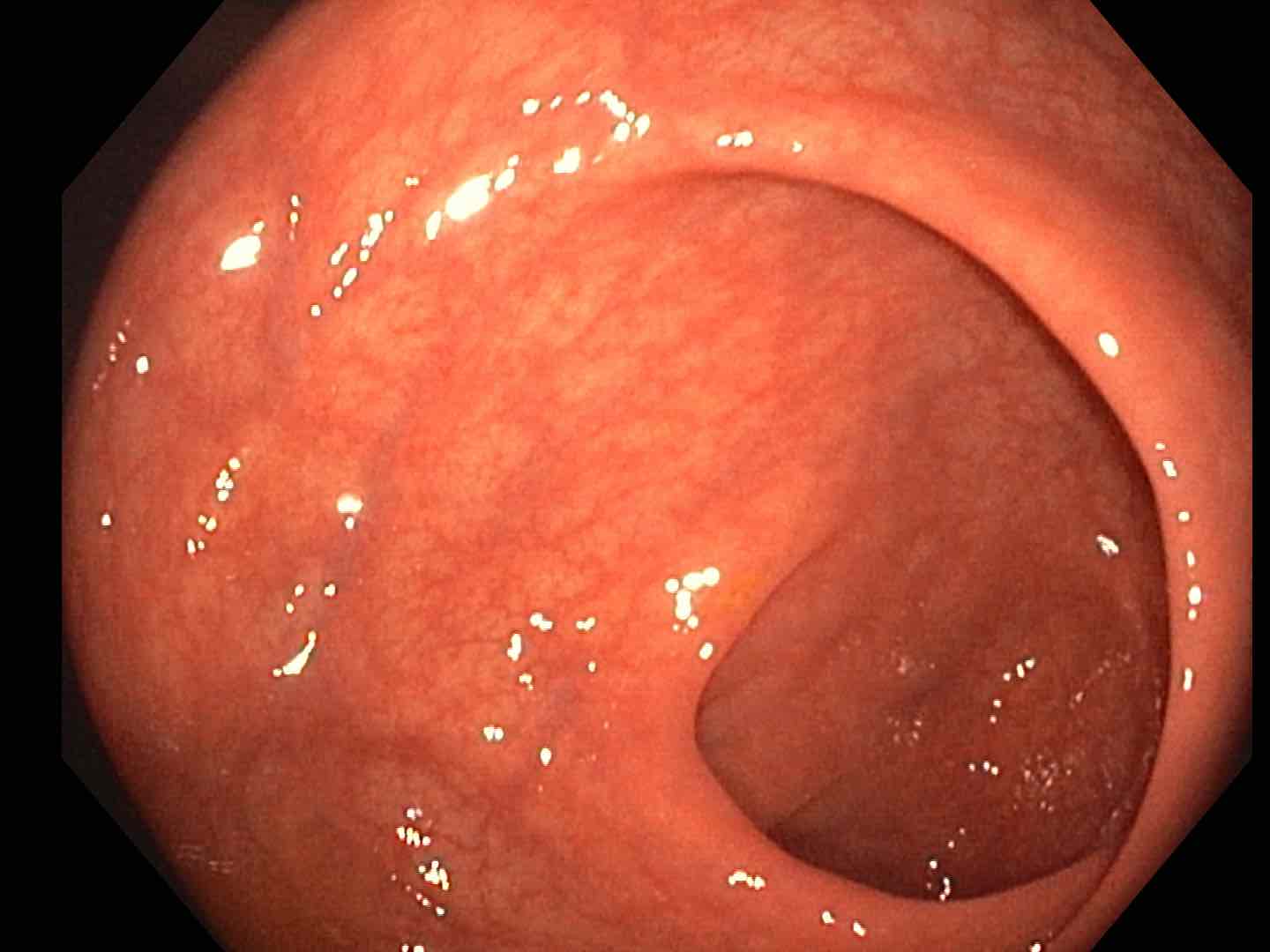}
		&\includegraphics[align=c,width=0.09\textwidth]{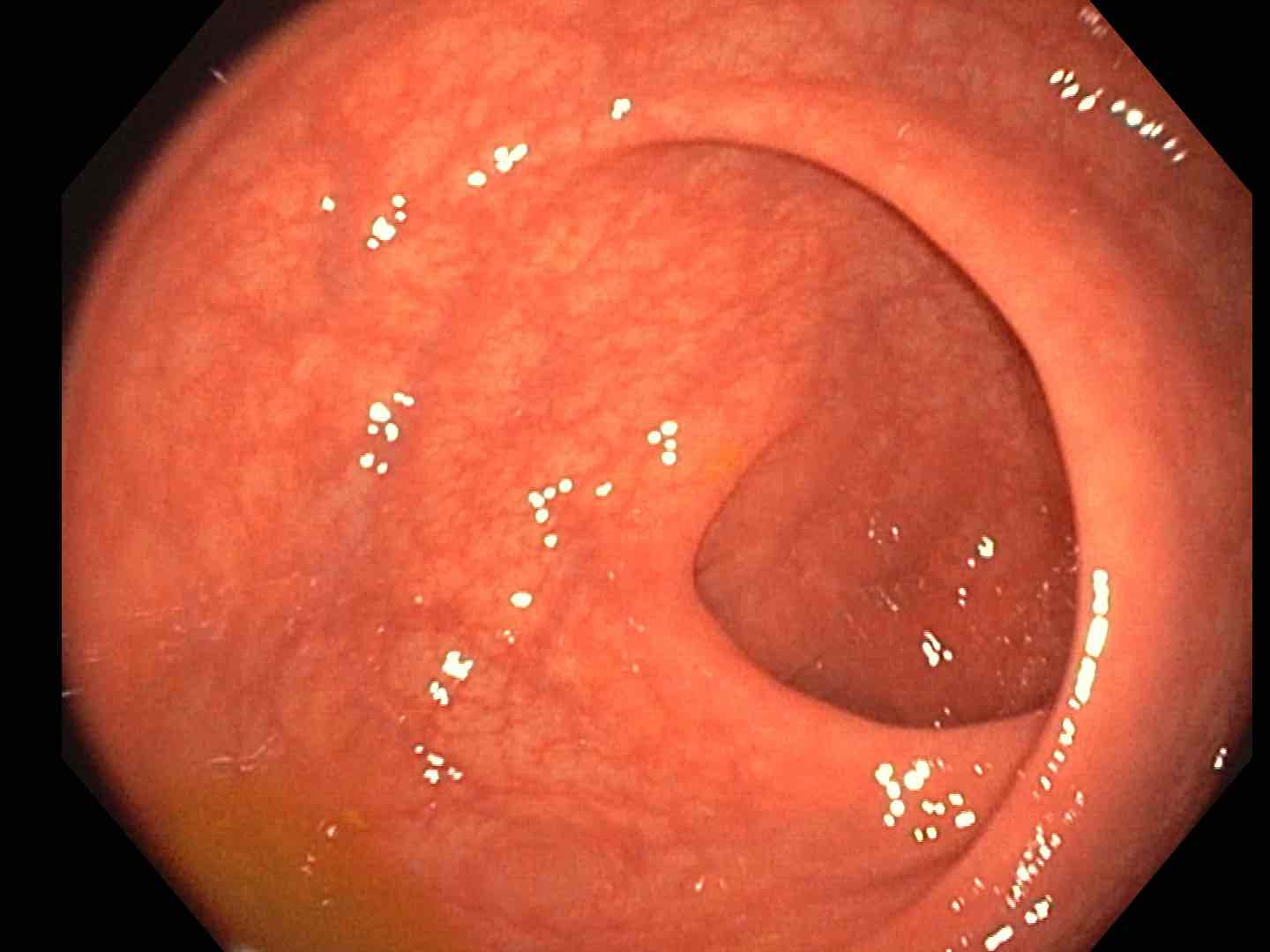}
		&\includegraphics[align=c,width=0.09\textwidth]{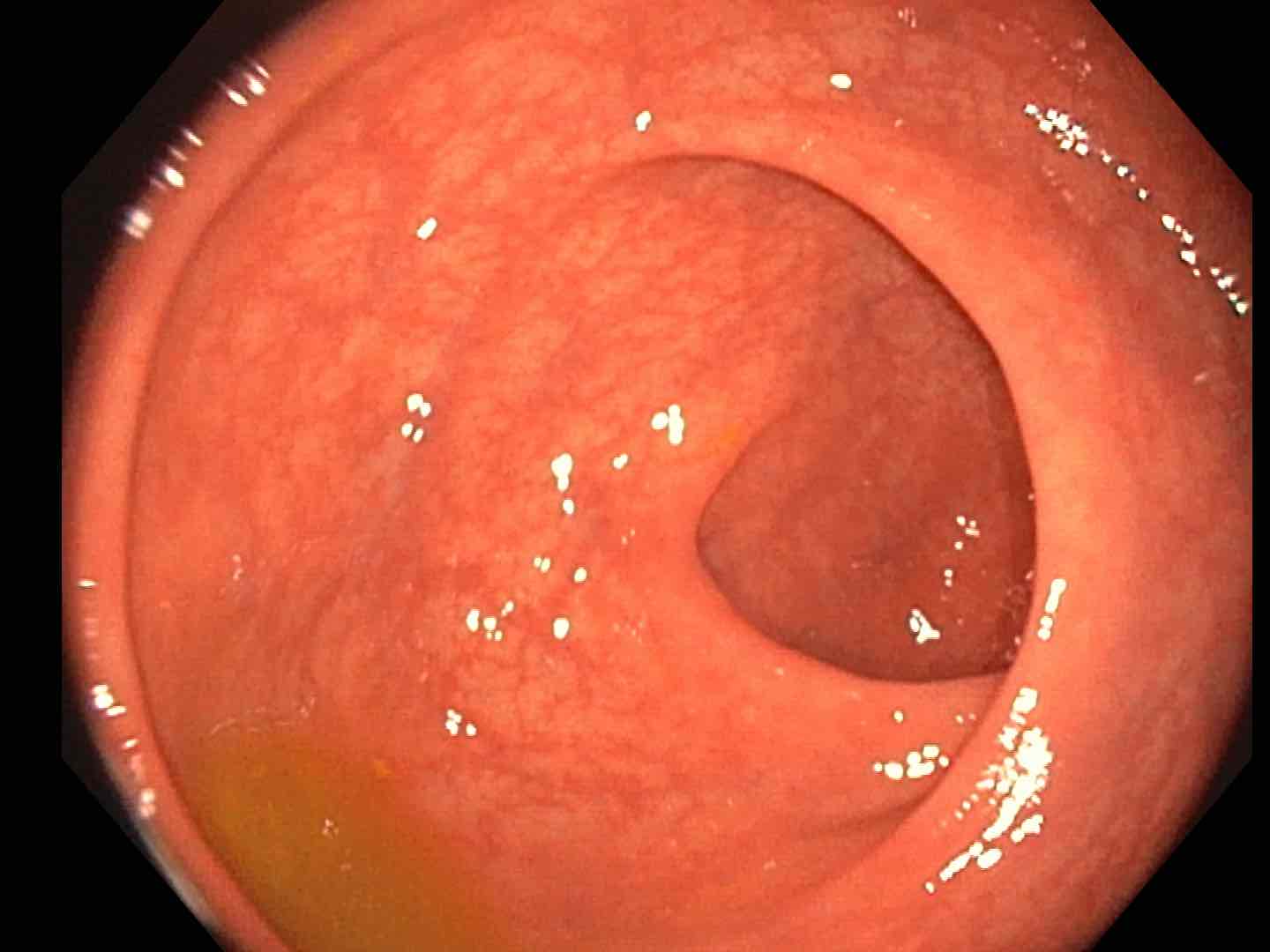}
		&\includegraphics[align=c,width=0.09\textwidth]{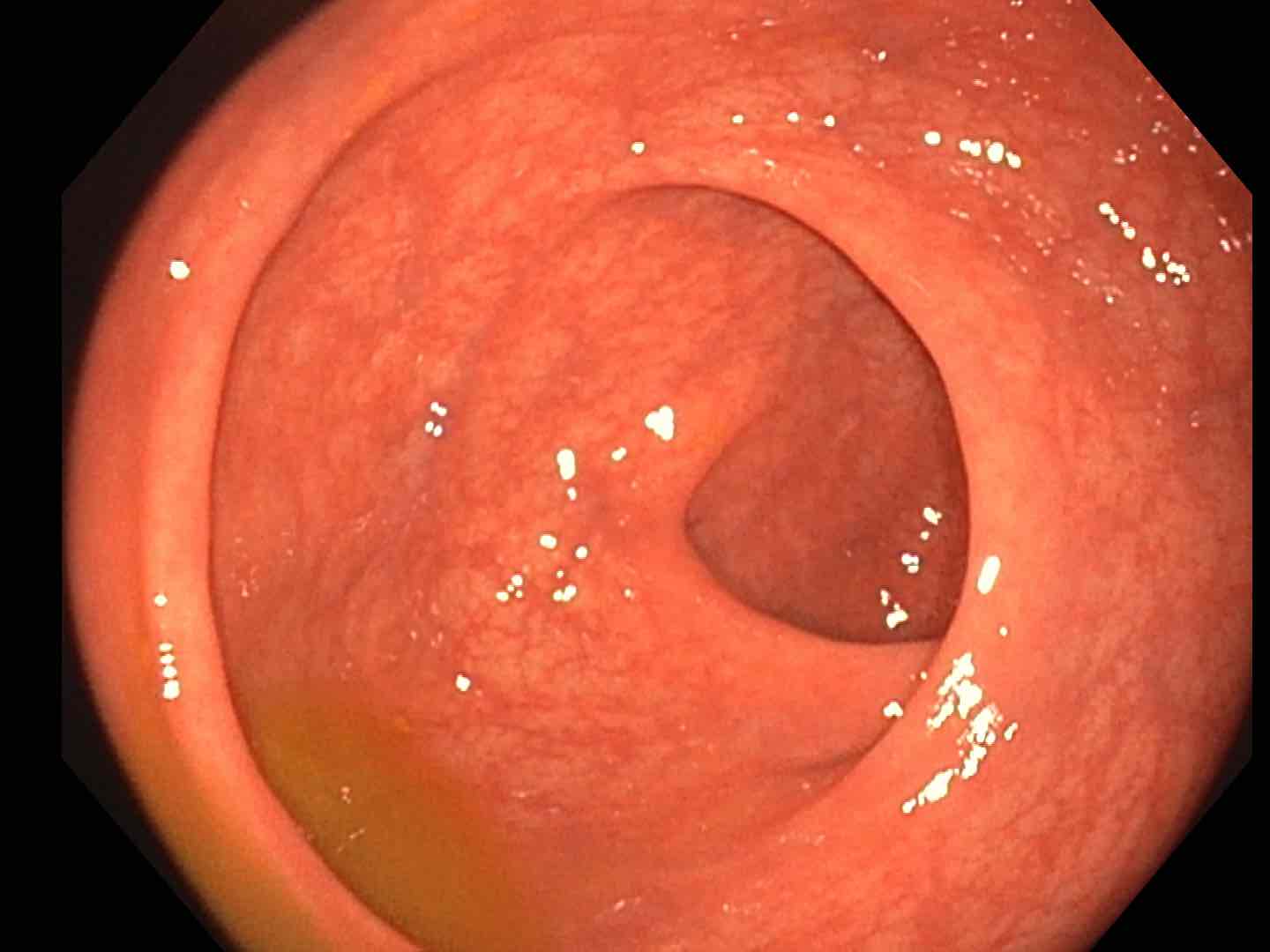}
		&\includegraphics[align=c,width=0.09\textwidth]{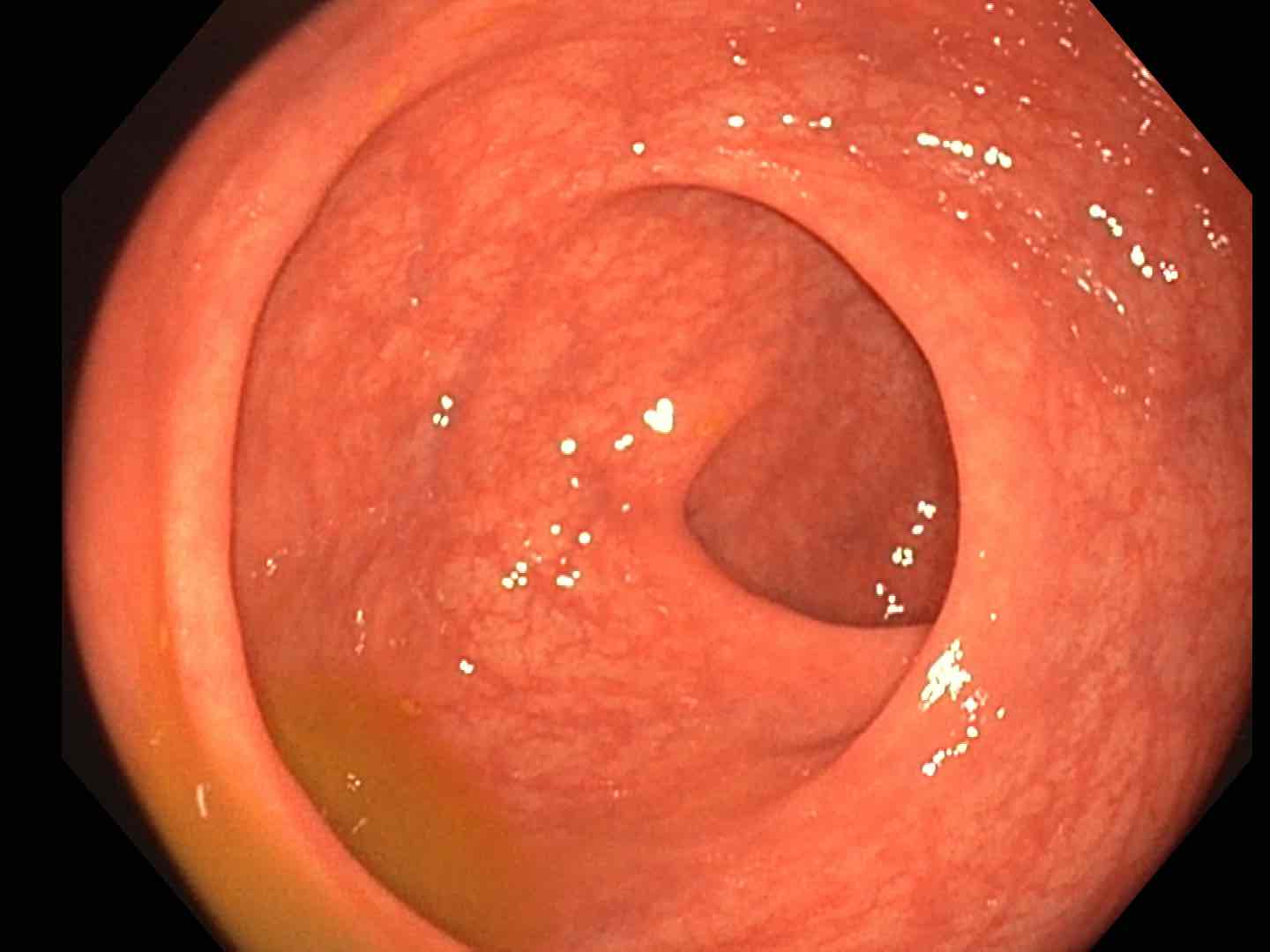}
		&\includegraphics[align=c,width=0.09\textwidth]{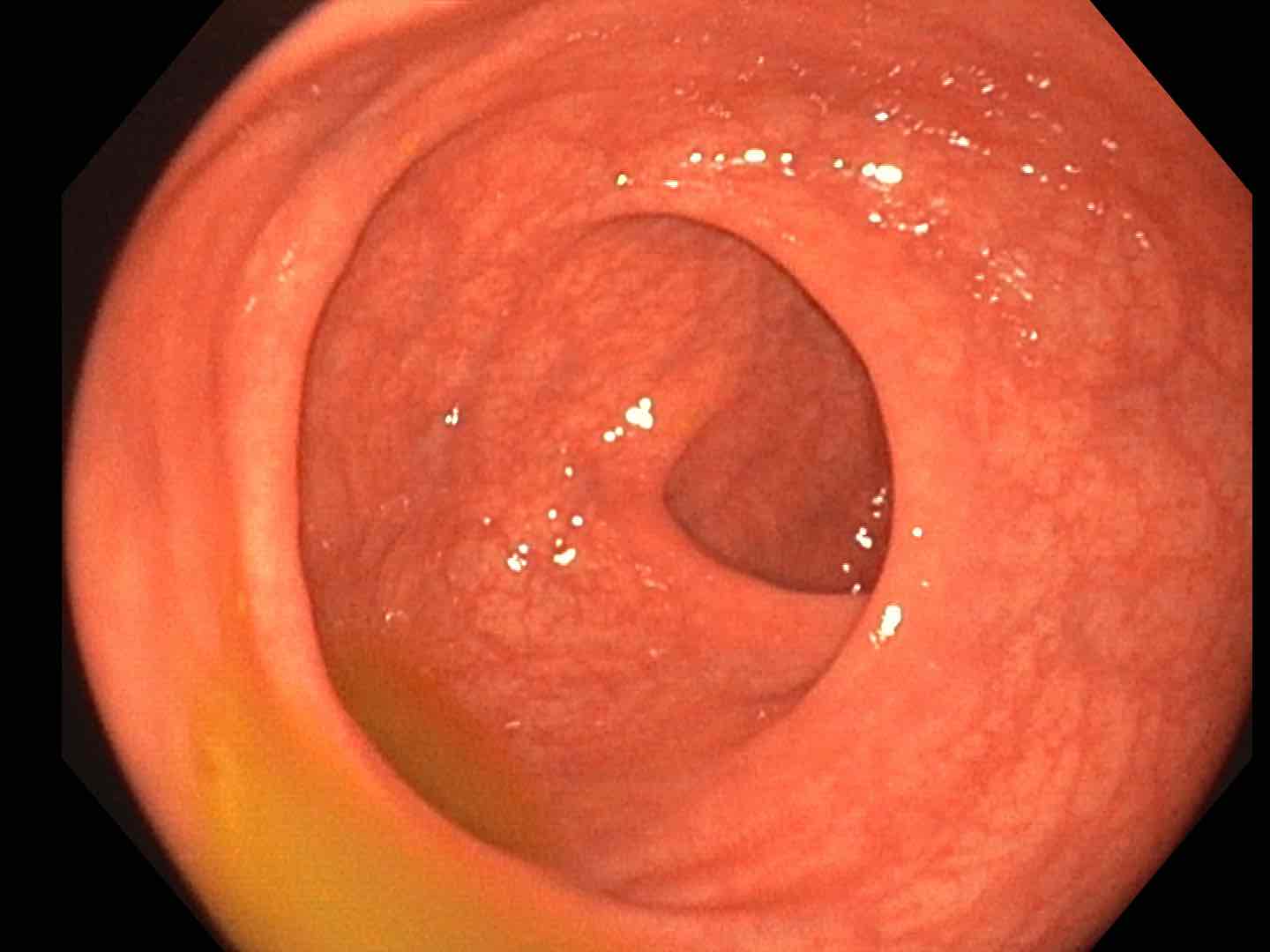}
		&\includegraphics[align=c,width=0.09\textwidth]{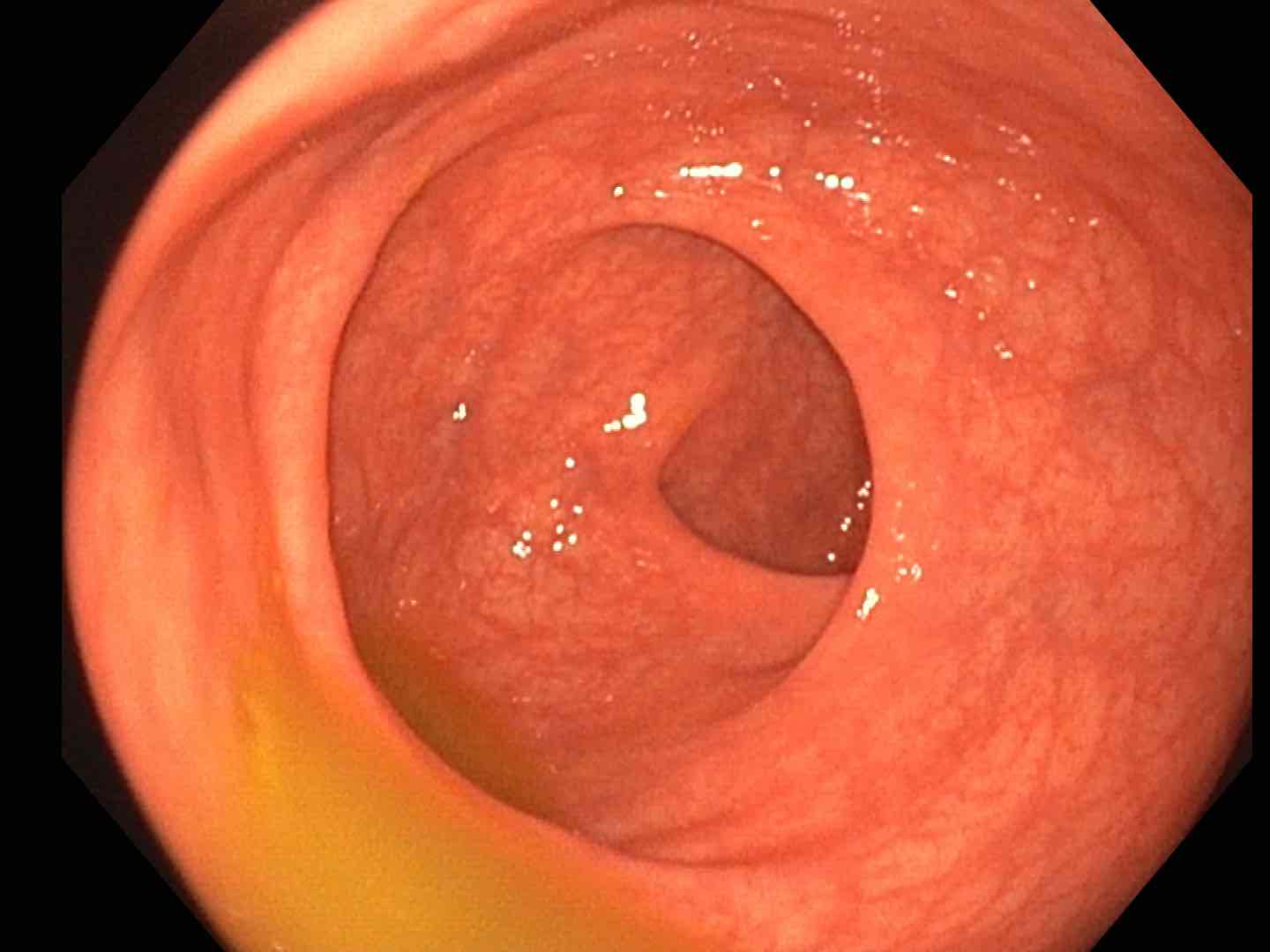}
		&\includegraphics[align=c,width=0.09\textwidth]{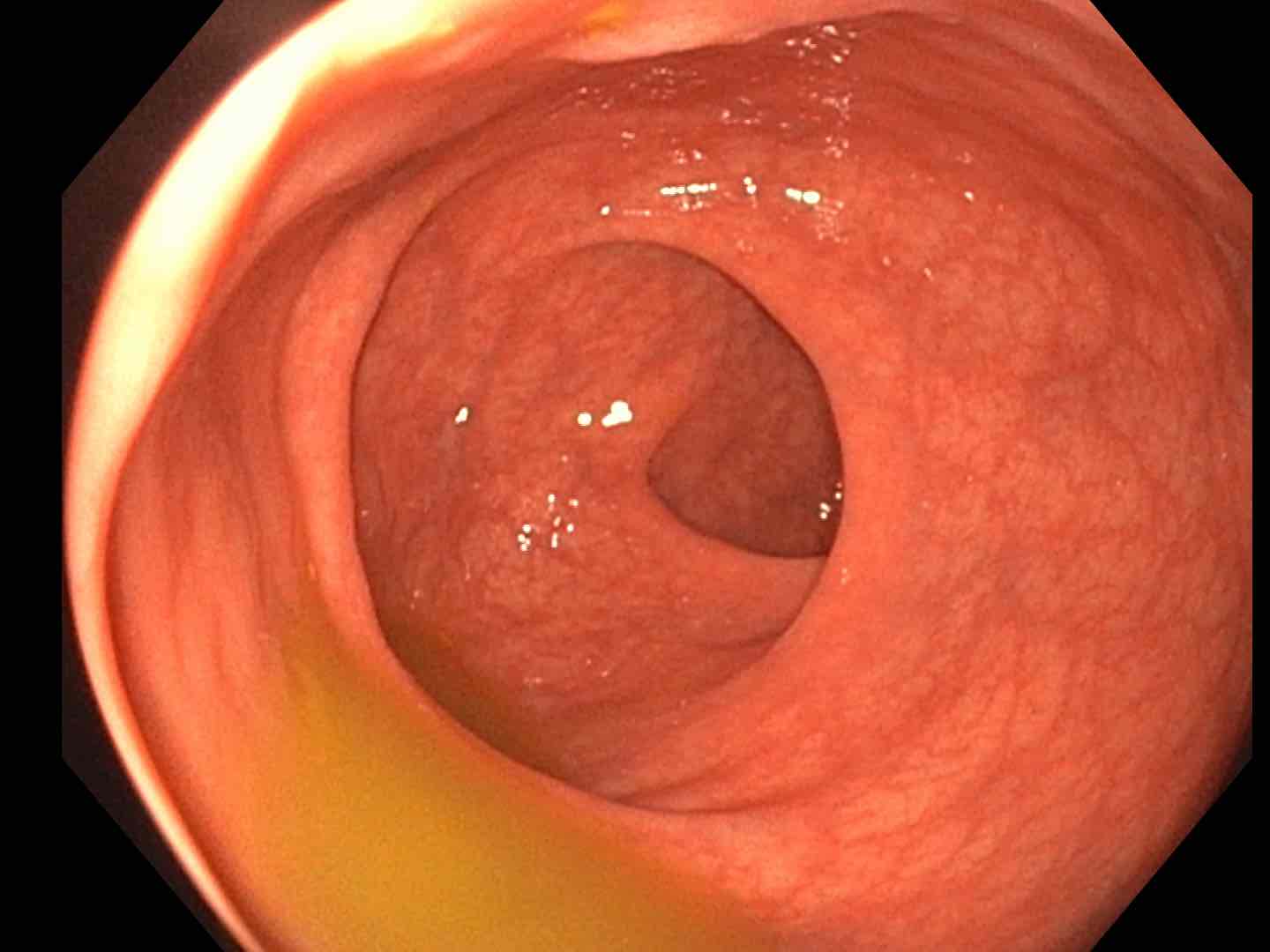}
		&\includegraphics[align=c,width=0.09\textwidth]{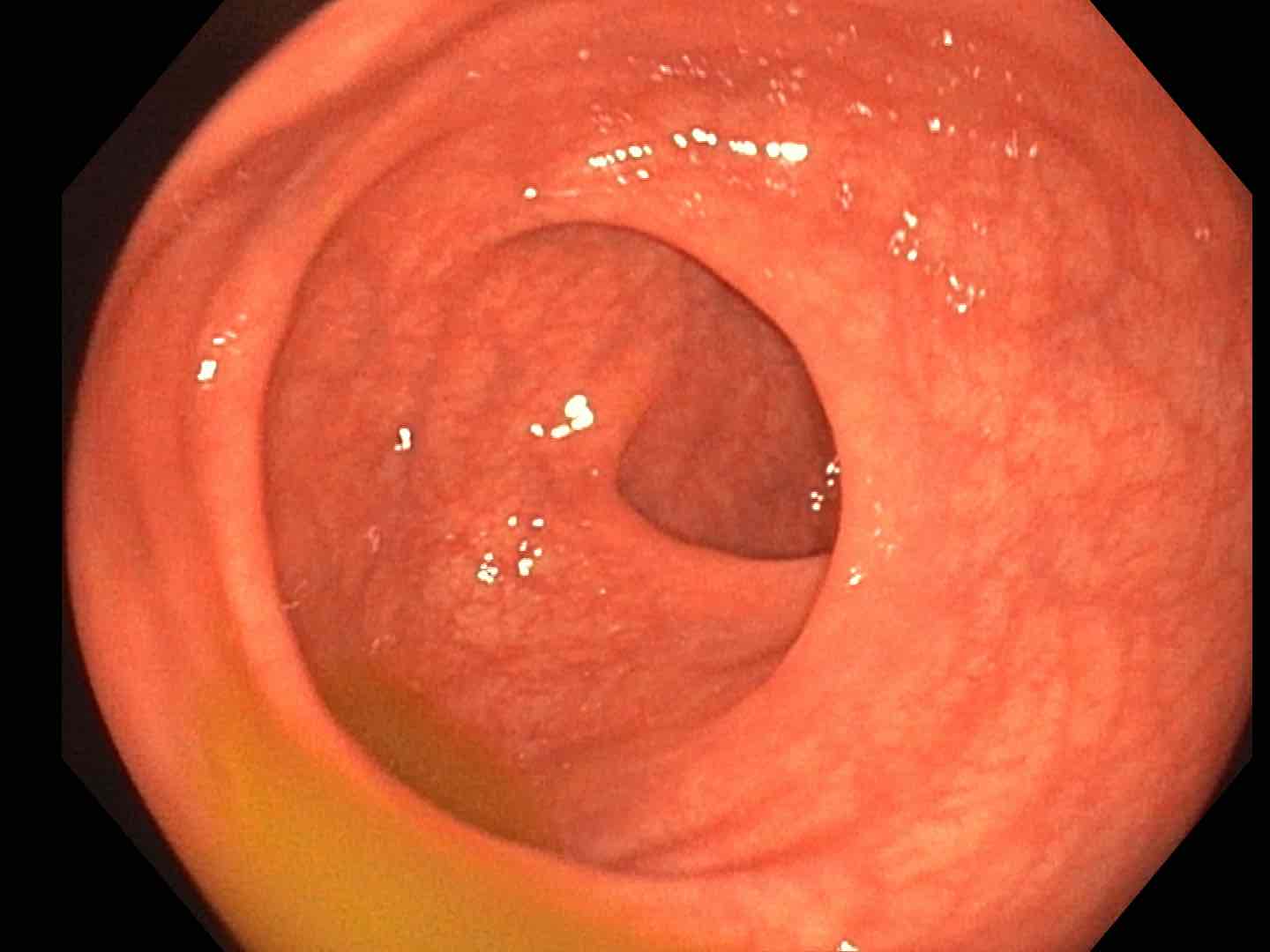}
		&\includegraphics[align=c,width=0.09\textwidth]{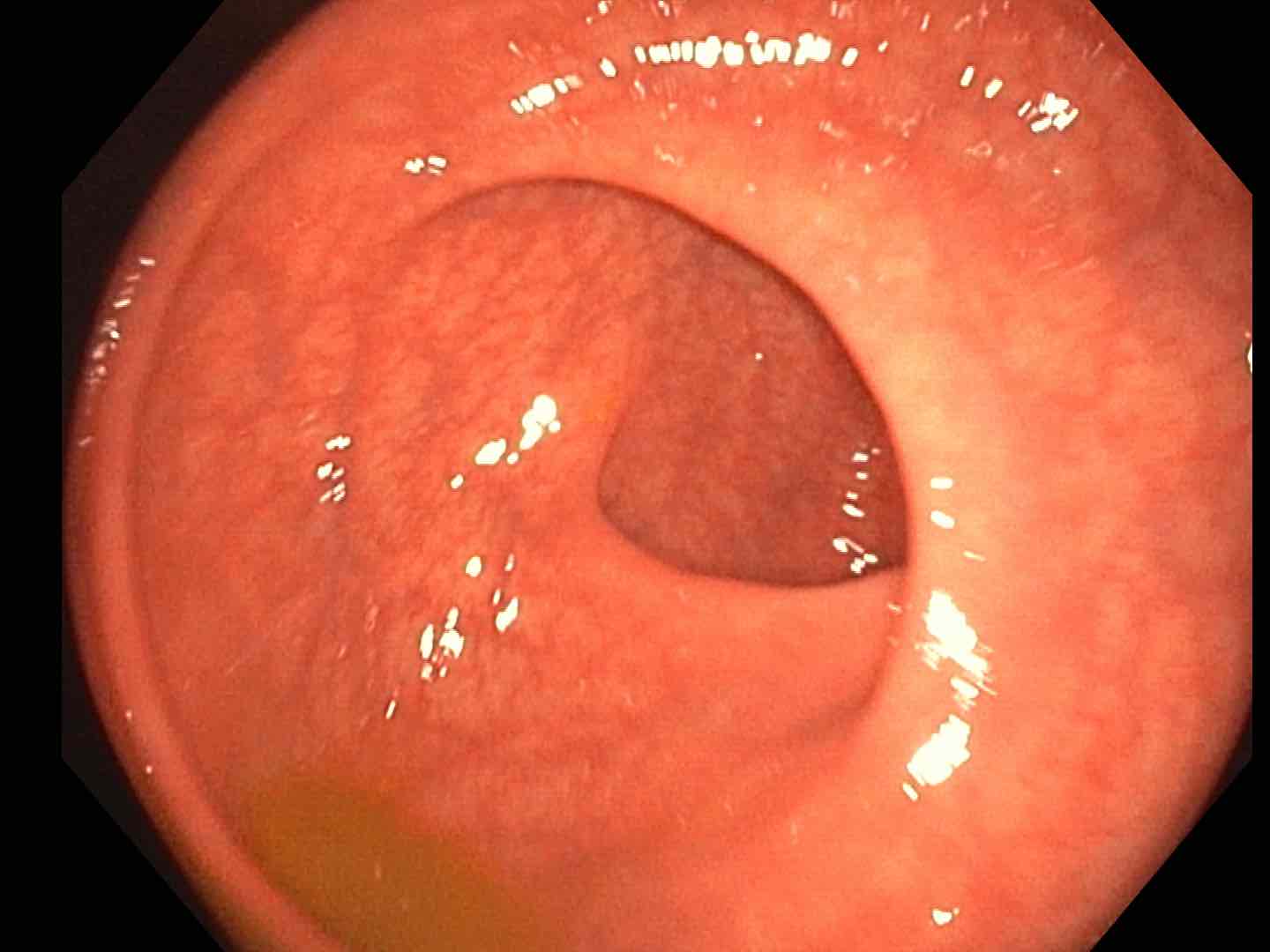}
	\end{tabular}} \\ $ $ \\
\footnotesize{ES on ES = EndoSLAM method trained on EndoSLAM data. ES on ours = EndoSLAM method trained on our data.}
	\caption{Comparison of predicted trajectories on three withdrawal (a-c), two insertion (d-e), and one withdrawal followed by insertion sequence(f).
	Top: Predicted and COLMAP trajectories. The first camera pose is set to be at (0,0,0) pointing in z-direction. Movement in positive z-direction corresponds to insertion relative to the first camera pose, while x- and y-axes represent left-right and up-down movements. Bottom: Ten images of each sequence. Self-supervised methods (green and red) fail to predict the correct direction and predict the opposite in at least half of the cases, while our method (orange) robustly predicts the correct direction along the z-axis in all cases, even when the camera changes direction as in (f).}
	\label{fig:colmap_traj}
\end{figure*}

In this section, we show that methods trained on the \textit{SimCol} data better generalize to real trajectories. We evaluate three methods on six real colonoscopy sequences form the publicly available EndoMapper dataset \cite{azagra2022endomapper}. We hand-select sequences that were successfully reconstructed with COLMAP \cite{schoenberger2016sfm} as provided by the authors of EndoMapper data. Upon visual examination, we found most reconstructed sequences to produce flat or noisy point clouds indicating errors in the reconstruction. We compared COLMAP trajectories of sequences producing reasonable 3D clouds with our own perception of the camera movement and chose the most coherent sequences as test set. Because we rely on visual inspection, the COLMAP results are not considered ground truths; however, they are a useful baseline.

In Table \ref{tab:real}, we compare three different algorithms: (i) our proposed bimodal method trained on our new dataset, (ii) the EndoSLAM algorithm using the publicly available network weights that were trained on EndoSLAM data, and (iii) the EndoSLAM algorithm trained on our data. We show the predicted poses in Figure \ref{fig:colmap_traj}. We scale trajectories resulting from the poses according to Equation \ref{eq:scale}. The errors in Table \ref{tab:real} describe the difference between COLMAP results and scaled predicted results. Let us first compare the EndoSLAM (ES) method trained on two different datasets; we will discuss the performance of our novel bimodal approach in the next section. In Table \ref{tab:real}, we can observe that ES trained on our data yields a smaller RTE in eleven out of twelve cases compared with the same method trained on the ES dataset. More importantly, in three cases, ES trained on ES fails when the same method trained on our method does not. On sequences (a), (b), and (c), the RTE produced by ES on ES is even larger than the respective mean step size. ES trained on our data better generalizes to the camera movement in the real sequence than the same method trained on the ES data. Although the domains are different, our data provided the network with more realistic camera poses to learn. Nonetheless, both ES methods fail to consistently predict the camera movements in both the original and the reverse direction. In the withdrawal sequences (a)-(c),  RTE $\rightarrow$ is especially large. Inversely, in the insertion sequences (d) - (e), RTE $\leftarrow$ is particularly large.
We plot the corresponding trajectories in the original direction in Figure \ref{fig:colmap_traj}. The first three sequences show withdrawals. While ES trained on ES fails to predict the correct direction in all three cases, ES trained on our data correctly interprets the movements as a  withdrawal in sequence (c). Sequences (d) and (e) show insertions. All methods suffer from drift, which we expect as we predict poses sequentially. However, ES trained on ES produces sharp direction changes in (d) when COLMAP's reconstruction is smooth. The last sequence shows a withdrawal followed by an insertion. Again, both ES baselines fail to predict the correct direction and the change in direction.

At first glance, it might be surprising that a synthetic dataset helps generalize to real data. Indeed, there is a domain gap between the appearance of real images and our synthetic data. However, our data has closed the gap between camera movements in existing datasets and camera movements in real colonoscopy videos. Our data allows a model to learn a correct prior for relative poses that is robust to the appearance gap.

\subsection{Bimodal distributions are more accurately learned with bimodal models} We have observed that the self-supervised baseline does not learn the bimodal pose distribution found in colonoscopies. In this section we show that supervised models are better suited to learn bimodal distributions especially when they employ a bimodal architecture. 
Table \ref{tab:real} reports how our supervised bimodal method performs on real data. Our bimodal method trained on our dataset (orange) yields the smallest RTE and ATE on almost all real sequences. ES trained on our data (red) yields the lowest ROT on $7/12$ scenes, however the lower rotation error fails to translate to a lower RTE, indicating that the method focuses on predicting accurate rotations at the expense of translations. As rotations are unimodaly distributed,  EndoSLAM is better suited to learn them than translations. Unlike ES, our method yields a similar RTE for the forward and backward directions and robustly predicts the trajectories in Figure \ref{fig:colmap_traj}. Even in sequences (a), (b), and (f), where ES fails, our method accurately follows the withdrawal and in (f) also the subsequent insertion. The bimodal model also predicts accurate trajectories on the insertion sequences (d) and (e).

Let us next evaluate our method on our proposed bimodally distributed synthetic data. In contrast to real data, our dataset provides labels for training and testing, which would make it unfair to compare a supervised method to a self-supervised one. Therefore, we compare our method to COLMAP. We apply COLMAP with standard settings to our data and report its performance in Table \ref{tab:results} (COLMAP). COLMAP only reconstructs a small subsection of images (14-18 images per trajectory corresponding to 6-8\% of frames). Accordingly, the ATE reflects only a small subsection, and we should not compare it to the ATE of other methods. The RTE and rotation error on reconstructed subsections of test sets 1 and 2 are extremely small. The RTE and rotation error on test trajectory 3 is comparable to ours, though our approach is robust and predicts the entire trajectory. CNNs for pose prediction in colonoscopy can thus be considered a practical alternative where feature-based methods fail to initialize.
\begin{figure}
	\centering
	\includegraphics[width=\columnwidth]{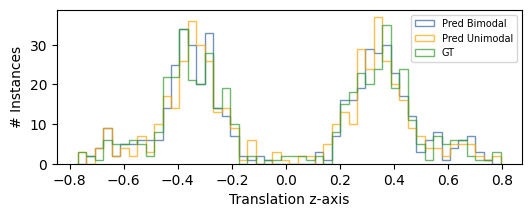}
	\caption{Histograms of z-translations in cm for our bimodal model,  unimodal model, and ground truth (GT) on test trajectory 1.}
	\label{fig:manhattan}
\end{figure}
\begin{table}
	\caption{Manhattan (L1-) errors for z-translation predicted by different models on our three test trajectories}
	\setlength{\tabcolsep}{3pt}
	\begin{center}
	{\def\arraystretch{1.2}
	\begin{tabular}{|p{100pt}|c|c|c|}
		\hline
		Test trajectory &1 &2 & 3 \\
		\hline
		\hline
		Bimodal, class supervision&132&\textbf{122}&\textbf{154}\\
		Unimodal, no supervision&168&152&162\\
		Bimodal, no supervision&\textbf{118}&142&170\\
		\hline
	\end{tabular}}\\ $ $ \\
	\end{center}
	\label{tab:manhattan}
\end{table}
Next, we evaluate the usefulness of a bimodal model compared to a unimodal approach in two experiments. We compare three different versions, all supervised with the ground truth pose: (i) a bimodal model with class supervision; (ii) a unimodal model that passes the ResNet-output to the pose net only; and (iii) a bimodal model that is \textit{not} trained with class labels. We implemented all models in PyTorch. For the bimodal models, we use the hyper-parameters $bin1 = 0.1\cdot k$, $bin2 = -0.1\cdot k$, $w_c = 0.1$, and $k=5$, where $w_c$, and $k$ were chosen based on grid search. The unimodal and bimodal models have the same number of weights in the pose net.
We compare the resulting histograms and pose errors of the models:

Firstly, we plot histograms of the predicted poses in Figure \ref{fig:manhattan}. Table \ref{tab:manhattan} evaluates the similarity of the ground truth and predicted histograms using the Manhattan (L1-) loss
\begin{align}
    L_{Manhattan} = \sum_{i} |h(i)_{gt} - h(i)_{pred}|.
\end{align}
From Table \ref{tab:manhattan} we gather that the bimodal model supervised with cross-entropy more accurately learns the distribution of z-translations in the test set than a unimodal model for all three test trajectories. All three models, however, closely replicate the true distribution (see Figure \ref{fig:manhattan}). 

\begin{figure}
\begin{minipage}{0.24\textwidth}
\centering
         	\includegraphics[width=\columnwidth]{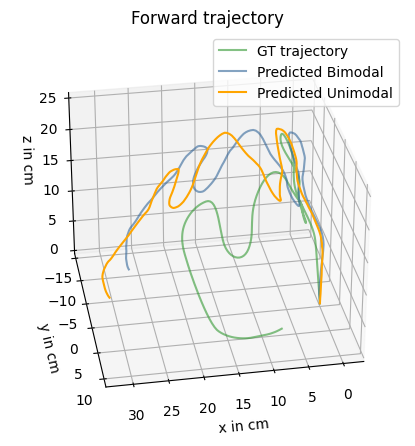}

\end{minipage}
\begin{minipage}{0.24\textwidth}
\centering
         	\includegraphics[width=\columnwidth]{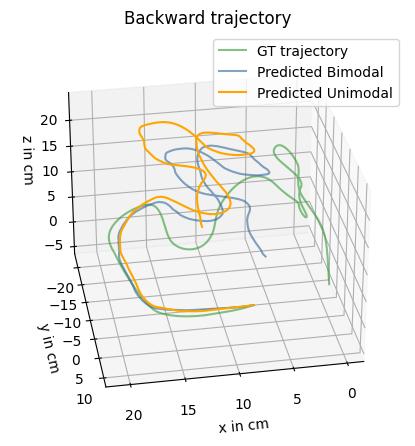}

\end{minipage}
\caption{Comparison of predicted trajectories produced by unimodal and our bimodal model and ground truth (GT) on test set 1. Please see supplementary material for videos.}
	\label{fig:supertraj}
\end{figure}

Secondly, we evaluate the RTE, ATE, and ROT as defined in Equations \ref{eq:ate}-\ref{eq:rot} on three test trajectories in Table \ref{tab:results}. However, we do not rescale the results according to Equation \ref{eq:scale}, as we expect the network to learn the absolute scale. 
As we estimate the pose between a frame pair that is $k=5$ frames apart, we calculate the errors for the five trajectories that begin with step 0, ..., 4 and report the average. For each trajectory, we predict poses in forward and backward directions. The bimodal approach with supervision yields the smallest RTE for each test trajectory and, therefore, the most accurate local camera poses. However, even the bimodal approach without class supervision yields smaller RTEs than the unimodal approach on all scenes. The network learns to distinguish bins without supervision, speaking to the ability of the correlation layer to distill information about relative movements. The higher error for our bimodal approach with class supervision but without correlation layer further supports the claim. To visualize the ATE, we plot whole trajectories for our bimodal model and the unimodal counterpart in Figure \ref{fig:supertraj}. The total length of the trajectory is roughly $1$ meter, so drift is expected. 

\subsection{Limitations}
We propose a dataset that replicates the movement of a colonoscope and show that methods trained on this data more robustly generalize to real video. Our bimodal approach improves supervised methods that regress relative camera pose directly. Yet, the applicability to real colonoscopy could be improved. The proposed dataset closes the label domain gap, but differences to real data remain and should be addressed in future work, like a lack of deformations. The appearance gap has not been approached by our method but can be addressed by domain adaptation methods. For a full 3D reconstruction, besides accurate poses and depths, global optimization and loop closure are required, as drift is universal when using local methods. Computational limitations are yet to be overcome to allow an on-the-fly reconstruction to give immediate feedback to colonoscopists for a higher clinical impact.

\section{Conclusions}
Relative camera pose prediction during colonoscopy remains exceptionally challenging. This work explains the limitations of self-supervised methods based on warping losses for camera pose prediction. We propose using supervised methods trained on synthetic data and show that the explicit modeling of a bimodal model can improve the accuracy of relative camera pose methods. Supervised methods have the additional benefit of being more robust to a lack of features and lighting inconsistency than standard feature-based SfM methods. Our dataset closes the domain gap between real and synthetic relative camera poses. Future work will solve the domain gap between the appearance of real and synthetic colonoscopy images, for instance, with generative networks as proposed in \cite{rau2019implicit}. Although supervised methods are currently better suited to learn bimodal distributions, an exciting direction for future work could be to further investigate bimodality within warping losses of self-supervised models.

\bibliography{bibliography.bib}
\bibliographystyle{ieeetr}
\end{document}